\documentclass{article}
\pdfoutput=1
\usepackage[preprint]{neurips_2023}
\usepackage[utf8]{inputenc} %
\usepackage[T1]{fontenc}    %
\usepackage{url}            %
\usepackage{booktabs}       %
\usepackage{amsfonts}       %
\usepackage{amsmath}       %
\usepackage{nicefrac}       %
\usepackage{microtype}      %
\usepackage{lipsum}
\usepackage{graphicx}
\usepackage{wrapfig}
\usepackage[size=small]{caption}
\usepackage{mathtools}
\usepackage{amssymb}
\usepackage{amsthm}
\usepackage[algoruled,boxed,lined,noend]{algorithm2e}
\usepackage{adjustbox} 
\usepackage{float}
\usepackage{amssymb}
\usepackage{pifont}
\usepackage{subcaption}
\newcommand{\cmark}{\ding{51}}%
\newcommand{\xmark}{\ding{55}}%
\usepackage{todonotes}
\usepackage{lineno}

\usepackage[superscript,biblabel]{cite}

\usepackage[dvipsnames]{xcolor}
\usepackage{booktabs}
\usepackage{multirow}
\usepackage{tabularx}

\definecolor{cvprblue}{rgb}{0.21,0.49,0.74}
\definecolor{citecolor}{HTML}{0071BC}
\definecolor{linkcolor}{HTML}{ED1C24}
\usepackage[colorlinks=true, linkcolor=blue]{hyperref}
\usepackage[capitalize]{cleveref}
\crefname{section}{Sec.}{Secs.}
\crefname{table}{Table}{Tables}
\crefname{figure}{Fig.}{Figs.}
\usepackage{placeins}
\usepackage[most]{tcolorbox}
\usepackage{float}
\usepackage{xspace}
\tcbset{
  aibox/.style={
    width=394.18663pt,
    top=10pt,
    colback=white,
    colframe=black,
    colbacktitle=black,
    enhanced,
    center,
    attach boxed title to top left={yshift=-0.1in,xshift=0.15in},
    boxed title style={boxrule=0pt,colframe=white,},
  }
}
\newtcolorbox{AIbox}[2][]{aibox,title=#2,#1}

\usepackage{colortbl}

\definecolor{qual-fig-green}{RGB}{0,144,11}
\definecolor{qual-fig-red}{RGB}{238,0,0}
\definecolor{qual-fig-purple}{RGB}{153,51,255}
\usepackage[dvipsnames]{xcolor}

\newcommand{\stoptocwriting}{\addtocontents{toc}{\protect\setcounter{tocdepth}{-5}}}
\newcommand{\resumetocwriting}{\addtocontents{toc}{\protect\setcounter{tocdepth}{\arabic{tocdepth}}}}

\usepackage{authblk}  

\title{Merlin: A Computed Tomography Vision-Language Foundation Model and Dataset}

\author[1,2,3$\dagger$]{\textbf{Louis Blankemeier}\thanks{Co-First Authors. \newline Corresponding author: akshaysc@stanford.edu}}
\author[2,3]{\textbf{Ashwin Kumar}$^*$}
\author[2,$\ddagger$]{\textbf{Joseph Paul Cohen}}
\author[2,3]{\textbf{Jiaming Liu}}
\author[11]{\textbf{Longchao
Liu}}
\author[1,2,3]{\textbf{Dave Van Veen}}
\author[4]{\textbf{Syed Jamal Safdar Gardezi}}
\author[5]{\textbf{Hongkun Yu}}
\author[2,3]{\textbf{Magdalini Paschali}}
\author[2,3]{\textbf{Zhihong Chen}}
\author[2,3]{\textbf{Jean-Benoit Delbrouck}}
\author[2,3]{\textbf{Eduardo Reis}}
\author[2,3]{\textbf{Robbie Holland}}
\author[5]{\textbf{Cesar Truyts}}
\author[2,6]{\textbf{Christian Bluethgen}}
\author[12]{\textbf{Yufu Wu}}
\author[13]{\textbf{Long Lian}}
\author[2,3]{\textbf{Malte Engmann Kjeldskov Jensen}}
\author[2,3]{\textbf{Sophie Ostmeier}}
\author[2,3,7]{\textbf{Maya Varma}}
\author[2,3,7]{\textbf{Jeya Maria Jose Valanarasu}}
\author[2,3]{\textbf{Zhongnan Fang}}
\author[8]{\textbf{Zepeng Huo}}
\author[ ]{\textbf{Zaid Nabulsi}$^\ddagger$, \textbf{Diego Ardila}$^\ddagger$, \textbf{Wei-Hung Weng}$^\ddagger$}
\author[5]{\textbf{Edson Amaro Junior}}
\author[9]{\textbf{Neera Ahuja}}
\author[7,8]{\textbf{Jason Fries}}
\author[2,9]{\textbf{Nigam H. Shah}}
\author[2,3]{\textbf{Greg Zaharchuk}}
\author[3]{\textbf{Marc Willis}}
\author[11]{\textbf{Adam Yala}}
\author[3]{\textbf{Andrew Johnston}}
\author[3]{\textbf{Robert D. Boutin}}
\author[4]{\textbf{Andrew Wentland}}
\author[2,3,8]{\textbf{Curtis P. Langlotz}}
\author[9]{\textbf{Jason Hom}}
\author[5]{\textbf{Sergios Gatidis}}
\author[2,3,8,10]{\textbf{Akshay S. Chaudhari}}

\affil[1]{Department of Electrical Engineering, Stanford University}
\affil[2]{Stanford Center for Artificial Intelligence in Medicine and Imaging, Stanford University}
\affil[3]{Department of Radiology, Stanford University}
\affil[4]{Department of Radiology, University of Wisconsin-Madison}
\affil[5]{Department of Radiology, Hospital Israelita Albert Einstein}
\affil[6]{Department of Radiology, University Hospital Zurich}
\affil[7]{Department of Computer Science, Stanford University}
\affil[8]{Department of Biomedical Data Science, Stanford University}
\affil[9]{Department of Medicine, Stanford University}
\affil[10]{Stanford Cardiovascular Institute, Stanford University}
\affil[11]{Computational Precision Health, University of California Berkeley}
\affil[12]{Department of Medical Imaging and Intervention, Chang Gung Memorial Hospital at Linkou}
\affil[13]{Department of Electrical Engineering and Computer Science, University of California Berkeley}

\begin{document}
\stoptocwriting
\maketitle
\begin{abstract}

Over 85 million computed tomography (CT) scans are performed annually in the US, of which approximately one quarter focus on the abdomen. Given the current shortage of both general and specialized radiologists, there is a large impetus to use artificial intelligence to alleviate the burden of interpreting these complex imaging studies while simultaneously using the images to extract novel physiological insights. Prior state-of-the-art approaches for automated medical image interpretation leverage vision language models (VLMs) that utilize both the image and the corresponding textual radiology reports. However, current medical VLMs are generally limited to 2D images and short reports. To overcome these shortcomings for abdominal CT interpretation, we introduce \emph{Merlin} - a 3D VLM that leverages both structured electronic health records (EHR) and unstructured radiology reports for pretraining without requiring additional manual annotations. We train Merlin using a high-quality clinical dataset of paired CT scans (6+ million images from 15,331 CTs), EHR diagnosis codes (1.8+ million codes), and radiology reports (6+ million tokens) for training. We comprehensively evaluate Merlin on 6 task types and 752 individual tasks, covering a variety of diagnostic, prognostic, and quality-related tasks. The non-adapted (off-the-shelf) tasks include zero-shot findings classification (31 findings), phenotype classification (692 phenotypes), and zero-shot cross-modal retrieval (image to findings and image to impressions), while model adapted tasks include 5-year chronic disease prediction (6 diseases), radiology report generation, and 3D semantic segmentation (20 organs). We perform internal validation on a test set of 5,137 CTs, and external validation on 44,098 CTs from three external sites spanning both abdomen and chest anatomies, and on two public CT datasets (VerSe, TotalSegmentator). Beyond these clinically-relevant evaluations, we assess the efficacy of various network architectures and training strategies to depict that Merlin has favorable performance to existing task-specific baselines. We derive data scaling laws to empirically assess training data needs for requisite downstream task performance. Our resource-friendly approach enables health systems to train their own foundation models with modest computational resources. We also release our trained models, code, and dataset, available at: \href{https://github.com/StanfordMIMI/Merlin}{https://github.com/StanfordMIMI/Merlin}.
\end{abstract}
\footnotetext{$^\ddagger$Work not related to position at Amazon or Google.}
\section{Main}
\label{sec:intro}
Over 85 million computed tomography (CT) scans are performed per year in the US~\cite{computed_tomography_exams,Harvard_Health_2021,winder2021we}, with approximately 300 million CTs performed globally~\cite{schockel2020developments}. Amongst these studies, CTs of the abdomen represent approximately one fourth of all examinations performed~\cite{doi:10.1148/radiol.2017161911}. These abdominal CT scans can consist of upwards of 300 slices per series with numerous anatomical structures that need to be examined, leading to time-consuming interpretation, often requiring 20 minutes per exam~\cite{udare2022radiologist}. Moreover, recent literature suggests that abdominal CT scans contain biomarkers of early diseases that routinely go unreported~\cite{bellolio2017increased, liu2022fully, lee2022abdominal, thibault2012body, bates2022ct, kuriyan2018body, blankemeier2022opportunistic, zambrano2023opportunistic}. With the current volume of medical imaging and a 6\% annual~\cite{Imaging_Technology_News_2024} increase in medical imaging utilization, the burden on radiologists is significant. Nonetheless, over the past decades, the number of radiology residency positions in the US has remained relatively constant at 1011 in 2006 and 1113 in 2020~\cite{wadhwa202215}. This is despite 1,800+ open radiologist positions on the American College of Radiology job board~\cite{acr_job_board} from a pool of roughly 32,000 total active radiologists in the US~\cite{num_radiologists}\footnote{Number of job openings on the ACR job board based on accessing the site on May 26, 2024.}. With the supply of new radiologists remaining relatively constant and an ever increasing utilization of medical imaging, the radiologist shortage is projected to expand to over 19,000 by 2036~\cite{practice_management_quality_informatics,How_Will_We_Solve_Our_Radiology_Workforce_Shortage?_|_American_College_of_Radiology, patel2020radiologists, rimmer2017radiologist}.

Machine learning (ML) has shown promise in various medical imaging tasks~\cite{ardila2019end,engautomated,cao2023large,wang2024screening}, inspiring optimism about its potential to offset the increasing burden faced by radiologists~\cite{practice_management_quality_informatics,langlotz2023future}. As of May 2024, of the 882 FDA-cleared ML-enabled devices, 671 (76\%) relate to radiology~\cite{number_of_ai_enabled_devices}. Despite this large prevalence, the current status-quo of training ML algorithms entails using unimodal (imaging-only) algorithms and retraining for new tasks from scratch using supervised ML with manually curated labels, even if it may be for the same modality or anatomy. In the medical imaging scenario, generating such labels requires expensive clinical expert time, limiting the development of capable AI models for a wide array of tasks.

Recent years have witnessed remarkable advancements in vision-language models (VLMs), a data-efficient alternative to supervised training. The contrastive language-image pretraining (CLIP) technique ~\cite{radford2021clip} demonstrated the efficacy of aligning text and visual representations in a shared embedding space as a means of supervising vision models with natural language. This paradigm enables leveraging internet-scale images and captions, demonstrating impressive image understanding capabilities in off-the-shelf (zero shot) settings or in settings that use subsequent adaptation (few shot learning)~\cite{schuhmann2022laion} for a large number of downstream tasks. Such models, trained on large-scale multi-modal pretraining datasets and enabling adaptation for multiple downstream tasks, are commonly referred to as foundation models. 

CLIP-based methods could be readily applied in the clinical setting by training with medical images and corresponding radiology reports that are generated routinely during clinical care, adding no additional data labeling cost. Many institutions, including ours, de-identify this data to maintain patient privacy while providing a unique setting for research on this high-quality human-annotated data. Recent ethical viewpoints describe how large-scale already-acquired clinical data could be responsibly used for secondary purposes such as training ML algorithms, \textit{ to ensure that the data are used for the benefit of future patients and society~\cite{larson2020ethics}}.  
Following such frameworks, VLMs like MedCLIP \cite{wang2022medclip}, BiomedCLIP \cite{zhang2023large}, LLaVA-Med \cite{li2023llava}, Med-Flamingo \cite{moor2023med}, Med-PaLM M \cite{tu2023towards}, RadFM \cite{wu2023generalist}, XrayGPT \cite{thawkar2023xraygpt}, ELIXR~\cite{xu2023elixr}, RoentGen~\cite{chambon2022adapting, bluethgen2024roentgen}, CheXagent~\cite{chen2024chexagent}, and MAIRA-1~\cite{hyland2023maira} demonstrate the potential of VLMs applied within the radiology domain.

Despite the burgeoning popularity of VLMs for radiology, most existing models focus on 2D modalities such as radiographs, notwithstanding that most medical imaging studies are 3D in nature\footnote{We note that in the clinical context, 3D medical imaging is considered as a subset of cross-sectional imaging where the pixels are isotropic. In this paper, we use the term 3D to align with the terminology used by the machine learning community describing the simultaneous processing of pixels originating from 3 spatial dimensions}. Many approaches extend 2D models to 3D by aggregating predictions slice-by-slice or in chunks of slices across the 2D image stacks that make up the 3D volume~\cite{huang2020penet, christensen2024vision} - an inefficient technique to parse the full 3D imaging volume. Unlike analysis of video, where successive frames have high correlation, there is limited 3D correlation in volumetric anatomical structures that rapidly change in all dimensions. This scenario is well-suited for 3D modeling where features spanning all 3 dimensions are synthesized to generate insights. Moreover, existing methods do not leverage supervision from the multiple data types, including EHR diagnosis codes and radiology reports, available in clinical settings.

The potential clinical impact of 3D medical imaging models may be significant. 3D medical VLMs could assist radiologists by flagging missed findings~\cite{patel2020radiologists}, accelerating image interpretation workflows, and serving as AI assistants that draft radiology reports~\cite{doi:10.1148/radiol.12121502,langlotz2023future}. The opportunity is particularly substantial for abdominal CT exams, which are the most commonly utilized 3D examination~\cite{doi:10.1148/radiol.2017161911} and require significant time for interpretation due to the number of anatomical structures that need to be examined~\cite{udare2022radiologist}.

Nonetheless, adapting existing methods for training VLMs to 3D medical imaging presents a challenge as a single volumetric image can comprise upwards of 300 individual 2D images, and the corresponding radiology reports can exceed 300 words~\cite{udare2022radiologist}. Training models on these large data samples can also require significant computational resources, which are often not available in academic institutions or within hospital systems where the data resides~\cite{ahmed2023growing}. Furthermore, clinically-relevant evaluations for benchmarking 3D VLMs across a suite of tasks is still lacking. Even in the clinical large language model (LLM) space, where progress is rapid, popular benchmarks based on medical licensing exams garner criticism for not reflecting real-world clinical use cases~\cite{sergeiai,fleming2023medalign}. Due to these training and evaluation challenges, there exists a dearth of methods for training and evaluating 3D medical imaging VLMs that can be adapted for a wide range of downstream tasks.

In this paper, we focus on developing and evaluating foundation VLMs for 3D abdominal CT scans to enhance image understanding across a variety of tasks. Our work provides the following contributions:

\begin{enumerate}
    \item We develop a training strategy and foundation model called \emph{Merlin}\footnote{Named after the legendary figure said to be able to perceive everything in the present, our model Merlin simultaneously processes the entirety of 3D information in a CT volume.} that leverages the structured and unstructured data within hospitals to train an abdominal CT visual model. Using this training strategy, we train a vision-language model on a single GPU using paired CTs (6,387,231 images from 15,331 CTs), EHR diagnosis codes (1,839,559 codes), and radiology reports (6,036,645 tokens). Merlin allows processing the entire 3D voxel data in a CT image at once.
    \item We evaluate Merlin on a comprehensive set of 6 task types and 752 individual tasks, covering a variety of diagnostic, prognostic, and quality-related tasks. The non-adapted (off-the-shelf tasks) include zero-shot findings classification (31 findings), phenotype classification (692 phenotypes), and zero-shot cross-modal retrieval (image to findings and image to impressions), while model adapted tasks include 5-year disease prediction (6 diseases), radiology report generation, and 3D semantic segmentation (20 organs). 
    \item  We perform internal validation on 5,137 CTs, as well as external validation on 44,098 CTs from three external sites spanning both abdomen and chest anatomies, and on two publicly available datasets focused on abdominal CT (VerSe~\cite{liebl2021computed} and TotalSegmentator~\cite{wasserthal2023totalsegmentator}). We demonstrate that our singular Merlin model outperforms carefully chosen task-specific baselines on our suite of benchmarking tasks.
    \item We compare Merlin to state-of-the-art finetuned 2D VLMs, 2D-to-3D lifted VLMs, and 3D vision-only models and show that (i) Merlin's vision-language pretraining outperforms vision-only pretraining, (ii) finetuning improves the performance of 2D and 2D-to-3D VLMs though they underperform relative to Merlin, and (iii) Merlin outperforms other baselines in both data scarce and fully supervised settings.
    \item We derive data scaling laws for Merlin, providing guidance about the data requirements for achieving specific levels of performance. We also present comprehensive ablation studies demonstrating the impact of various training strategies and the role of the different clinical data types on model performance.
    \item We release our trained \href{https://huggingface.co/stanfordmimi/Merlin}{models}, \href{https://github.com/StanfordMIMI/Merlin}{code}, and \href{https://stanfordaimi.azurewebsites.net/datasets/60b9c7ff-877b-48ce-96c3-0194c8205c40}{dataset}. Our dataset, the Merlin Abdominal CT dataset, has been manually reviewed to ensure that there is no personal health information in 25,494 CT images and corresponding radiology reports.
\end{enumerate}
\section{Results}
\begin{figure}[th!]
\centering
\includegraphics[width=1.0\textwidth]{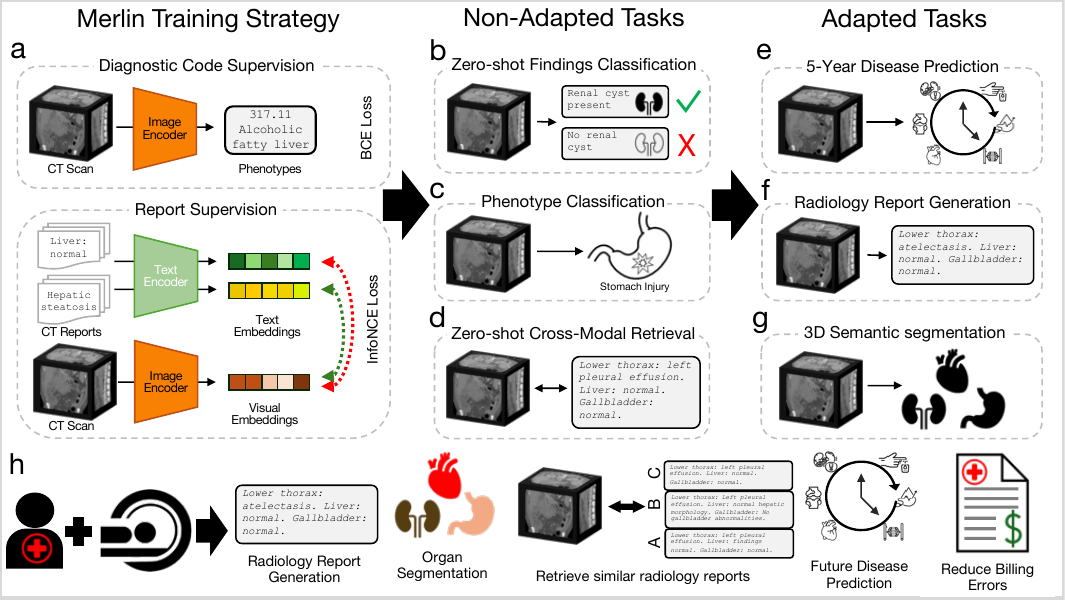}
\caption{\textit{Overview of Merlin training and evaluation.} (a) Merlin training strategy. Diagnosis codes from the EHR are used as labels for Merlin training, with a binary cross entropy loss. Radiology reports are also used for training, with an InfoNCE loss~\cite{oord2018representation}. Training with diagnosis codes and radiology reports is either staged or performed in a multi-task manner. Merlin is then evaluated on non-adapted tasks that can be performed without any architectural or weight modifications. These include (b) zero-shot findings classification, (c) phenotype classification, and (d) zero-shot cross-modal retrieval. Adapting Merlin enables us to perform (e) 5-year disease prediction, (f) radiology report generation, and (g) 3D semantic segmentation. (h) Illustration of Merlin’s potential clinical applications. Merlin can assist radiologists by generating radiology reports and segmenting organs from CT scans. Additionally, it can facilitate the retrieval of radiology reports from patients with similar CT scans, aid clinicians in identifying potential future disease for patients, and help minimize billing errors by assigning ICD codes.}
\label{fig:overview_figure}
\end{figure}

We present results across 6 evaluation task types comprising 752 individual tasks, that can be performed off-the-shelf without adaptation (Figure \ref{fig:overview_figure}b - Figure \ref{fig:overview_figure}d), or with additional adaptation (Figure \ref{fig:overview_figure}e - Figure \ref{fig:overview_figure}g). The non-adapted tasks that we evaluate Merlin on include zero-shot classification (31 classification tasks; Figure \ref{fig:overview_figure}b), phenotype classification (692 classification tasks; Figure \ref{fig:overview_figure}c), and zero-shot cross-modal retrieval (image to findings and image to impressions; Figure \ref{fig:overview_figure}d). The adapted tasks that we evaluate Merlin on include 5-year disease prediction (6 classification tasks; Figure \ref{fig:overview_figure}e), radiology report generation (Figure \ref{fig:overview_figure}f), and 3D segmentation (20 organs; Figure \ref{fig:overview_figure}g). We report that training Merlin required approximately 160 hours on a single NVIDIA A6000 GPU.

\subsection{Zero-shot Findings Classification}
\begin{figure}[h!]
\centering
\includegraphics[width=0.95\textwidth]{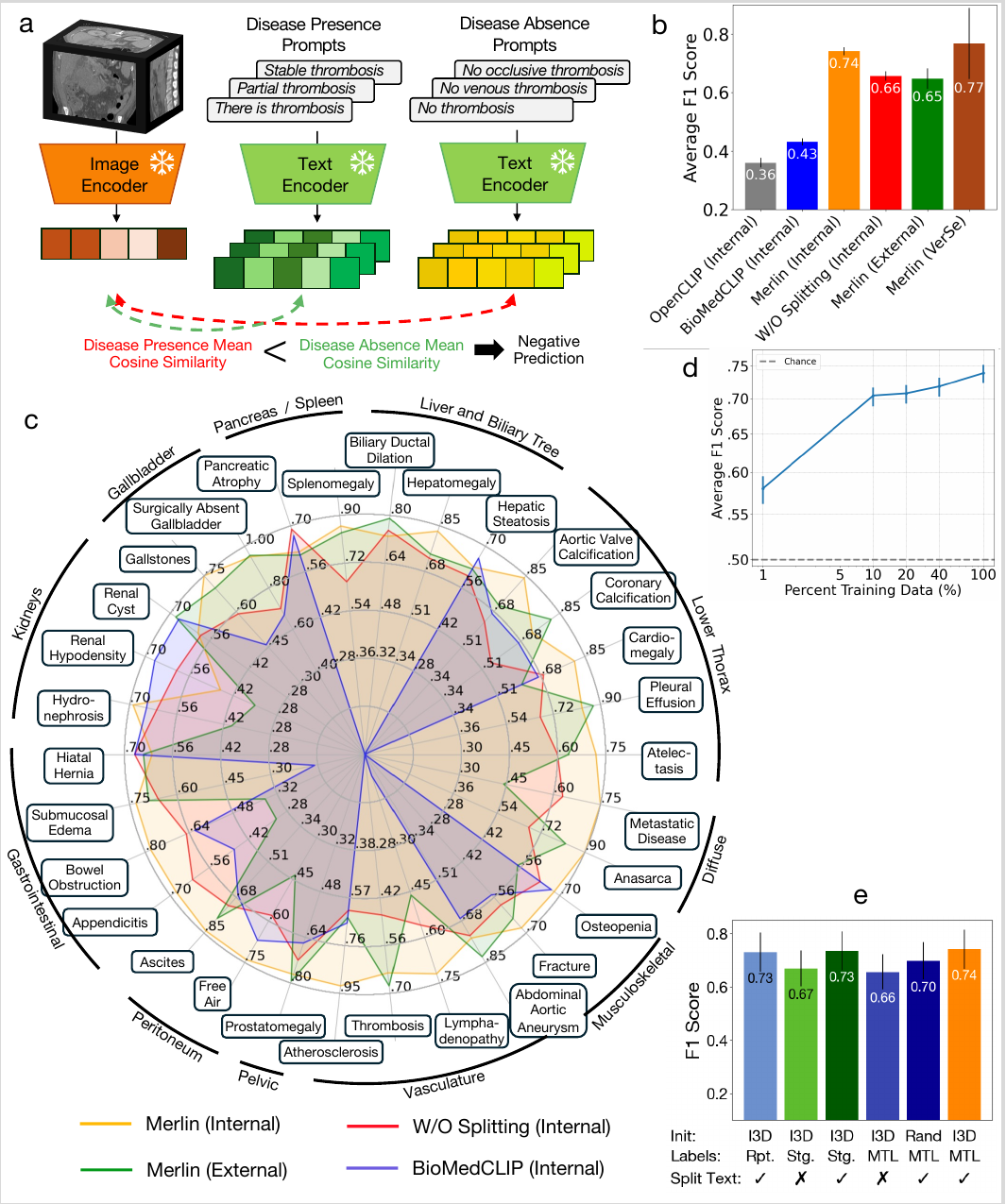}
\caption{\textit{Zero-shot findings classification.} (a) Depicts how zero-shot classification is performed where text embeddings from disease presence prompts and disease absence prompts are compare to the image embedding. (b) We compare the performance of off-the-shelf 2D OpenCLIP~\cite{cherti2022reproducible} with K=1 pooling, finetuned 2D BioMedCLIP~\cite{zhang2023biomedclip} with average pooling, Merlin on an internal dataset, and Merlin without radiology report splitting. We further evaluate Merlin on an external clinical dataset and the VerSe~\cite{loffler2020vertebral} external fracture detection dataset. (c) Performance of finetuned 2D BioMedCLIP, Merlin, and Merlin without report splitting on the internal dataset, as well as Merlin on the external dataset, across 30 findings assessed on abdominal CT scans. (d) Merlin zero-shot classification performance improves with increasing pretraining dataset size. (e) An ablation study across various aspects of Merlin's pretraining strategy. "Rpt." is shorthand for "report" and indicated training with radiology reports only. Staged (Stg.) refers to performing weakly supervised training with EHR in a first training stage and then training with radiology reports in a second stage. This is in contrast to multi-task learning (MTL) where EHR and radiology reports are used for training simultaneously.}
\label{fig:zero_shot}
\end{figure}

Zero-shot findings classification assesses Merlin's ability to classify the presence of common imaging findings based on text prompts that a user can develop, which are likely distinct from prompts seen during training. In Figure \ref{fig:zero_shot}b we evaluate Merlin zero-shot classification across 30 abdominal CT findings on our internal and external clinical datasets. Merlin achieves an average F1 score of 0.741 (95\%CI [0.727-0.755]) on the internal validation dataset and an average F1 score of 0.647 (95\%CI [0.607-0.678]) on the external validation dataset, significantly outperforming off-the-shelf 2D OpenCLIP~\cite{cherti2022reproducible} with K=1 pooling and finetuned 2D BioMedCLIP~\cite{zhang2023biomedclip} with average pooling (p < 0.001) in both settings (Figure \ref{fig:zero_shot}b). Additional zero-shot ablation results for OpenCLIP and BiomedCLIP are presented in Supplemental Figure \ref{suppl_fig:top_k}. Qualitatively, we observe in Figure \ref{fig:zero_shot}c that Merlin's external performance remains high on diseases with coarse-grained or salient features, e.g. pleural effusion, splenomegaly, ascites, surgically absent gallbladder, prostatomegaly, anasarca, and abdominal aortic aneurysm. Performance expectedly decreases on more challenging findings that require more subtle and fine-grained features, e.g. appendicitis, metastatic disease, lymphadenopathy, and free air. We also plot Merlin performance without radiology report splitting (W/O splitting in Figure \ref{fig:zero_shot}c). Radiology report splitting refers to splitting the radiology reports into sections describing different anatomical structures for subsequent contrastive learning (e.g. \textit{"liver: normal"} and \textit{"vasculature: patent"}). Without radiology report splitting, Merlin achieves an average F1 score of 0.656 (95\%CI [0.640, 0.671]), similar to the Merlin external evaluation performance. On a separate external dataset (VerSe~\cite{loffler2020vertebral} vertebral fracture detection), Merlin achieves a zero-shot F1 of 0.767 (95\%CI [0.623-0.867]). In Figure \ref{fig:zero_shot}d, we establish quantitative relationships to assess how expanding the pretraining dataset would improve zero-shot classification. This analysis helps determine the extent of pretraining data necessary for obtaining a specified zero-shot classification performance. 

Compared to the ablations, Merlin (I3D initialization~\cite{carreira2017quo} where 2D ImageNet pretrained weights are reused within the 3D model, multi-task learning with EHR and radiology reports versus training in stages, and radiology report splitting) performs the best with an F1 score of 0.741 (95\%CI [0.727-0.755]) (Figure \ref{fig:zero_shot}e). Report splitting and staging training across the EHR and radiology reports results in second best performance. Learning directly from the radiology reports without any EHR supervision, along with radiology report splitting, results in the third best performance. These settings achieve F1 scores of 0.735 (95\%CI [0.719-0.748]) and 0.730 (95\%CI [0.714, 0.744]), respectively. The largest impact on performance is driven by the choice of whether or not to split the radiology reports. Relative to Merlin, without report splitting, the performance drops by an average of 7.9 F1 points (p << 0.01).

We further reformulated the zero-shot classification task into a supervised zero-shot findings classification task to compare zero-shot Merlin against supervised Merlin baselines using 10\% and 100\% pretraining data (Figure \ref{fig:baseline_experiments}d). In line with prior work from the original CLIP paper~\cite{radford2021clip}, zero-shot Merlin outperformed all supervised baselines in both data-scarce (10\% training data) and fully-labeled (100\% training data) supervised experiments, achieving a 45.0\% increase in F1 score with 10\% training data and a 29.0\% increase with 100\% training data (Figure \ref{fig:baseline_experiments}d). Specifically, zero-shot Merlin outperformed supervised Merlin in the 100\% training data setting, achieving a 16.0\% increase in F1 score (0.741; 95\% CI [0.727-0.755] versus 0.641; 95\% CI [0.574-0.701]). Additionally, Merlin trained with 100\% of the data achieved a 13.0\% increase in F1 score than Merlin trained on 10\% of the data, which had an F1 score of 0.577 (95\% CI [0.516-0.634]).

\subsection{Phenotype Classification}

The goal of this task is to use CT scans to directly predict phenotypes~\cite{denny2013systematic} (based on groupings of ICD codes) that were assigned to patients during their hospital admission that included the CT scan. We evaluate the performance of Merlin in predicting 692 phenotypes defined in PheWAS~\cite{denny2013systematic}. We find that Merlin reaches a macro-average area under the receiver operating characteristic curve (AUROC) across phenotypes of 0.812 (95\% CI, 0.808-0.816), and achieves AUROCs over 0.85 for 258 phenotypes (37\% of all phenotypes), and AUROCs over 0.9 for 102 phenotypes (15\% of all phenotypes). 

\begin{figure}[th!]
\centering
\includegraphics[width=1.0\textwidth]{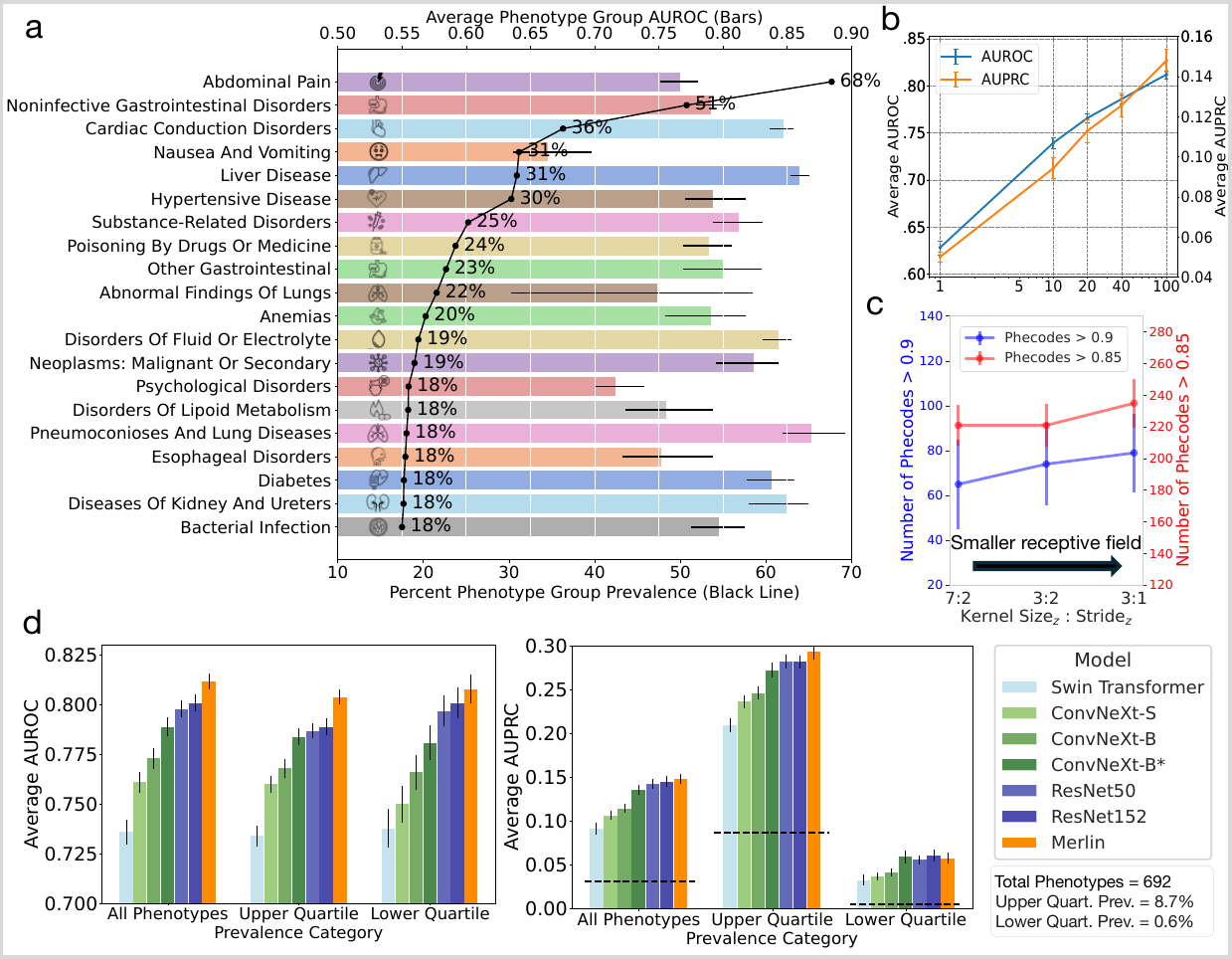}
\caption{\textit{Phenotype classification.} (a) Average AUROC performance for the top 20 phenotype groups listed in order of prevalence (black line). (b) Data scaling law experiments that measure how average AUROC and AUPRC scale across the 692 phenotypes as the amount of pretraining data varies. (c) Average AUROC as a function of model stem hyper-parameters. We find that a smaller receptive field yields better performance. (d) Average AUROC (left chart) and AUPRC (right chart) across all 692 phenotypes, the top quartile of 173 phenotypes, and the botton quartile of 173 phenotypes across several baseline models. All baseline models are trained using the phenotypes in the pretraining dataset. The dashed lines denote random chance performance. Note that Merlin, which uses the best performing backbone of ResNet152, is further trained using radiology reports.}
\label{fig:phecodes}
\end{figure}

In Figure \ref{fig:phecodes}a, we group the 692 phenotypes into groups of similar phenotypes using established methods (exclusion ranges defined in the PheWAS database~\cite{denny2013systematic}) and report prevalence and average performance within the top 20 most prevalent groups in the internal test set. We find that abdominal pain is the most prevalent phenotype group (68\% prevalence), followed by noninfective gastrointestinal disorders (51\% prevalence). Measuring average model performance across the phenotypes within each group, we find that Merlin performs well in detecting diseases across a range of organ systems, including the liver, kidneys, ureters, and gastrointestinal tract.

In Figure \ref{fig:phecodes}b we compute data scaling law curves to assess how Merlin performance improves with respect to training data. Similar to zero-shot classification, we find that increasing training data improves performance.

In Figure \ref{fig:phecodes}c and Figure \ref{fig:phecodes}d, we compare performance across 7 model variations, with different model backbones, architectures, and training paradigms. 
We make the following observations: i) model performance consistently improves across model size, ii) 
smaller convolutional receptive fields improve performance, with smaller z-dimension kernel sizes and strides in the model stem performing best (Figure \ref{fig:phecodes}d), iii) ResNet backbones, with in-plane and out-of-plane kernel size of 3, outperform ConvNeXt~\cite{liu2022convnet} backbones (in-plane kernel size of 7) that have out-of-plane kernel sizes of 3 (ConvNeXt-B*) or 7 (ConvNeXt-B) at all layers of the network (Figure \ref{fig:phecodes}c). The Swin Transformer~\cite{liu2021swin} backbone performs the worst, with a window size of 7 and patch size of 4. Due to the computational complexity of the attention mechanism, the number of parameters of the Swin Transformer is lowest (3 million). The trend that we observe across model variations as measured by average AUROC aligns with the trend described by average AUPRC.

\subsection{Zero-shot Cross-Modal Retrieval}
\begin{figure}[th!]
\centering
\includegraphics[width=1.0\textwidth]{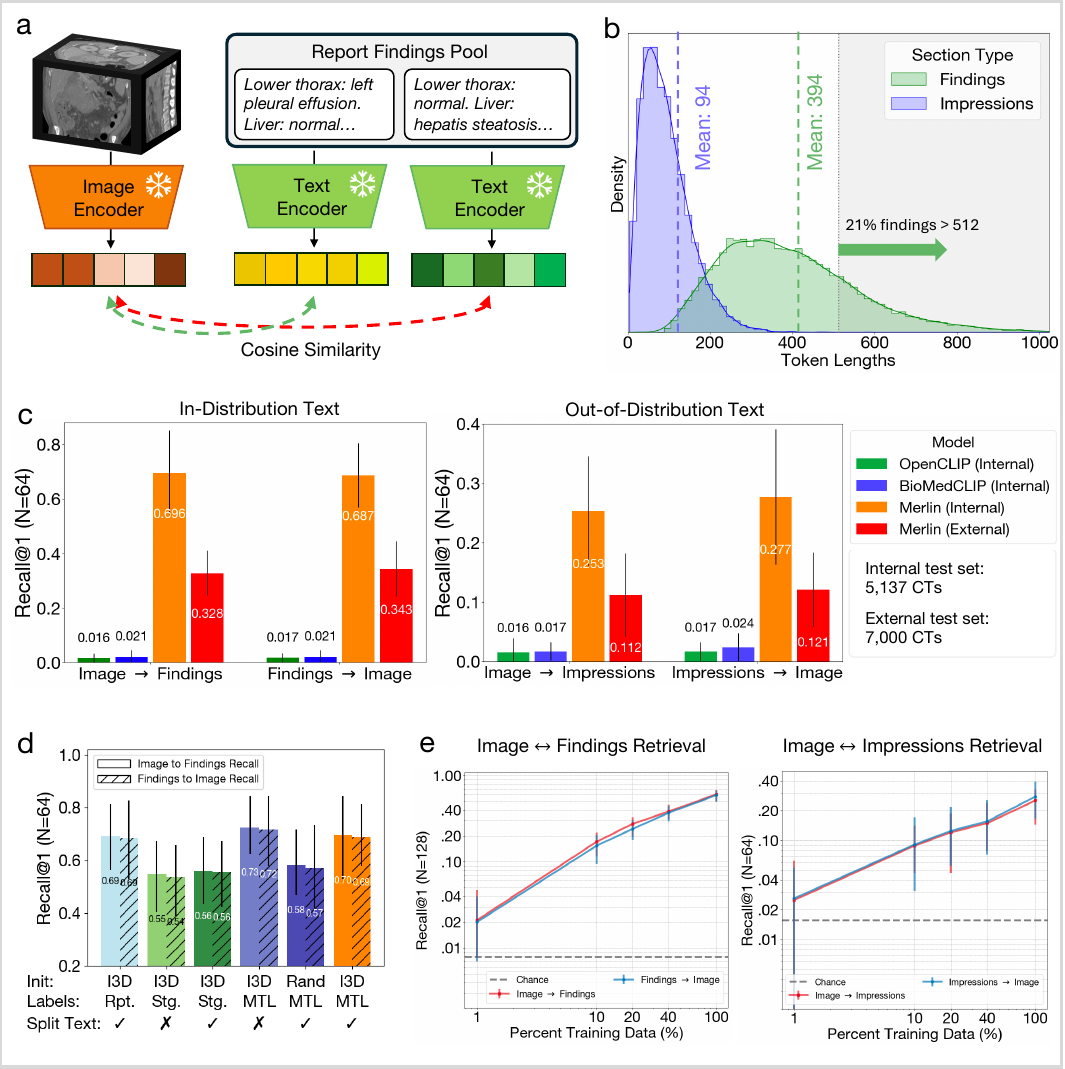}
\caption{\textit{Zero-shot cross-modal retrieval.} (a) Schematic demonstrating how we perform retrieval. We compute the cosine similarity between Merlin report embeddings and CT embeddings, enabling us to rank CT and report pairs in order of similarity. (b) A distribution of the findings section and impressions section lengths shows that 21\% of findings have sequence lengths greater than 512 tokens. (c) Top-1 recall out of pools of 64 findings sections (left), which is considered an in-distribution evaluation as Merlin is trained using findings sections. We also report top-1 recall on out-of-distribution impressions sections (right). (d) An ablation study that examines the impact of using I3D ImageNet initialization, multi-task learning (MTL) versus staged training (Stg.) with EHR and reports versus training with reports only (Rpt.), and splitting the radiology report text into anatomical sections. (e) Data scaling law experiments that examine the impact of pretraining dataset size on retrieval performance. The dashed lines indicate random chance performance.}
\label{fig:retrieval}
\end{figure}

Zero-shot cross-modal retrieval evaluates the model's ability to match a CT image with the corresponding radiology report findings or impressions section and vice versa. In Figure \ref{fig:retrieval}b, we plot the distribution of radiology report findings and impressions lengths. We find that 21\% of findings sections have lengths that exceed 512 tokens, motivating our choice of the clinical Longformer~\cite{li2022clinical} text encoder (OpenCLIP~\cite{cherti2022reproducible} and BioMedCLIP~\cite{zhang2023biomedclip} only allow maximum token lengths of 77 and 256, respectively). We find that Merlin significantly (p $<<$ 0.01) outperforms OpenCLIP and BioMedCLIP on the task of 3D retrieval for retrieving the correct findings out of 64 cases (Image $\rightarrow$ Findings in Figure \ref{fig:retrieval}c). We find similar performance for retrieving the correct CT scan out of 64, given a findings section (Findings $\rightarrow$ Image in Figure \ref{fig:retrieval}c). We also find that on even larger pools of data that Merlin outperforms existing baselines (Supplemental Figure \ref{fig:retrieval_extended}).

Retrieving the impressions section from radiology reports provides evidence that Merlin generalizes to out-of-distribution text, given that Merlin training uses the findings sections. Figure \ref{fig:retrieval}c demonstrates that Merlin generalizes to the task of retrieving the correct impressions section given a scan (Image $\rightarrow$ Impressions) and retrieving the correct scan given an impressions section (Impressions $\rightarrow$ Image). On the external test set, retrieval performances decreases (Figure \ref{fig:retrieval}c). However, it is important to note that the structure and language used in reports varies significantly across institutions. Nonetheless, the external Merlin performance remains 5-7x better than the external baselines.

We conduct a model ablation study (Figure \ref{fig:retrieval}d) to investigate i) the impact of ImageNet initialization with I3D weights, ii) the impact of either staged training or multi-task learning using the EHR and radiology reports, and iii) the impact of training with full reports or reports that are split across anatomical regions. We find that the top-performing model variation uses I3D initialization, multi-task learning with the EHR and radiology reports, and no report splitting. We hypothesize that report splitting does not improve performance on the retrieval task since retrieval prompting is performed with the full report sections. Splitting the report into subparts and additionally learning directly from the radiology reports without any EHR supervision results in the second and third best performance, respectively.

Finally, we find that model performance improves with pretraining dataset size both for image and findings retrieval and image and impressions retrieval via the data scaling curve in Figure \ref{fig:retrieval}e.

\subsection{Multi-Disease 5-Year Prediction}
\begin{figure}[t]
\centering
\includegraphics[width=1.0\textwidth]{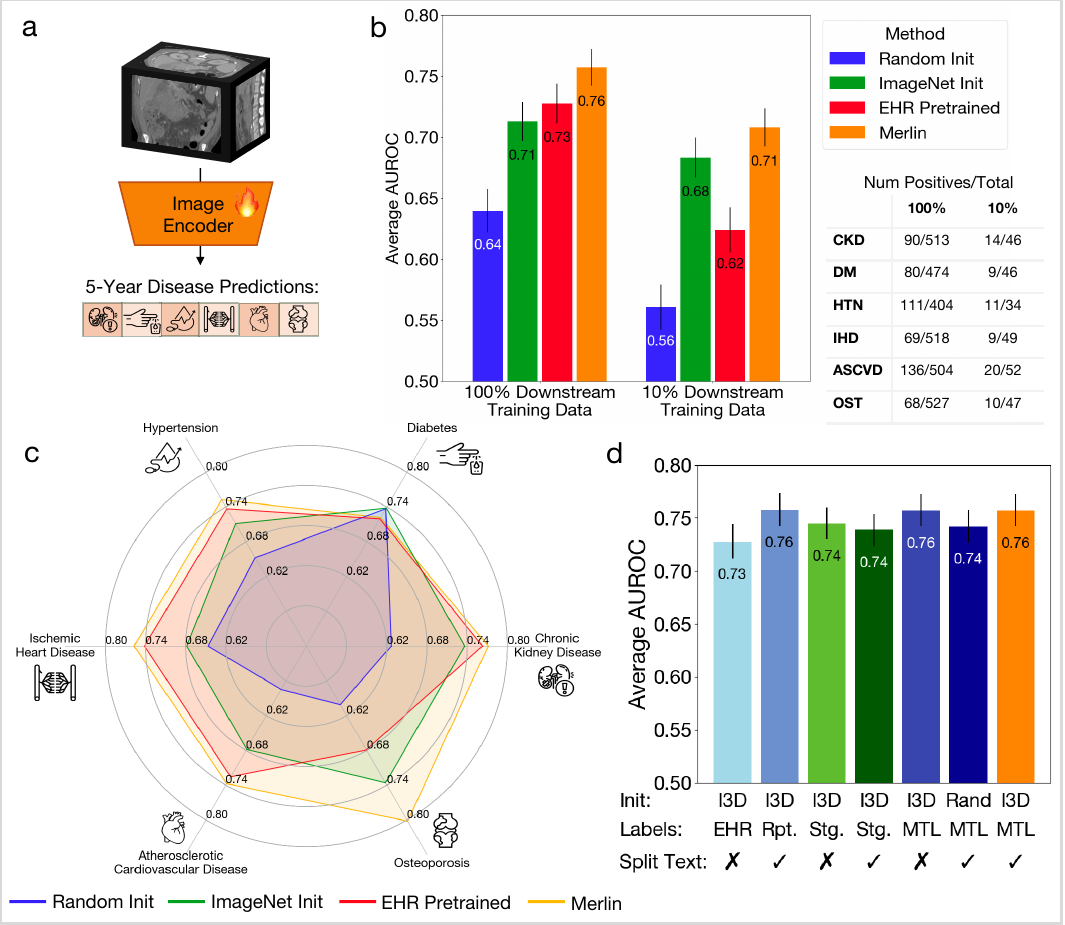}
\caption{\textit{Multi-disease 5-year prediction.} (a) We fine-tune Merlin for predicting chronic disease onset in otherwise healthy patients within 5-years. (b) We compare Merlin to other baseline model variations fine-tuned for the same task. We find that with both 100\% and 10\% of downstream training data, Merlin outperforms the other model variations. (c) Comparison of Merlin chronic disease prediction performance to a model trained using only phenotypes (EHR Pretraining), an ImageNet I3D initialized model, and a randomly initialized model. (d) An ablation study that measures the impact of various aspects of Merlin's training strategy. We find that training with EHR and radiology reports, using staged training (Stg.) or multi-task learning (MTL), and training with radiology reports only (Rpt.), all outperform training with EHR only.}
\label{fig:disease_prediction}
\end{figure}

Multi-disease 5-year prediction measures the model's ability to predict whether a patient will develop a chronic disease within 5 years, given that they are not diagnosed with the disease when their CT scan was acquired. We fine-tune Merlin to opportunistically predict which patients, who are healthy at baseline, will be diagnosed with any of 6 chronic diseases (chronic kidney disease, osteoporosis, cardiovascular disease, ischemic heart disease, hypertension, and diabetes) in the ensuing 5 years based on their CT scan. On average across these 6 diseases, Merlin predicts disease onset within 5 years with an AUROC of 0.757 (95\%CI [0.743, 0.772]) using 100\% of the downstream labels, and outperforms the ImageNet pretrained (I3D) image-only model by 7\% (Figure \ref{fig:disease_prediction}b). Even using 10\% of labels, Merlin predicts disease onset within 5 years with an AUROC of 0.708 (95\%CI [0.692, 0.723]), and outperforms the ImageNet pretrained model by 4.4\% (Figure \ref{fig:disease_prediction}b). Using Merlin for disease risk stratification can produce similar accuracy as using an ImageNet pretrained model, while utilizing 10x reduced labeled training data. These results depict that even fewer than 25 positive cases (10\% of training data) can be used to build future disease risk stratification models using Merlin. Our model ablation study demonstrates that 3 configurations, all of which use I3D initialization, provide similar performance. Report training only with report splitting (AUROC of 0.758 (95\%CI [0.743, 0.773])), multi-task EHR and report training without report splitting (AUROC of 0.757 (95\%CI [0.743, 0.772])), and our Merlin configuration with multi-task EHR and report training, along with report splitting, (AUROC of 0.757 (95\%CI [0.743, 0.772])) all produce comparable results.

\subsection{Radiology Report Generation}
\begin{figure}[th!]
\centering
\includegraphics[width=1.0\textwidth]{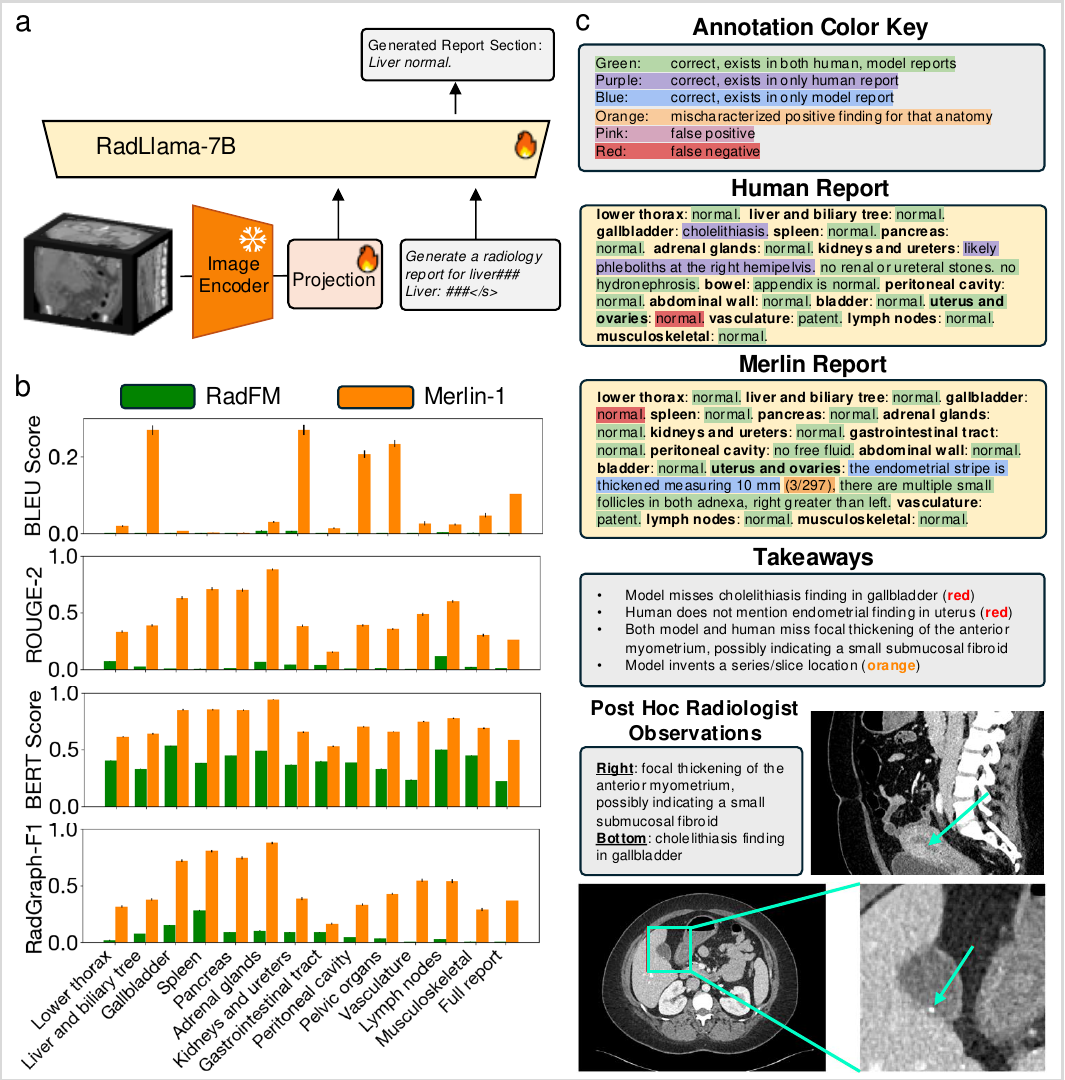}
\caption{\textit{Radiology report generation.} (a) To enable report generation, we extract the last hidden layer embeddings from Merlin and modify the dimension of these embeddings using a projection layer. We generate the report section by section and therefore also embed a report section prompt. The resulting vision and language tokens are used as input to a language model to generate a report section. (b) We compare the performance of our model against RadFM, using 4 metrics, across each report section and the full report. (c) We provide a densely annotated example of human and Merlin generated reports. We bold the report section headers in the human and Merlin generated reports. We include "uterus and ovaries" in green, as Merlin needs to deduce the correct pelvic anatomy.}
\label{fig:report_generation}
\end{figure}

Radiology report generation evaluates Merlin's capacity for generating reports based on the CT images. We select RadFM~\cite{wu2023generalist} as a baseline, as it is a multi-modal text generation model where the training dataset includes abdominal CT scans. Based on the quantitative metrics of RadGraph-F1~\cite{delbrouck2022improving}, BERT Score~\cite{zhang2020bertscore}, ROUGE-2~\cite{rouge}, and BLEU score~\cite{bleu}, Merlin consistently outperforms RadFM across anatomical sections and the full report findings (Figure \ref{fig:report_generation}b).

Qualitatively, we observe that Merlin generates reports with correct structure, where findings are placed within the correct anatomical section. When Merlin predicts the presence of a finding, the finding or a related finding usually exists in the image. For example, in the Merlin generated report in Figure \ref{fig:report_generation}c, Merlin correctly identifies that "the endometrial strip is thickened". Nonetheless, we observe that Merlin tends to under-report positive findings. For instance, in the example in Figure \ref{fig:report_generation}c, Merlin does not report the cholelithiasis finding, which is present in the human generated report and in the CT image. As this is an early demonstration of radiology report generation based on CT scans, we anticipate room for improvement in the generated reports.

\subsection{3D Semantic Segmentation}

\begin{figure}[th!]
\centering
\includegraphics[width=1.0\textwidth]{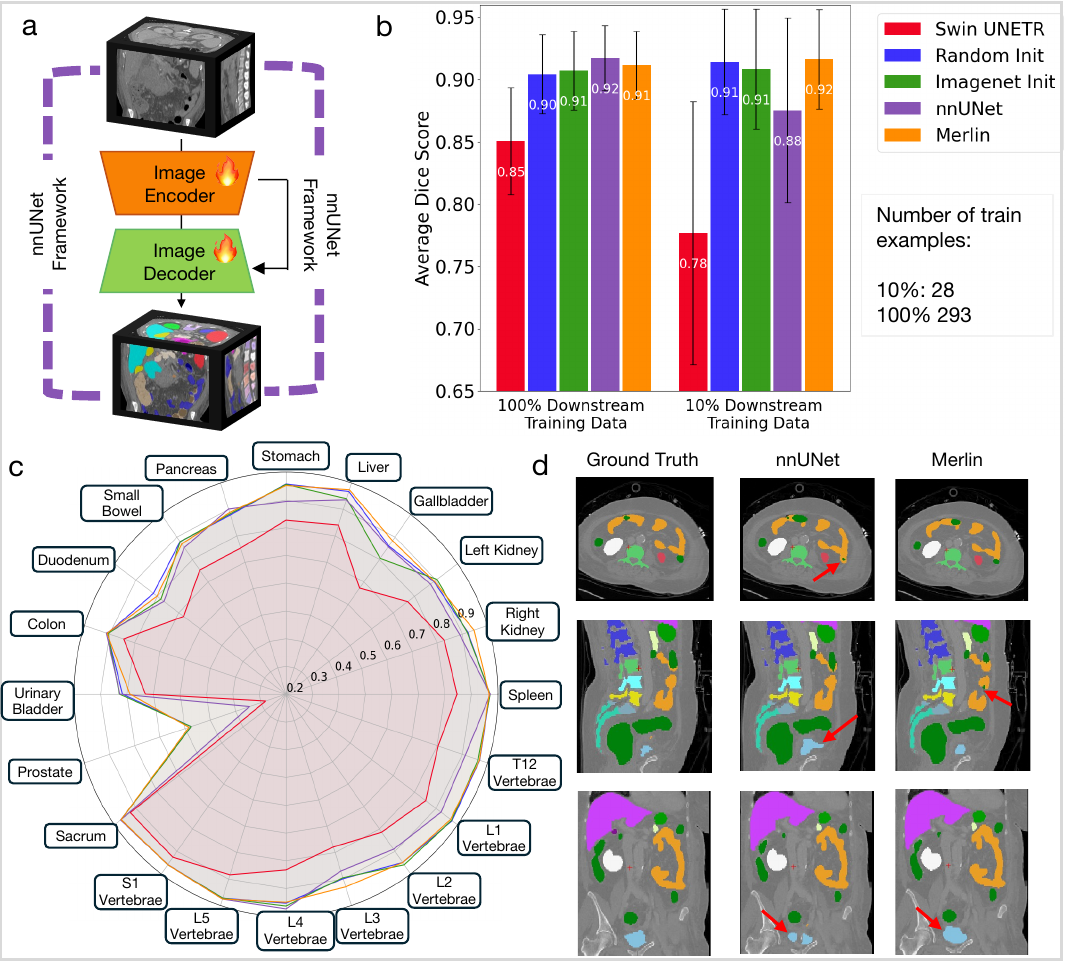}
\caption{\textit{3D semantic segmentation.} (a) To adapt Merlin and other architectures for segmentation, we add a decoder and skip connections between the encoder and decoder. We conduct all segmentation experiments within the nnUNet framework. (b) We compare model variations using average Dice score across 20 organs that appear in abdominal CT. We compare performance of models trained using 100\% of training cases and also simulate the data scarce regime with 10\% of training cases. (c) We report Dice scores for 20 organs across 5 model variations using 10\% of training cases. (d) We qualitatively compare segmentations between the ground truth labels, nnUNet, and Merlin with 10\% of training cases. The red arrows indicate inconsistencies made by the model relative to the ground truth. The same patient was sampled from the Total Segmentator~\cite{wasserthal2023totalsegmentator} test set.}
\label{fig:segmentation}
\end{figure}

3D semantic segmentation evaluates whether Merlin learns geometric information about various anatomical structures. We conduct all experiments within the nnUNet framework to ensure a fairer comparison between model architecture and weight initialization by standardizing data-preprocessing, training, and inference. We find that with 10\% of training cases, Merlin outperforms nnUNet by 4.7\% in average Dice score (Figure \ref{fig:segmentation}b). This demonstrates that Merlin pretraining is particularly beneficial in label scarce scenarios. However, the nnUNet model in the 100\% training data domain performs slightly better than the Merlin-initialized model, with a Dice score difference of approximately 0.006 (Figure \ref{fig:segmentation}b). Among the different initialization strategies for the Resnet152 encoder-decoder framework (ImageNet, random, and Merlin), the performance difference is around 0.01. However, we observe a much larger performance gap, a 6.3\% decrease in Dice score, between the fully convolutional segmentation models and the Swin UNETR architecture for this task. Overall, Merlin’s weight initialization outperforms other Resnet-152 initialization strategies in both the 10\% and 100\% training data settings and nnUNet in the 10\% training data setting across all 20 organs. 

Across 20 different organs in the test set, we compared the average Dice score of segmentation models (Figure \ref{fig:segmentation}c). On a per-organ basis, in the 10\% training data domain, Merlin achieves better performance than nnUNet on 12 out of 20 organs, with one organ showing an improvement of over 10\% and five organs having a 5\% improvement. Notably, Merlin demonstrates a 41.0\% improvement in Dice score on the prostate compared to nnUNet. Figure \ref{fig:segmentation}d presents qualitative segmentation results, where the red arrows indicate differences in the predicted segmentations relative to the ground truth. We observe that nnUNet makes mistakes in all three slices, while Merlin makes a mistake in two of the three slices. Top row, center column: nnUNet undersegments a small part of the descending colon. Middle row, center and right columns: nnUNet oversegments the urinary bladder and Merlin undersegments the inferior small bowel. Bottom row, center column: both nnUNet and Merlin improperly segment the urinary bladder. 

The slightly improved segmentation performance of nnUNet over Merlin in the 100\% domain may be attributed to two features. nnUNet’s 3D full-resolution architecture is shallower compared to the 152 layers of Merlin's Resnet152 (optimized for discriminative tasks), which may reduce potential overfitting. Furthermore, the hyperparameters and pre-processing in the nnUNet framework are specifically tuned for segmentation for a 3D full-resolution architecture, ensuring that the effective receptive field size is appropriately large for the patch size~\cite{isensee2021nnu}. This contrasts with the weight-initialized, inflated Resnet152 architecture used in Merlin, which outperforms nnUNet in the 10\% training data setting while maintaining state-of-the-art performance across numerous tasks.

\subsection{Alternative Architecture Experiments} \label{result:beyond_merlin} 

\begin{figure}[th!]
\centering
\includegraphics[width=0.9\textwidth]{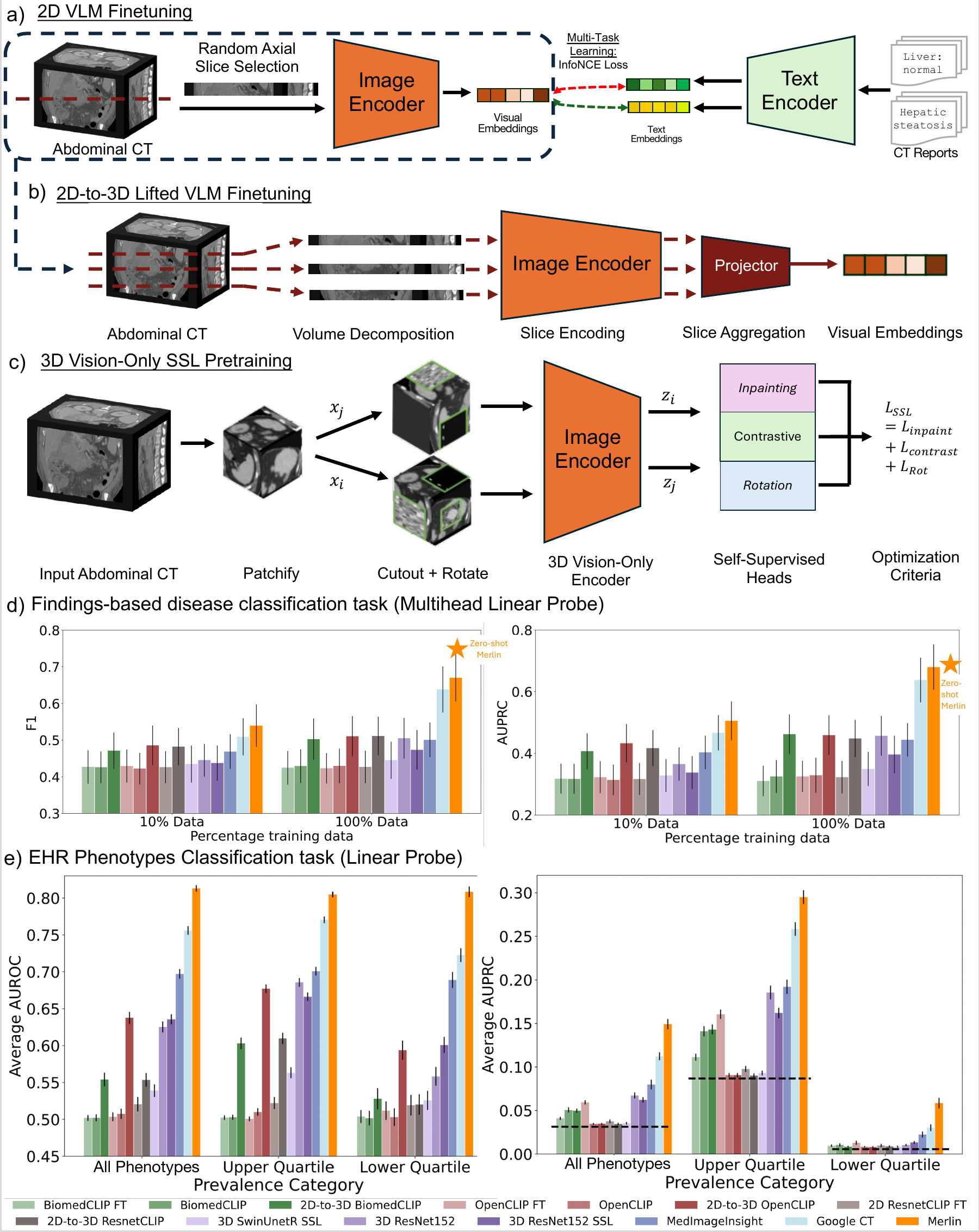}
\caption{\textit{Alternative architecture experiments.} The baselines, which include three 2D-to-3D lifted vision-language models, five 2D vision-language models, and two 3D vision-only models, are presented to evaluate different training strategies for the AbCT dataset. The baselines serve as points of comparison against Merlin. Below, we illustrate the training procedure for the baseline models. (a) 2D vision-language model (VLM) finetuning engages both the image and text encoders. Specifically, we adapted the image encoder to process randomly selected axial slices from abdominal CT scans and trained the model in a multi-task learning framework, jointly leveraging EHR phenotypes and radiology reports. (b) We finetuned the 2D-to-3D lifted VLM by decomposing the abdominal CT volume into individual slices, which were processed through the image encoder. A neural network slice projector then combined the slice-wise embeddings into a single embedding. (c) Vision-only SSL pretraining using SwinUNETR and Resnet152 entails patchifying a large 3D volume into 3D patches and applying patch-level augmentations, with inpainting, contrastive, and rotational losses. (d) Average F1 (left chart) and AUPRC (right chart) on 10\% and 100\% pretraining data for the findings-based disease classification task. (e) Average AUROC (left chart) and AUPRC (right chart) across all 1692 phenotypes, the top quartile of 173 phenotypes, and the bottom quartile of 173 phenotypes across several baseline models. All baseline models are trained using the phenotypes in the 100\% pretraining dataset. The dashed lines denote random chance performance. Note that Merlin is the best-performing model between these two tasks.}
\label{fig:baseline_experiments}
\end{figure}

Our results from linear probe experiments demonstrate the advantages of Merlin's 3D vision-language pretraining strategy over 2D finetuned VLMs, 2D-to-3D lifted VLMs, 3D vision-only models, Microsoft's MedImageInsight model~\cite{codella2024medimageinsight}, and Google CT foundation model~\cite{yang2024advancing} (Figure \ref{fig:baseline_experiments}). We further show advantages of Merlin pretraining in the data-scarce (10\% training data setting) on the findings-based disease classification task (Figure \ref{fig:baseline_experiments}d) and on the EHR phenotypes task (Supplemental Figure \ref{fig:phecode_classification_extended}b). These performance gains were also observed in full finetuning evaluations documented in Supplementary Section \ref{suppl:embedding_comparison}. Specifically, through our experiments, we observe that: (i) vision-language pretraining outperforms vision-only pretraining, (ii) finetuning improves the performance of 2D and 2D-to-3D VLMs though they underperform relative to Merlin, (iii) Merlin outperforms recent CT embedding foundation models (Figure \ref{fig:baseline_experiments}).

\noindent\textbf{2D VLMs:} The 2D VLMs demonstrated relatively consistent performance between both finetuning and no finetuning scenarios on the linear probe classification tasks, with finetuned 2D ResnetCLIP performing comparably to the other finetuned models. For example, when trained on 100\% of the data for the findings-based disease classification task, ResnetCLIP achieved an F1 score of 0.427 (95\% CI [0.381-0.470]), which was similar to OpenCLIP (0.423, 95\% CI [0.381-0.463]) and BiomedCLIP (0.424, 95\% CI [0.378-0.0.470]). However, 2D VLMs underperformed compared to Merlin, with a 54.7\% lower F1 score for findings-based disease classification task and 59.3\% lower AUROC on the EHR phenotypes task in the 100\% training data setting (Figure \ref{fig:baseline_experiments}d-e). 

\noindent\textbf{2D-to-3D lifted VLMs:} 2D-to-3D lifted VLMs showed consistent performance across tasks and demonstrated performance improvements compared to their 2D VLM counterparts. On the findings-based disease classification task, they achieved a 19.7\% increase in F1 score and a 42.9\% increase in AUPRC compared to finetuned 2D VLMs in the 100\% training data setting (Figure \ref{fig:baseline_experiments}d). Similarly, on the EHR phenotype task, 2D-to-3D lifted VLMs outperformed their 2D VLMs baselines, with a 14.2\% increase in AUROC and a 47.0\% increase in AUPRC (Figure \ref{fig:baseline_experiments}e). Despite these improvements, 2D-to-3D lifted VLMs underperformed relative to Merlin. Specifically, Merlin had 32.1\% higher F1 scores on the findings-based disease classification task and a 39.8\% higher AUROC on the EHR phenotype classification task compared to the 2D-to-3D lifted VLMs (Figure \ref{fig:baseline_experiments}d-e). 

\noindent\textbf{3D vision-only models:} 3D models trained from scratch or with SSL pretraining had a 41.1\% decrease in F1 score on the findings-based disease classification task and a 35.8\% decrease in AUROC on the EHR phenotypes task compared to Merlin in the 100\% training data setting  (Figure \ref{fig:baseline_experiments}d-e). Among the 3D vision-only baselines, SwinUNETR trailed 3D Resnet SSL, with a 6.3\% decrease in F1 score on the findings-based disease classification task and a 18.0\% decrease in AUROC on the EHR phenotypes task. The two 3D Resnet models underperformed relative to Merlin, with a 36.9\% decrease in F1 score on the findings-based disease classification task and a 28.6\% decrease in AUPRC on the EHR phenotypes task (Figure \ref{fig:baseline_experiments}d-e). Although 3D vision-only models outperformed the 2D and 2D-to-3D VLMs with a 6.02\% increase in F1 score on the findings-based disease classification task and 12.0\% increase in AUROC on the EHR phenotypes task, Merlin still achieved the best quantitative performance across all tasks (Figure \ref{fig:baseline_experiments}d-e).

\noindent\textbf{CT Embedding Foundation Models:} Although MedImageinsight showed a 52.5\% increase in AUROC from the 10\% to 100\% training data setting on the EHR phenotype task (Supplemental figure \ref{fig:phecode_classification_extended}), it still underperformed relative to Merlin with a 33.8\% lower F1 on the findings-based disease classification task and 16.6\% lower AUROC on the EHR phenotype classification task (Figure \ref{fig:baseline_experiments}d-e). Merlin also outperformed the 3D Google CT foundation model, achieving a 5.01\% higher F1 and 7.68\% higher AUROC on the findings-based disease and EHR phenotype tasks (Figure \ref{fig:baseline_experiments}d-e). We therefore find that all alternative architecture baselines underperform relative to Merlin across all label settings and tasks.

\subsection{External Validation Experiments} \label{result:external_val}

\begin{figure}[th!]
\centering
\includegraphics[width=1.0\textwidth]{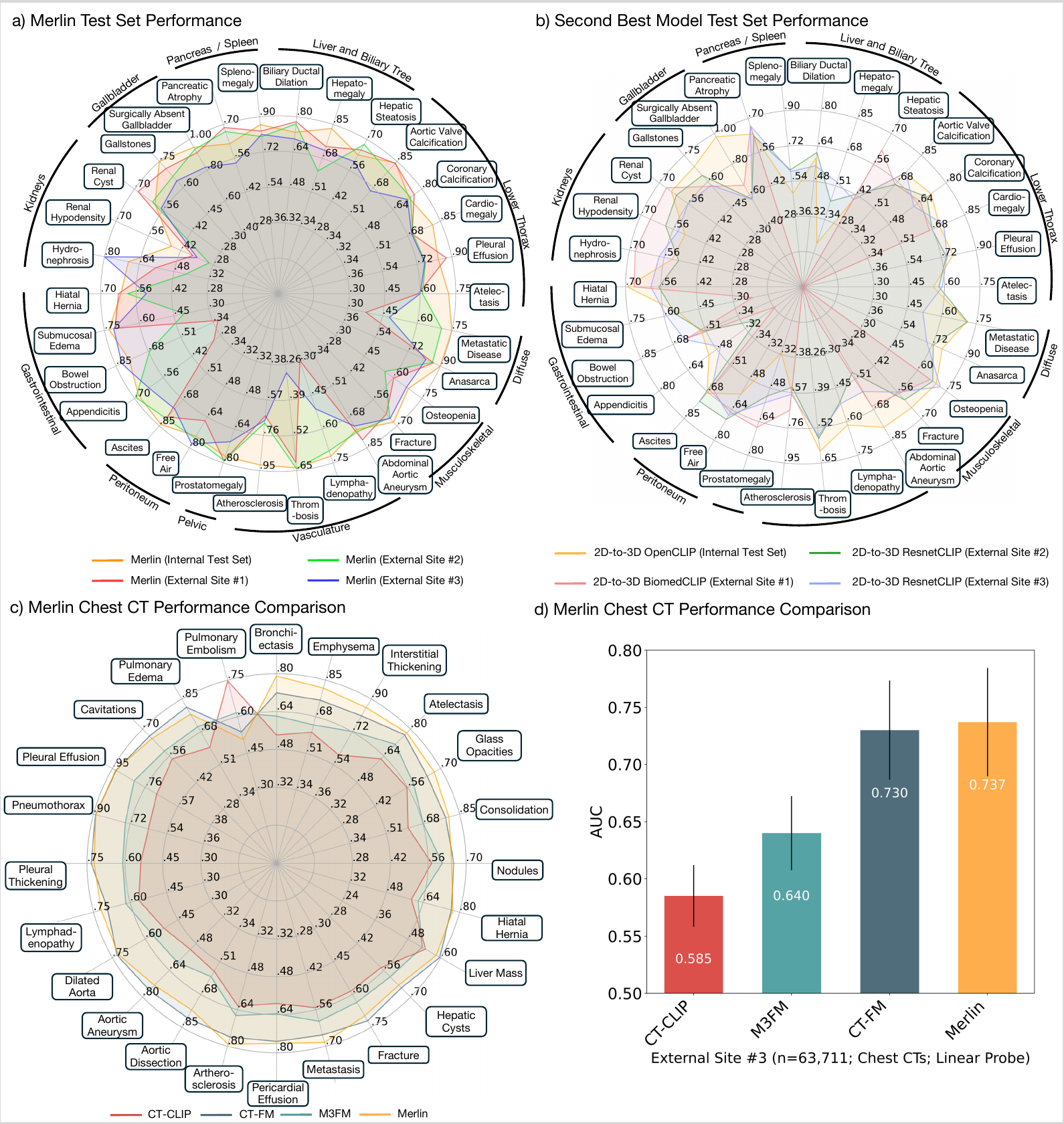}
\caption{\textit{External validation on abdominal and chest CTs.} Merlin’s strong performance across 44,098 abdominal and chest CTs from external sites. (a) Merlin's performance across internal and three external sites across 30 findings assessed on abdominal CT scans, where Merlin achieved the highest performance across all datasets. (b) Performance breakdown of the second-best model at each site (selected by second-highest average F1) on the same abdominal CT findings. (c)  We train a series of embedding models on External Site \#3’s chest CT dataset and benchmark them against recent chest CT foundation models. We here present AUC performance comparison for each model across 24 diseases on External Site \#3’s dataset. (d) Average AUC performance of Merlin compared to recent Chest CT FMs on External Site \#3’s dataset. Across all baselines, Merlin consistently achieves the highest performance. We note that Merlin was trained exclusively on abdominal CT scans, while all other comparison models were trained explicitly on chest CT scans. Despite this training anatomy discordance, Merlin exhibits the strongest performance on Chest CT scans out of all comparison models.}
\label{fig:external_val}
\end{figure}

We draw from a dataset of over 100,000 external CT scans to evaluate Merlin on a total of 44,098 external CT scans, comprising 37,855 abdominal CTs from three external sites and 6,243 chest CTs from site \#3. Merlin demonstrates consistent and accurate performance across sites and anatomies, overcoming distributional shifts between training and external testing sets (Supplemental Table ~\ref{tab:dataset_characteristics}), and consistently outperforms alternative architecture baselines and chest CT foundation models (Figure ~\ref{fig:external_val}).

\noindent\textbf{Zero-shot findings classification:} We evaluated Merlin on the 30 abdominal findings part of our zero-shot evaluation framework, \textit{without any finetuning}, using abdominal CT scans for our three external datasets (Figure \ref{fig:external_val}) to assess the robustness of its embeddings to out-of-distribution data sources. For comparison, we include alternative architectures 2D/2D-to-3D BiomedCLIP, OpenCLIP, and ResnetCLIP. Across all sites, Merlin outperformed the second-best alternative architecture baseline by 19.7\% on average F1 and demonstrated consistent disease-specific performance across external cohorts (Figure ~\ref{fig:external_val}a-b). Specifically, at out-of-state site \#1, Merlin exceeded 2D-to-3D BiomedCLIP by 34.4\%; at in-state site \#2, it outperformed 2D-to-3D ResnetCLIP by 15.7\%; and at in-state site \#3, it maintained an 8.9\% advantage over 2D-to-3D ResnetCLIP. These differences largely reflect distribution shifts in patient age, slice thickness, scanner manufacturer, and radiologist reporting patterns (Supplemental Table ~\ref{tab:dataset_characteristics}). We attribute Merlin's improved performance here to its full 3D image encoder pretraining. 

\noindent\textbf{Chest CT linear probe evaluation.} We evaluate Merlin against recent chest CT FMs~\cite{hamamci2024developing, niu2025medical, pai2025vision} and find that, even when training \textit{only} a linear probe on a frozen Merlin image encoder, Merlin outperforms chest CT FMs by 12.3\% on average AUC (Figure \ref{fig:external_val}c-d). Given that Merlin, which was exclusively trained on abdominal CT scans, outperforms other vision-language models trained explicitly on chest CT scans, this demonstrates the generalizability of the Merlin model. Specifically, Merlin surpasses CT-CLIP by 24.7\%, M3FM by 14.0\%, and achieves equivalent performance to CT-FM. We attribute Merlin’s strong performance on this task to this combination of full-volume 3D processing and diverse multi-modal pretraining.

\section{Discussion}
\label{sec:discussion}

In this study, we curate a high-quality clinical dataset comprising of paired CT scans (10,628,509 total 2D images from 25,528 CT scans), EHR data (2,995,293 ICD codes), and radiology reports (10,051,571 total tokens in the findings sections). With this dataset, we train Merlin, a 3D vision-language foundation model for interpreting abdominal CT scans that leverages structured EHR data and unstructured radiology report supervision. Merlin is trained on a single GPU, demonstrating that compute-constrained hospitals and research institutions can build their own models, as well as highlighting the opportunity to train larger models with additional compute and data. We demonstrate that Merlin can generalize to 6 types of downstream tasks with 752 individual tasks. Non-adapted tasks include zero-shot findings classification (31 classes), phenotype classification (692 classes), and zero-shot cross-modal retrieval (including both image to findings and image to impressions retrieval). Model adapted tasks include 5-year disease prediction (6 classes), radiology report generation, and 3D semantic segmentation (20 organs). We outperform carefully chosen single-task baseline models across all tasks. We further contextualize our methods with comprehensive ablation studies that demonstrate Merlin's performance as a function of the underlying model architecture and the design considerations for the use of the underlying data. For training our Merlin foundation model, we demonstrate the benefit of I3D weight initialization, multi-task learning with EHR and reports, and splitting the radiology reports into anatomical sections. Overall, we depict how Merlin may assist in interpretation of abdominal CT scans and mitigate the burden on radiologists, while simultaneously adding value for new biomarker discovery and future disease risk stratification using pre-existing datasets.

Unlike traditional radiology AI tools that largely focused on bespoke classification tasks~\cite{mello2023clinical, alis2024choosing}, Merlin offers a suite of capabilities to support clinicians across various diagnostic, prognostic, and research tasks (Figure \ref{fig:overview_figure}h). When a patient undergoes an abdominal CT scan prescribed by their provider, Merlin can streamline the clinical workflow. As the radiologist reviews the CT scan, Merlin can generate a preliminary draft of the radiology report and provide a list of common findings, aiding with interpretation efficiency. Additionally, Merlin’s multimodal retrieval capabilities enable content-based image retrieval, allowing finding similar prior studies. This ability can facilitate improved clinical decision-making by providing radiologists and referring providers with relevant patient cases for comparison. Beyond diagnostic tasks, Merlin can assist with opportunistic analysis of the image for diagnosing current or future disorders using its ability of organ segmentation and future disease prediction. Using the abdominal CT image, Merlin can also identify the appropriate ICD-9 and ICD-10 codes, allowing for appropriate diagnostic coding for patients. This automation may reduce the risk of billing errors and claim denials, which can arise from incorrect coding~\cite{kusnoor2020narrative, weiner2018point}.

Prior approaches for 3D CT pretraining focus on image-only pretraining~\cite{huang2023self,tang2022self,valanarasu2023disruptive}. These methods leverage masked autoencoders (MAEs)~\cite{he2021masked}, which train a model to reconstruct the input image, given that some part of the input is masked out. This approach has been demonstrated to be beneficial for 3D segmentation tasks, where the decoder network is included in pretraining. However, the quality of the latent space representations for a variety of downstream tasks on CT images has not been demonstrated. It is possible that while MAEs require the latent space to encode the geometric information necessary to reconstruct the input, this geometric information reduces the effectiveness of the latent representations for image classification tasks. For image classification, it may be beneficial to discard this geometric information. In contrast to MAEs, Merlin trains the latent representations directly using supervision from EHR diagnosis codes and radiology reports. We demonstrate that these representations are useful for a variety of downstream task types, including tasks involving text and images, which cannot be accomplished using image-only pretraining methods.

Recent concurrent research has started to investigate 3D vision language models for radiology, particularly focusing on chest and head CT scans~\cite{hamamci2024foundation, yang2024advancing}. However, the extent of experiments and 3D evaluations are limited. These studies lack systematic ablation studies and do not utilize all available clinical data, including EHR. In addition to exploring the use of multiple clinical data types for supervision, we carry out a wide array of experiments and evaluations that contextualize the methodological choices we make in the development of Merlin.

Through the ablation studies, we find that I3D weight initialization~\cite{carreira2017quo} is helpful for all tasks that we examine with ablation studies. Furthermore, training with both EHR diagnoses and radiology reports is beneficial over training with either alone. The manner of combining these supervision sources is important, with multi-task training using EHR and radiology reports outperforming staged training. Nonetheless, report-only contrastive pretraining accounts for the vast majority of the gains, and the modest additional improvement from incorporating EHR diagnoses may not justify the added complexity in settings where such data are difficult to obtain. Additionally, we find that report splitting, while essential for zero-shot classification, slightly degrades zero-shot retrieval performance. This reinforces that pretraining is most effective when the pretraining data distribution matches the downstream task distribution. Splitting the radiology reports results in text that better aligns with prompts for zero-shot classification, but less closely resembles the radiology report text used for retrieval tasks.

We also demonstrate the advantages of Merlin’s 3D vision-language pretraining strategy over finetuned 2D VLMs, 2D-to-3D lifted VLMs, 3D vision-only models, and recent CT embedding foundation models\cite{codella2024medimageinsight, yang2024advancing}. Specifically, we find that (i) vision-language pretraining outperforms vision-only pretraining, (ii) finetuning improves the performance of 2D and 2D-to-3D VLMs though they underperform relative to Merlin, (iii) Merlin outperforms recent CT embedding foundation models, and (iv) Merlin outperforms other baselines in both data-scarce and fully supervised settings. Merlin outperformed 3D vision-only SSL and non-SSL methods, suggesting that language alignment provides a stronger self-supervision signal during CLIP pretraining compared to 3D vision-only SSL approaches. Additionally, 2D VLMs exhibited limited capacity to capture volumetric representations required for evaluation tasks, presumably due to their inability to process full 3D volumes. Similarly, 2D-to-3D lifted models failed to effectively encode volumetric information, as their projection layers insufficiently compressed multi-slice data relative to Merlin’s full-volume processing. Furthermore, in the supervised findings-based disease classification task, zero-shot Merlin still outperformed linear probing, mirroring the results observed in the original CLIP experiments~\cite{radford2021clip}, likely due to its ability to leverage the pretrained alignment between vision and language modalities. This alignment enabled zero-shot inference to generalize well across tasks, even without task-specific finetuning. Overall, our findings suggest that 3D VLM pretraining through Merlin demonstrates trends consistent with other foundation model studies~\cite{wornow2023ehrshot}.

Beyond outperforming alternative architecture baselines, Merlin demonstrates strong generalizability across diverse external datasets. Merlin maintained consistent and strong performance when evaluated on over 44,000 external CT scans spanning multiple sites and anatomies. These results highlight the model’s ability to handle distributional shifts between training and testing cohorts, including differences in patient demographics, acquisition parameters, and reporting practices. Importantly, these findings suggest that Merlin’s pretraining captures features that can generalize across sites, providing a foundational backbone for CT foundation model development and applications.

In light of our findings, we identify key areas that could significantly enhance the performance of Merlin and other future 3D medical imaging models:
\begin{enumerate}
    \item \textit{Increasing dataset size:} Expectedly, our initial data scaling results show that using a larger pretraining dataset improves Merlin's performance (Figure \ref{fig:zero_shot}d, Figure \ref{fig:phecodes}b, Figure \ref{fig:retrieval}e). Our data scaling curves across tasks can serve as a useful prospective reference measure in evaluating adequate sizes of training datasets for a requisite task performance. We note that as we add more data, we may need to update our model architecture, which was optimized at our current data scale.
    \item \textit{Improving image resolution:} Utilizing higher resolution images is expected to enhance model performance, as suggested by prior studies on vision-language models~\cite{laurenccon2024matters,li2023monkey,radford2021learning}. This may only hold until the resolution used for training is equal to the resolution of the original CT scan, and we are no longer downsampling the CT scan. It is important to note that increasing the physical resolution during image acquisition may not always yield better results. Radiologists often do not use the highest resolution available during acquisition in abdominal CT scans, particularly regarding slice thickness, since this may come at the expense of lower signal-to-noise ratio (SNR).
    \item \textit{Optimizing batch size:} Improving batch size is a challenge with 3D medical imaging where the size of the images is significant. However, prior work demonstrates that large batch sizes are beneficial for contrastive pretraining~\cite{radford2021learning, chen2020simple}.
    \item \textit{Extending to additional anatomies and modalities:} Our Merlin model lays the groundwork for training anatomy- and modality-specific radiology foundation models. Future work can explore the relative benefits of pretraining on multiple anatomies for the same modality, multiple modalities for the same anatomy, or both. For such experiments, it is imperative to benchmark the benefit that each modality or anatomy provides, and we hope that our rigorous evaluation strategy can help highlight optimal data mixtures for training next-generation radiology foundation models.
\end{enumerate}

Beyond the strengths of our study, there are also limitations of our model and our work. First, towards our goal of allowing compute-constrained health systems to train and finetune foundation models, we train and evaluate Merlin using a single GPU. This may limit the generalizability of some of our findings as task performance may increase by simply scaling up compute resources. Second, the limited number of publicly-available baselines for abdominal CT, especially VLM baselines, makes it challenging to understand the relative efficacy of our methods. On the other hand, we believe that Merlin could establish a strong and task-agnostic initial baseline for abdominal CT. Third, further exploration is required to optimize Merlin for the tasks of radiology report generation and 3D segmentation. For radiology report generation, there are numerous parameters that can be tuned, primarily regarding the adapter and LLM. While we devise an initial performant strategy for direct report generation, we leave further optimization of the adapter and LLM to future work. Likewise, adapting Merlin for segmentation requires adding a decoder network. Future work can explore additional architectural parameters and optimization strategies for decoder training.
\section{Methods}

\subsection{Datasets: Paired CT Scan, Unstructured Radiology Report, and Structured EHR}
\label{sec:data}
Medical imaging presents the opportunity to capture supervision signal from both the structured information in the EHR, as well as the unstructured information in radiology reports. We focus on abdominal CT images, as abdominal CTs are the most common 3D imaging examination~\cite{doi:10.1148/radiol.2017161911}. 

All datasets used in this study were under IRB approval with a waiver of informed consent due to the use of retrospective data. To collect the dataset, we identified patients from our academic medical center who underwent consecutive abdominal CT examinations from December 2012 to October 2018. This resulted in a high quality clinical dataset comprising 18,321 patients, from inpatient (37\%), emergency services (35\%), outpatient (16\%), and observation (8\%). We collected the following data for each patient:

\noindent \textbf{CT Studies:} We obtain full abdominal CT studies, each comprising multiple CT series. From each study, we select the series with the most axial slices, maximizing the amount of information in the CT volume. This results in 10,628,509 2D images from 25,528 CTs. This sampling may include the non-contrast series and put the image embeddings at a disadvantage relative to the radiology reports since contrast enhancement patterns may not be apparent on the non-contrast series. To remedy this, we run an open source abdominal CT contrast-phase detection algorithm that has been validated previously~\cite{reis2024automated} on the selected CT series. We use this algorithm as the image meta-data related to contrast phase is often missing or inaccurate. We find that 97\% of the selected series are portal venous phase, while 2.4\%, 0.45\%, and 0.26\% scans are delayed, arterial, and non-contrast phases, respectively. \\
\noindent \textbf{Radiology Reports:} We compile the associated radiology reports for each CT study. These reports consist of multiple sections, most predominantly the findings and impressions sections. The findings section includes detailed observations from each organ system. The impressions section contains a description of the most clinically important findings. We use only the findings sections for training, given the granularity of information provided and previous work demonstrating the efficacy of this approach~\cite{van2023exploring}. We find that there are 10,051,571 tokens from the radiology report findings sections in our dataset. \\
\noindent \textbf{EHR:} Beyond the images and reports, we acquire EHR data for each patient. In this work, we leverage the diagnosis information, in the form of ICD codes, for model training. To link ICD codes with CT scans, we collect the ICD codes that were assigned during the patient encounter that generated the CT. There are 954,013 assigned ICD9 codes with 5,686 unique ICD9 code values in our dataset. There are 2,041,280 ICD10 codes with 10,867 unique ICD10 code values. In total, there are 2,995,293 assigned codes in the dataset with 16,553 unique values.

\subsection{Vision-Language-EHR Model}
\label{sec:model}

We aim to leverage both structured EHR and unstructured radiology report information as supervision signal to train a CT visual model. 

\noindent \textbf{CT Scan Preprocessing:} We reformat all CT scans so that the first axis points from left to right, the second from posterior to anterior, and the third from inferior to superior. We then resample the in-plane axial images to 1.5mm resolution and the out-of-plane slice thickness to 3mm spacing using bilinear interpolation. We map the Hounsfield unit range -1000:1000 to the range 0:1, clipping values that fall outside of this range. Finally, we pad and center crop to 224 by 224 pixels in-plane and 160 pixels out-of-plane.

\noindent \textbf{EHR Preprocessing:} From the EHR, we extract all ICD-9 and ICD-10 codes that are assigned during the hospital visits where the abdominal CT studies are carried out. Furthermore, we observe that often codes can comprise disparate characters but have similar underlying phenotypic meaning. For example, the ICD-9 code V12.55 denotes a personal history of pulmonary embolism, where ICD-9 code 415 represents acute pulmonary heart disease. While one of these codes describes a history and one describes an acute indication, their imaging phenotypes may appear similar. Thus, we leverage the PheWAS Phecode Mapping~\cite{denny2013systematic} to map 16,553 ICD-9 and ICD-10 codes to 1,692 hierarchical phenotypes. Furthermore, we apply phenotype expansion where if a subject is positive for a more specific phenotype during a particular hospital visit, we propagate this positive label throughout the hierarchical phenotype tree so they are also positive for all less specific phenotypes. Grouping the ICD codes in our dataset results in 1,692 total phenotypes. For each CT image, we thus have an associated binary vector with a 0 indicating the absence of a phenotype during the corresponding hospital visit and a 1 indicating the presence of the phenotype. This supervision signal is coarse grained in that the phenotypes may not be directly associated with pixel values in the abdominal CT scan; they are associated with the patient's health status more generally. Thus, the EHR phenotype codes serve as weak supervision for our model training.

\noindent \textbf{Radiology Report Preprocessing:} Using regular expressions, we extract the findings section from each radiology report. Due to the long radiology reports, we hypothesize that contrastive training may overfit to short and salient parts of the reports to solve the task. Furthermore, short prompts used for subsequent zero-shot classification would present a significant domain shift compared to the full reports. Thus, we split the reports into anatomical sections and during each training step, alternate between presenting the full reports and a single anatomical section. We rotate through the anatomical sections every other step, presenting the model a single anatomy per batch to allow the model to compare across different descriptions of the same anatomy. To generate these sections, we use regular expressions. If the regular expressions fail for a particular report and anatomy, we use the full report. The sections that we consider are lower thorax, liver and biliary tree, gallbladder, spleen, pancreas, adrenal glands, kidneys and ureters, gastrointestinal tract, peritoneal cavity, pelvic organs, vasculature and lymph nodes, and musculoskeletal.

\noindent \textbf{Model Architecture:}
\label{Model Architecture}
Merlin uses an inflated 3D (I3D) ResNet152 for the image encoder. Inflation refers to reusing 2D pretrained model weights and copying those weights across the 3rd dimension of the 3D convolutional kernels~\cite{carreira2017quo}. Given the long token lengths of the reports (Figure~\ref{fig:retrieval}b), we use clinical Longformer~\cite{li2022clinical} as the text encoder due to its longer context length (4,096) than other biomedical pretrained masked language models and general domain CLIP text encoders. Previous work found that pretrained text encoders with longer context length perform better, given longer captions~\cite{zhang2023large}. We also perform architecture ablation studies where we investigate 3D Swin Transformer~\cite{liu2021swin} and ConvNeXt~\cite{liu2022convnet} architectures (Figure~\ref{fig:phecodes}c). In addition to architecture ablations, we investigate how the out-of-plane stride and kernel size in the model stem of the ResNet152 impact performance (Figure~\ref{fig:phecodes}d). The model stem in the ResNet152 is a convolutional layer at the input of the network with an in-plane kernel size of 7 and stride of 2, followed by a max pooling layer with a kernel size of 3 and a stride of 2. The model stem thus reduces size of the input by a factor of 4 in each dimension. The model stem ablation only adjusts parameters in the convolutional layer of the stem. We maintain the in-plane settings as they are in the 2D version of the models to leverage the 2D pretraining. Furthermore, we assess how inflating the ConvNeXt model with an out-of-plane kernel size of 7, matching the 2D kernel size of the original 2D weights, performs compared to inflating the ConvNeXt model with an out-of-plane kernel size of 3. We refer to the model with an out-of-plane kernel size of 3 as ConvNeXt-B* in Figure~\ref{fig:phecodes}c. Additionally, we compare various model sizes in Figure~\ref{fig:phecodes}c. We perform these architecture experiments using the phenotype classification task due to the simplicity of this supervised task and the strong evaluation signal that results from averaging performance across a large number of phenotypes.

\noindent \textbf{Model Training:} We use binary cross entropy for the phenotype classification loss and InfoNCE~\cite{oord2018representation, radford2021clip} loss for contrastive learning with the radiology reports, where the embeddings for both the image and text encoders in our implementation of the InfoNCE loss had a dimension of 512. This matches the embedding dimension used in the ViT-Base model in OpenCLIP experiments~\cite{cherti2023reproducible}. We use an AdamW~\cite{loshchilov2017decoupled} optimizer with an initial learning rate of 1e-5, betas=$(0.9, 0.999)$, and a cosine learning rate scheduler with number of epochs for decay to 0 set to 300. We use gradient checkpointing for both the visual and text encoders and train with FP16 mixed precision. This allows us to maximize a batch size of 18 on a single 48GB A6000 GPU.

\noindent \textbf{Multi-Task Learning Versus Staged Training:} In addition to training using the EHR phenotypes and radiology reports jointly in a multi-task manner, we consider staging the training. In this formulation, we first train the Merlin image encoder using the EHR phenotypes in stage 1. In stage 2, we perform contrastive training with the radiology reports. To prevent catastrophic forgetting~\cite{kirkpatrick2017overcoming} of the EHR information learned in stage 1, we include the phenotype loss function during stage 2 training, with a low relative weight. We use the same hyper-parameters for stage 2 as we do for multi-task training. For stage 1, we use an AdamW optimizer with an initial learning rate of 1e-4 and betas=$(0.9, 0.999)$, as well as an exponential learning rate scheduler with gamma = 0.99. We use a batchsize of 22 on a single A6000 GPU.

\noindent \textbf{Data Splits:} We divide the pretraining dataset into splits of size 60\% (15,331 CTs) for training, 20\% (5,060 CTs) for validation, and 20\% (5,137 CTs) for testing. We ensure that multiple CTs from a single patient do not exist in a single split. The external dataset, from another university medical center, consists of 7,000 CTs which are used for testing. The external dataset
from out-of-state site 1 consisted
of 6,997 abdominal CT scans used for testing. In-state external sites
2 and 3 included 25,997 and 4,872 abdominal CT scans, respectively,
used for testing; site 3 additionally included 6,243 chest CT scans used
for testing.

\subsection{Evaluations}
\label{sec:evaluation}

We select evaluation tasks consisting of non-adapted tasks, which Merlin can perform out of the box without additional adaptation. These tasks include zero-shot findings classification, phenotype classification, and zero-shot cross-modal retrieval. We also select tasks that require adapting Merlin, which include 5-year disease prediction, radiology report generation, and 3D semantic segmentation. We select both non-adapted tasks and adapted tasks to demonstrate Merlin's effectiveness without fine-tuning, as well as its adaptability when fine-tuned for specific applications. To comprehensively evaluate Merlin's performance against alternative architectures, we compare Merlin to finetuned 2D VLMs, 2D-to-3D lifted VLMs, and 3D vision-only models in data-scarce and fully-supervised settings.

\subsection{Non-Adapted Tasks}
\label{sec:zero_shot_findins}
\noindent\textbf{Zero-Shot Findings Classification:} We consult three radiologists to develop a list of 30 findings that exhibit a diversity of size and level of difficulty for human diagnosis. For each finding, we generate lists of disease presence phrases that describe possible sub-types, locations, and severities of the finding. We similarly generate a list of disease absence phrases, which are ways of describing a negative finding. For example, a disease presence phrase for ascites could be \textit{``large volume ascites present"} and a disease absence prompt could be \textit{``no ascites"}. We use these phrases to mine the radiology reports for negative and positive examples. To standardize the chance metric value across findings, we balance the number of positive and negative examples for each finding. After extracting these examples, we manually review them to ensure that the labels are accurate.

To perform zero-shot classification, we follow the procedure in Figure~\ref{fig:zero_shot}a where we embed the CT scan using the image encoder. We use the disease presence and absence phrases as prompts for zero-shot classification. We embed each of the prompts using the text encoder. We then compute the cosine similarity between the CT scan embedding and each of the prompts. Computing the mean cosine similarity across the disease presence prompts and mean similarity across the disease absence prompts allows us to classify the CT scan as positive or negative for a given finding. We adapt existing 2D baselines (OpenCLIP~\cite{cherti2022reproducible} and BioMedCLIP~\cite{zhang2023biomedclip}) for 3D zero-shot classification by embedding every axial slice of a given CT image. To compute a similarity score between a CT scan and a given prompt, we take the mean of the cosine similarities between each axial slice and the prompt embedding.

We also perform zero-shot spine fracture detection using the VerSe dataset~\cite{loffler2020vertebral}. All CTs in the VerSe dataset were evaluated for fracture severity at each vertebral level in the thoracolumbar spine. We reorient and resample the VerSe CTs to match Merlin pretraining. For volumes with fractures, we create a probability map from the fracture grading annotations and sample high-probability locations to obtain representative sub-volumes. We map the Hounsfield unit range -1000:1000 to 0:1 and spatially pad and crop the images to $224\times224\times160$. For volumes without fractures, we center crop the CTs.

We perform an ablation study where we measure the impact of I3D initialization, staged versus multi-task training with EHR and radiology reports, and splitting the report text with every other batch for finer grain contrastive learning (Figure \ref{fig:zero_shot}e). We also examine how zero-shot performance varies across the pretraining dataset sizes of 1\%, 10\%, 20\%, 40\%, and 100\% of the total training set (Figure \ref{fig:zero_shot}d).

\noindent\textbf{Phenotype Classification:} \label{method:phenotype_cls} To generate the labels for this task, we group ICD-9 and ICD-10 codes into phenotypes using the PheWAS~\cite{denny2013systematic} phecode mapping. We compare Swin Transformer, ConvNeXt, and ResNet backbone architectures (Figure \ref{fig:phecodes}c), as well as various model stem hyper-parameters (Figure \ref{fig:phecodes}d). See section \ref{Model Architecture} for additional details. We report average AUROC and AUPRC using all phenotypes that have more than 20 positive examples in the test set, in order to ensure a meaningful measure of performance. We additionally assess how phenotype classification performance varies across the pretraining dataset sizes of 1\%, 10\%, 20\%, 40\%, and 100\% of 15,331 total pretraining samples.

\noindent\textbf{Zero-Shot Cross-Modal Retrieval:} \label{method:zero_shot} In Figure \ref{fig:retrieval}(a) we illustrate the procedure for performing zero-shot retrieval. First, we sample a pool of report findings sections. Then, we embed each of the findings sections using the text encoder and the corresponding CT scans using the image encoder. We then compute the cosine similarities between the images and each of the report findings sections. Based on these cosine similarities, we rank the report findings in order of their similarity to each of the images.

To evaluate retrieval performance, we divide the test dataset into non-overlapping pools and within each pool evaluate how often the cosine similarity between corresponding CT images and reports is the highest (Recall@1). To compute the overall score, we average performance across all pools. We compute retrieval performance for surfacing the most similar findings section given a CT image (Image $\rightarrow$ Findings in Figure \ref{fig:retrieval}c) and the most similar CT image given a findings section (Findings $\rightarrow$ Image in Figure \ref{fig:retrieval}c). We compute retrieval performance between images and findings sections on both our internal test dataset (5,137 pairs) and our external dataset (7,000 pairs).

To validate that the model is not learning a shortcut to match images with findings sections, which seems plausible given the length of the findings sections, we also evaluate retrieval between the images and impressions sections. Given that during training, Merlin is only exposed to findings sections from the reports, evaluating on semantically similar, yet distinct impressions sections can provide a measure of generalizability. We also perform this analysis on our internal test dataset and the external dataset (Figure \ref{fig:retrieval}c).

Similar to zero-shot classification, we adapt existing 2D baselines (OpenCLIP and BioMedCLIP) for 3D cross-modality retrieval by embedding each axial slice of the CT volumes.

As with zero-shot classification, we perform ablation studies where we examine the impact of I3D initialization, staged versus multi-task training with EHR and radiology reports, and radiology report splitting (Figure \ref{fig:retrieval}d). We also perform data scaling experiments that assess how retrieval performance varies across pretraining dataset sizes of 1\%, 10\%, 20\%, 40\% and 100\% of the full pretraining dataset (Figure \ref{fig:retrieval}e).

\subsection{Adapted Tasks}

\noindent\textbf{Multi-Disease 5-Year Prediction:} Following previous work \cite{blankemeier2022opportunistic}, we choose 6 chronic diseases based on their prevalence and the potential for beneficial interventions following early diagnoses: chronic kidney disease (abbr. CKD; prevalence = 36 million in US~\cite{ckd}), diabetes mellitus (abbr. DM; prevalence = 35 million in US~\cite{dm}), hypertension (abbr. HTN; prevalence = 120 million in US~\cite{htn}), ischemic heart disease (abbr. IHD; prevalence = 21 million in US~\cite{ihd}), atherosclerotic cardiovascular disease (abbr. CVD; prevalence = 24 million in US~\cite{ascvd}), and osteoporosis (abbr. OST; prevalence = 10 million in US~\cite{ost}). 

To create the labels for this task, we extract disease ICD codes of interest from the EHR. We assign patients to one of 4 categories for each disease based on their ICD codes. Formally, we define $t_d$ to be the time point marking the onset of the disease based on the presence of ICD codes, and $t_s$ to be the time point marking the date of the CT scan. $t_a$ is the time point after a CT scan that defines a window starting with the scan date ($t_s$ to $t_a$), where the presence of a new diagnosis indicates a positive case. We set $t_a$ to 5 years to ensure an adequate time duration for patient followup following the CT imaging as well as providing an adequate duration for intervention in future prospective studies. We define $t_h$ to be the time point marking the last date of record in a patient's history. With these time points, we classify images into the following categories: 
\begin{itemize}
    \item \noindent\textbf{Class 0 (Healthy):} The patient is not diagnosed with the specified diseases before their CT scan date ($t_s$) or during the window from $t_s$ to $t_a$. If a patient is diagnosed with the disease, it happens after $t_a$. For patients to be assigned to class 0, they must have 5 years of followup in their EHR history. 
    \item \noindent\textbf{Class 1 (Progressors):} The patient is not diagnosed with the disease before their CT date, $t_s$, but receives a positive diagnosis between $t_s$ and $t_a$. 
    \item \noindent\textbf{Class 2 (Already Progressed):} The patient is already diagnosed with the disease before the CT scan date, $t_s$.
    \item \noindent\textbf{Class 3 (Censored):} The patient has not been monitored long enough to rule out the disease. In this case, $t_h$ is before $t_a$ and the patient does not receive a positive diagnosis before the end of their EHR history, $t_h$.
\end{itemize}

For training, we use class 1 examples as positive examples and class 0 examples as negative examples. We fine-tune Merlin on examples that are held out from Merlin pretraining and validation. Furthermore, we do multi-task multi-disease prediction where the fine-tuned Merlin has one head per disease as shown in Figure \ref{fig:disease_prediction}a. Since we train jointly on all the diseases, we mask labels in a batch that are from classes other than 0 and 1 for a specific disease. We use binary cross entropy loss, an AdamW optimizer with a learning rate of 1e-5, an exponential learning rate decay with $\gamma=0.8$, and a batch size of 8 on a single A6000 GPU. We use 60\%, 20\%, 20\% train, validation, and test splits, and measure performance in both the low-label regime (10\% of training labels with 9-20 positive examples per disease), as well as the medium-label regime (100\% of available training labels with 69-136 positive examples per disease). The number of per-disease positive and negative examples are given in Figure \ref{fig:disease_prediction}.

\noindent\textbf{Radiology Report Generation:} Figure~\ref{fig:report_generation}a demonstrates the procedure for adapting Merlin for report generation. We first extract image features from the last hidden layer of Merlin, which has size 7x7 in-plane, 10 out-of-plane, and a feature dimension of 2048. We use a single linear adapter layer to map the features to size 4096. For text generation, we use RadLlama-7B~\cite{chen2024chexagent}, a version of Llama2-7B that is fine-tuned on MIMIC~\cite{johnson2023mimiciv} radiology reports and clinical notes. We fine-tune the linear adapter, as well as 5\% of RadLlama-7B parameters using low-rank adaptation (LoRA)~\cite{hu2021lora}. We generate the full reports section by section, where we use the following prompt template:
\begin{verbatim}
<visual tokens>Generate a radiology report for <organ system>###
<report section>###</s>
\end{verbatim} 

For training, we use 8 gradient accumulation steps with a local batch size of 6, giving us a total batchsize of 48. We use an AdamW optimizer with betas=$(0.9, 0.999)$, a learning rate of 1e-4, and a cosine learning rate scheduler with 500 warmup steps and decay to 0 over 500 epochs.

We compare our model performance on report finding section generation versus RadFM~\cite{wu2023generalist} using four metrics: BLEU score~\cite{bleu} and ROUGE-2~\cite{rouge}, which primarily assess syntactic similarity; RadGraph-F1~\cite{delbrouck2022improving}, which assesses findings and modifiers of findings (e.g. location and severity); and BERT score~\cite{zhang2020bertscore}, a model-based score which assesses semantic similarity. The comparison is conducted both section-by-section and for the full findings sections. Subsequently, we apply qualitative frameworks~\cite{van2023radadapt, vanveen2024clinical} to compare our model generated findings to radiologist findings in Figure~\ref{fig:report_generation}c. This consists of densely annotating individual phrases to be correct, mischaracterized, false positive, or false negative.

\noindent\textbf{3D Semantic Segmentation:} To adapt Merlin for segmentation, we add a UNet~\cite{ronneberger2015u} decoder and skip connections between the Merlin encoder and the decoder. Each block of the decoder consists of a 3D transpose convolution operation, with a kernel size of 2 and a stride of 2 in each dimension, followed by two 3D convolutions, with kernel sizes of 3 and strides of 1 in each dimension. Each of the two 3D convolution operations is followed by a 3D batchnorm and ReLU activation. We also add skip connections where outputs from the Merlin ResNet blocks are concatenated with outputs from the 3D transpose convolution in the decoder before being passed to the subsequent 3D convolutional layers. To mirror pretraining, we segment full volumes of size 224x224x160.

For model training and evaluation, we filter body CT scans from the Total Segmentator dataset using the following study types: ct pelvis, ct abdomen-pelvis, ct abdomen, ct thorax, ct thorax-abdomen, and ct thorax-abdomen-pelvis. This gives us 401 scans total. We also select the following organs that appear in abdominal CT: stomach, liver, gallbladder, left kidney, right kidney, spleen, prostate, T12 vertebrae, L1 vertebrae, L2 vertebrae, L3 vertebrae, L4 vertebrae, L5 vertebrae, S1 vertebrae, sacrum, urinary bladder, colon, duodenum, small bowel, and pancreas. We use the official Total Segmentator test split for testing, which has 34 scans after filtering. We split the remaining 367 scans into 80\% training examples (293 scans) and 20\% validation examples (74 scans). In addition to training with all of the training scans, we simulate the label scarce setting by sampling 10\% of training scans randomly.

We evaluated various network architectures and initialization strategies within the nnUNet framework. Specifically, we compared the nnUNet 3D full-resolution architecture with other configurations, including a Swin UNETR~\cite{tang2022self}, an architecture specifically designed for 3D medical image segmentation, and a Resnet152 encoder-decoder architecture initialized with random, Imagenet, and Merlin weights (Figure \ref{fig:segmentation}a). The training setup followed the default nnUNet configuration, including data augmentation steps, a 1000-epoch training schedule, and a polynomial learning rate scheduler. We evaluated the model in both 10\% and 100\% training data domains using Dice score.

\subsection{Alternative Architecture Experiments}

\noindent\textbf{2D VLMs:} We trained three 2D vision-language foundation models: a 2D Resnet152 vision encoder paired with a clinical longformer text encoder, OpenCLIP~\cite{cherti2023reproducible} finetuned on abdominal CT data, and BiomedCLIP~\cite{zhang2023biomedclip} finetuned on abdominal CT data (Figure \ref{fig:baseline_experiments}a). Each model used MTL, incorporating EHR phenotypes and radiology reports split into anatomical sections, to ensure a comparable training approach to the 3D Merlin model. This setup finetuned both the image and text encoders using the EHR and a contrastive loss head. During training, we randomly selected a single slice from each 3D volume as input. The Resnet152 model was initialized with ImageNet weights. All models were finetuned with a batch size of 128, 100 epochs, using 100\% of the training data. 

\noindent\textbf{2D-to-3D lifted VLMs:} To adapt the aforementioned 2D vision-language models for 3D data, we finetuned the models by processing each slice of the abdominal CT volume individually and concatenating them using a learned projection layer (Figure \ref{fig:baseline_experiments}b). Instead of naive averaging, we propose transforming the features from each highly-redundant 2D slicewise embeddings from the 3D volume into an unified embedding through this learned layer. More specifically, the 160 slice embeddings are processed through a sequential linear transformation, first reducing them to 64 embeddings with a ReLU non-linearity, and then further compressed into a single embedding of size 512. The image encoder and classification heads from the VLM were then finetuned for 10 epochs on 100\% of the training data, with maximal attainable batch sizes of 2, 8, and 16 for 2D Resnet152, OpenCLIP, and BiomedCLIP, respectively.

\noindent\textbf{3D vision-only models:} We trained two state-of-the-art 3D vision-only foundation models: Swin UNETR\cite{hatamizadeh2022unetr} and Inflated 3D (I3D) Resnet152, using a self-supervised learning (SSL) pretraining strategy~\cite{tang2022self} (Figure \ref{fig:baseline_experiments}c). The SSL pretraining process involved randomly cropping abdominal CT volumes into subvolumes, followed by augmentations such as random cutout and rotation. These augmented subvolumes were then fed into the models, with pretraining tasks including inpainting, contrastive learning, and rotation prediction to learn meaningful image representations, based on prior work~\cite{tang2022self}. For Swin UNETR, pretraining was performed with a batch size of 1, a sliding window size of 2, and 50K training steps. For I3D Resnet152, the model utilized convolutional embeddings with convolutional transpose layers for the L1 reconstruction loss. Finetuning the I3D Resnet152 was conducted with a batch size of 3, a sliding window size of 2, and 50K training steps.

\subsection{Alternative Architecture Evaluation}

To ensure a fair comparison between the vision-only and vision-language baselines, we evaluate the models on two tasks: findings-based disease classification and EHR phenotype classification. We here describe the evaluation tasks, which involve training linear-probes while keeping the image encoder frozen. This setup allows us to compare pretrained image representations in isolation and compare Merlin to recent CT embedding foundation models\cite{codella2024medimageinsight, yang2024advancing}. Full finetuning results, where both the image encoder and classification heads are trained, are presented in Supplemental Section \ref{suppl:embedding_comparison}. For the 3D vision-only models, the embedding dimensions were 768 for SwinUNETR and 2048 for I3D Resnet152. For the 2D and the 2D-to-3D VLMs, the embedding dimensions were 768 for BiomedCLIP, 2048 for ResnetCLIP, and 1024 for OpenCLIP. For the CT embedding foundation models, MedImageInsight used an embedding dimension of 1024, while Google CT Foundation used an embedding dimension of 1408. Details on processing the Merlin dataset through these CT embedding foundation models are provided in Supplemental Section \ref{suppl:embedding_comparison}.

\noindent\textbf{Findings-based disease classification task:} The findings-based disease classification task was adapted from the zero-shot findings classification task (Section \ref{method:zero_shot}), where the 30 findings originally used for zero-shot evaluation were reformulated for supervised learning. We parsed the radiology reports and categorized each finding as positive, negative, or missing from the report. While most conventional strategies involve discarding missing labels during training, we adopted a more inclusive strategy to use all our data. We framed the task as a multilabel classification task, introducing thirty separate binary heads, one for each finding. Image embeddings were used as input to the multihead linear classifiers, and the loss was masked for subjects missing a specific finding before backpropagation\cite{xue2024ai, blankemeier2022opportunistic}. Linear probes were trained from scratch using Kaiming initialization with the AdamW optimizer, an initial learning rate of 1e-5, and a cosine decay scheduler (300 epochs to zero learning rate), for a total of 10 epochs using binary cross-entropy loss on the abdominal CT dataset (15,331 scans for training and 5,060 for validation). We evaluated the performance of these baselines using 10\% and 100\% of the training data. This approach standardizes image encoder evaluation across various model architectures, including 3D vision-only foundation models, 2D vision-language models, 2D finetuned vision-language models, 2D-to-3D lifted vision-language models, MedImageInsight, Google CT foundation model, and Merlin.

\noindent\textbf{Phenotype classification} For the EHR phenotype classification task, we train a linear probe that maps the embeddings to 1,692 target labels. The probe consists of a single linear layer with input dimensionality matching the model's image encoder embedding dimensions and an output of 1,692 classes. We use the same optimizer, scheduler, training hyperparameters, and image dataset as in the multihead linear probe experiments (findings-based disease classification task). All linear probes were trained from scratch using Kaiming initialization to maintain consistent comparisons across models. Similar to the findings-based disease classification task, we finetuned the models on 10\% and 100\% of the training data to assess classification performance across different dataset sizes.

\subsection{External Validation Experiments}

We evaluate Merlin on 44,098 external CT scans, comprising 37,855 abdominal CTs from three external sites [Site \#1: n = 6,997, out-of-state site; Site \#2: n = 25,986, in-state site; Site \#3: n = 4,872, in-state site] and 6,243 chest CTs from Site \#3.

\noindent\textbf{Zero-shot findings classification:} We evaluate Merlin on the zero-shot findings classification task and compare Merlin against alternative architecture baselines: 2D/2D-to-3D BiomedCLIP, OpenCLIP, and ResnetCLIP. All models were fine-tuned on the Merlin dataset in a multi-task manner, analogous to Merlin's training. This setup ensured that observed differences reflected architectural paradigms (2D versus 2D-to-3D versus full 3D [Merlin]) rather than differences in pretraining data quality. Ground-truth labels for zero-shot classification were derived by matching positive and negative prompts to the findings sections of radiology reports using regular expressions. The same positive and negative prompts as in the internal test set (Section ~\ref{sec:zero_shot_findins}) were used. For the External Site \#3, labels were generated using the Qwen3-30B LLM~\cite{yang2025qwen3} applied to the findings section of their abdominal CT reports. 

\noindent\textbf{Chest CT linear probe evaluation:} We leveraged a dataset of 63,711 chest CTs from External Site \#3, comprising 57,468 scans for training and 6,243 for testing. We specifically only trained a linear probe on the frozen Merlin image encoder to assess the quality of the baseline Merlin embeddings. Each scan is paired with a radiology report, and labels were derived from the findings section using the Qwen-30B LLM~\cite{yang2025qwen3} to identify 24 conditions: lung nodules, consolidation, ground-glass opacities, atelectasis, interstitial thickening, emphysema, bronchiectasis, pulmonary embolism, pulmonary edema, lung cavitations, pleural effusion, pneumothorax, pleural thickening, lymphadenopathy, dilated aorta, aortic aneurysm, aortic dissection, atherosclerosis, pericardial effusion, metastases, fracture, hepatic cysts, liver mass, and hiatal hernia. We trained only a \textit{linear probe} using balanced logistic regression to correct for skewed disease prevalence, without explicit resampling. The implementation was based on the scikit-learn library, mapping each model's pretrained embeddings to the extracted labels. From a model comparison perspective, we evaluated Merlin against recent chest CT foundation models—CT-CLIP~\cite{hamamci2024developing}, M3FM~\cite{niu2025medical}, and CT-FM~\cite{pai2025vision}—by training a separate linear probe for each disease and model.

\subsection{Statistical Analysis}
We compute 95\% confidence intervals using bootstrapping with 1000 samples with replacement for all experiments except for cross-modal retrieval. For cross-modal retrieval, we divide the test dataset into pools of size N. For each pool, we compute the average retrieval performance for all examples in the pool. We then compute the 95\% confidence intervals using the distribution of performances across pools. All p-values that we report are one-sided p-values.

\subsection{External Validation Methods}

\subsection*{Acknowledgements} A.S.C. receives research support from NIH grants R01 HL167974, R01HL169345, R01 AR077604, R01 EB002524, R01 AR079431, P41 EB 027060 and P50 HD118632; the Advanced Research Projects Agency for Health (ARPA-H) Biomedical Data Fabric (BDF) and
the Chatbot Accuracy and Reliability Evaluation (CARE) programmes (contracts AY2AX000045 and 1AYSAX0000024-01); and the Medical Imaging and Data Resource Center (MIDRC), which is funded by the National Institute of Biomedical Imaging and Bioengineering (NIBIB) under contract 75N92020C00021 and through the ARPA-H. C.B. receives research support from the Promedica Foundation.

\subsection*{Contributions}
L.B. and A.K. collected data, developed code, trained models, ran
experiments, analysed results, created figures and wrote the manuscript. All authors reviewed the manuscript and provided revisions and feedback. J.P.C., A.K., D.V.V., M.P., Z.C., J.-B.D., E.R., R.H., C.B., M.E.K.J., S.O., M.V., J.M.J.V., Z.F., Z.N., D.A., W.-H.W., S.G. and A.S.C. provided
technical advice. J.P.C. developed the counterfactual generation method for CT scans and ran counterfactual experiments. A.K. carried out model evaluations on external datasets. J.L. developed a Merlin dataset inference pipeline for the Google CT. D.V.V. assisted in collecting
zero-shot evaluation labels and facilitated radiologist annotations of generated reports. S.J.S.G., H.Y. and A.W. ran model inference on external clinical dataset 1. L. Liu, L. Lian, Y.W. and A.Y. ran model inference on external clinical dataset 3. M.P., Z.C., J.-B.D., E.R., C.T. and E.A.J. assisted with model evaluations. Z.H. and J.F. assisted with dataset anonymization. C.B., E.A.J., N.A., G.Z., M.W., A.J., R.D.B., A.W., C.P.L., M.W., J.H. and S.G. provided clinical input and feedback. C.B. provided counterfactual annotations. S.G. and C.B. provided annotations of generated reports. N.H.S., C.P.L., S.G. and A.S.C. provided research support for the project. A.S.C. guided the project, serving as principal investigator and advising on technical details and overall direction. No funders or third parties were involved in study design, analysis or writing.

\FloatBarrier

\bibliographystyle{unsrt}
\bibliography{main}

\clearpage
\appendix
\tableofcontents

\resumetocwriting


\section{Extended Methods}
\label{appendix:methods}

\subsection{Dataset Details}

\begin{table}[h!]
\begin{center}
\begin{tabular}{lcccccc}
\toprule
\textbf{Split} & \textbf{2D Slices} & \textbf{CTs} & \textbf{Findings Tokens} & \textbf{ICD9 Codes} & \textbf{ICD10 Codes} &
\textbf{Patients} \\
\midrule
Train & 6,387,231 & 15,331 & 6,036,645 & 577,998 & 1,261,561 & 11,010 \\
Validation & 2,099,217 & 5,060 & 1,985,925 & 178,194 & 379,733 & 3,644 \\
Test & 2,142,061 & 5,137 & 2,029,001 & 197,821 & 399,986 & 3,667 \\
\midrule
Total & 10,628,509 & 25,528 & 10,051,571 & 954,013 & 2,041,280 & 18,321 \\
\bottomrule
\end{tabular}
\end{center}
\caption{\textit{Summary of pretraining dataset splits.}}
\label{suppl_tab:pretraining_summary}
\end{table}

\begin{table}[h!]
\begin{center}
\begin{tabular}{lcc}
\toprule
\textbf{Demographic} & \textbf{Patients (n = 18,321)} & \textbf{Value} \\
\midrule
\textbf{Age} & - & 53.8$\pm$19.5 \\
\textbf{Gender} & \\
\quad Female & 10,254 & 55.97\% \\
\quad Male & 8,065 & 44.02\% \\
\textbf{Self-Reported Race/Ethnicity} & \\
\quad Non-Hispanic White & 8,660 & 47.27\% \\
\quad Asian & 2,673 & 14.59\% \\
\quad Black & 952 & 5.20\% \\
\quad Hispanic White & 515 & 2.81\% \\
\quad Pacific Islander & 294 & 1.60\% \\
\quad Native American & 65 & 0.35\% \\
\quad Unknown & 5,127 & 27.98\% \\
\textbf{Patient Class} & \\
\quad Inpatient & 6,834 & 37.31\% \\
\quad Emergency Services & 6,388 & 34.87\% \\
\quad Outpatient & 2,959 & 16.16\% \\
\quad Observation & 1,452 & 7.93\% \\
\bottomrule
\end{tabular}
\end{center}
\caption{\textit{Internal dataset characteristics.} The age value is provided as mean $\pm$ standard deviation. All other values are provided as percentages of the total patients (n = 18,321).}
\label{suppl_tab:internal_dataset}
\end{table}

\begin{table}[h!]
\begin{center}
\begin{tabular}{lcc}
\toprule
\textbf{Demographic} & \textbf{Patients (n = 5,804)} & \textbf{Value} \\
\midrule
\textbf{Age} & - & 61.4$\pm$16.5 \\
\textbf{Gender} & \\
\quad Female & 2,982 & 51.38\% \\
\quad Male & 2,822 & 48.62\% \\
\textbf{Self-Reported Race/Ethnicity} & \\
\quad Non-Hispanic White & 4,576 & 78.84\% \\
\quad Black & 270 & 4.65\% \\
\quad Hispanic White & 198 & 2.76\% \\
\quad Asian & 87 & 1.50\% \\
\quad Native American & 31 & 0.53\% \\
\quad Pacific Islander & 4 & 0.07\% \\
\quad Unknown & 602 & 10.37\% \\
\bottomrule
\end{tabular}
\end{center}
\caption{\textit{External dataset characteristics.} The age value is provided as mean $\pm$ standard deviation. All other values are provided as percentages of the total patients (n = 5,804).}
\label{suppl_tab:external_dataset}
\end{table}

\FloatBarrier

\clearpage
\section{Extended Results}

\subsection{Zero-shot Classification}

\begin{table*}[h!]
\small
\begin{center}
\setlength{\tabcolsep}{5.6pt}
\begin{tabular}{llllc}
\toprule
\textbf{$Method$}
& \textbf{Average F1 Score}

\\ 
\midrule
\small OpenCLIP (Internal) & .276 [.262, .288] \\
\small BioMedCLIP (Internal) & .285 [.274, .295] \\
\small Merlin (Internal) & .741 [.727, .755] \\
\small Merlin (External) & .647 [.607, .678] \\
\small Merlin (VerSe) & .767 [.630, .867] \\
\bottomrule
\end{tabular}
\end{center}
\caption{\textit{Zero-shot classification.} The internal and external numbers (first 4 rows) represent averages over the 30 findings for the internal clinical dataset and the external clinical dataset respectively. The bottom row represents F1 score for zero-shot classification of vertebral fractures on the VerSe dataset.}
\label{table:zero_shot}
\end{table*}
\begin{table*}[h!]
\small
\begin{center}
\setlength{\tabcolsep}{5.6pt}
\begin{tabular}{lllc}
\toprule
\textbf{$Init$}
& \textbf{$Labels$}
& \textbf{$Split$}
& \textbf{Average F1 Score}
\\
&
& \textbf{$Text$}

\\ 
\midrule
I3D & Report & \cmark & .730 [.714, .744] \\
I3D & Staged & \xmark & .669 [.653, .683] \\
I3D & Staged & \cmark & \cellcolor[HTML]{EFEFEF}.735 [.719, .748] \\
I3D & MTL & \xmark & .656 [.640, .671] \\
Rand & MTL & \cmark & .698 [.681, .711] \\
I3D & MTL & \cmark & \cellcolor[HTML]{C0C0C0}\textbf{.741 [.727, .755]} \\
\bottomrule
\end{tabular}
\end{center}
\caption{\textit{Zero-shot classification ablation study.} We measure zero-shot performance as we vary parameters of I3D versus random initialization, staged training versus multi-task learning (MTL) with EHR and radiology reports, and training with the full findings sections versus using radiology report splitting.}
\label{table:zero_shot_ablation}
\end{table*}

\underline{Scaling Law Experiments:} We expand Figure \ref{fig:zero_shot}d by providing a quantitative analysis of how increasing the pretraining dataset affects zero-shot performance. This analysis allows us to estimate the amount of pretraining data needed to achieve a given zero-shot classification performance. The resulting power law is $\overline{F1} = 0.458D^{0.0524}$, where $D$ represents the number of paired pretraining CT imaging and report data.

\begin{figure}[h!]
\centerline{\includegraphics[width=1.0\textwidth]{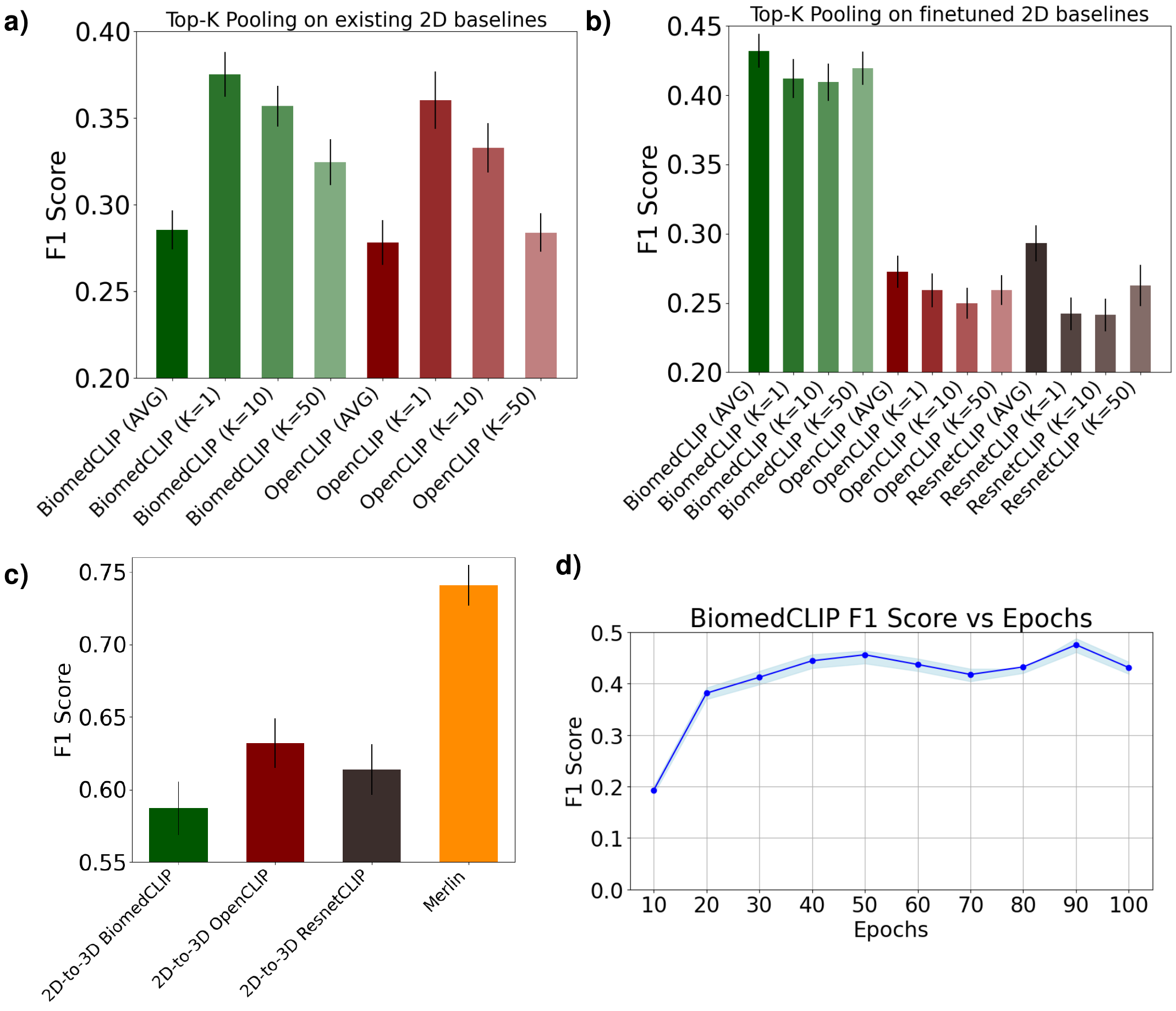}}
\caption{\textit{Zero-shot findings classification extended.} (a) Top-k pooling on existing 2D BiomedCLIP and OpenCLIP baselines. (b) Top-k pooling on 2D finetuned BiomedCLIP, OpenCLIP, and ResnetCLIP baselines. Finetuning involved training both the image and text encoders following the strategy outlined in Figure \ref{fig:baseline_experiments}b. (c) 2D-to-3D lifted VLM zero-shot evaluation. Merlin tends to perform better on this task compared to baselines from other training paradigms. (d) BiomedCLIP's F1 on zero-shot findings classification task across training epochs.}
\label{suppl_fig:top_k}
\end{figure}

\underline{Additional 2D and 2D-to-3D Experiments:} We expanded our analysis to present the top-k pooling results for existing baseline models and included the zero-shot findings evaluation to encompass finetuned 2D and 2D-to-3D lifted VLMs (Figure \ref{suppl_fig:top_k}). We observed that top-k pooling does increase F1 score compared to mean-pooling.

We first evaluated off-the-shelf 2D BiomedCLIP and OpenCLIP models, without finetuning, by calculating average F1 scores for top-k pooling (k=1, 10, 50). Our findings indicate that top-k pooling improves F1 scores for zero-shot evaluation compared to average pooling, with performance decreasing as k increases (Figure \ref{suppl_fig:top_k}a). For instance, when comparing k=1 to the average (AVG), BiomedCLIP shows a 32.0\% increase in F1 score, while OpenCLIP shows a 29.0\% increase in F1 score.

We repeated the top-k pooling experiments after 2D finetuning of BiomedCLIP, OpenCLIP, and ResnetCLIP (Figure \ref{suppl_fig:top_k}b). Interestingly, after finetuning, pooling across averaged slices yielded the highest F1 scores. When comparing average pooling F1 scores between the finetuned and non-finetuned versions of BiomedCLIP and OpenCLIP, we observed that finetuning increased BiomedCLIP’s F1 score by 52.0\%, whereas OpenCLIP’s F1 score decreased by 2.1\%. For instance, with the finetuned models, average pooling outperformed K=1 pooling, with BiomedCLIP showing a 4.9\% increase in F1 score, OpenCLIP a 5.0\% increase, and ResnetCLIP a 22.0\% increase. When comparing the finetuned 2D baselines on average pooling, BiomedCLIP achieved the highest F1 score of 0.432 (95\% CI [0.419-0.443]), while OpenCLIP and ResnetCLIP demonstrated lower F1 scores of 0.272 (95\% CI [0.260-0.283]) and 0.293 (95\% CI [0.279-0.305]), respectively. This trend may be attributed to the 2D finetuning process, which involved selecting a random slice for training. By minimizing reliance on specific slices, this approach reduces potential advantages from pooling. To assess potential overfitting during 2D finetuning, we examined the model’s performance across training epochs by tracking changes in zero-shot F1 score. The analysis revealed minimal signs of overfitting, with training performance saturating around 30 epochs (Figure \ref{suppl_fig:top_k}d).

Finally, we compared the 2D-to-3D lifted VLMs to Merlin on the zero-shot findings classification task. Among the aggregation strategies, 2D-to-3D OpenCLIP achieved the highest F1 score of 0.632 (95\% CI [0.613-0.647]), followed by 2D-to-3D ResnetCLIP with an F1 score of 0.614 (95\% CI [0.595-0.623]) and BiomeCLIP with 0.587 (95\% CI [0.568-0.604]) (Figure \ref{suppl_fig:top_k}c). Despite these results, OpenCLIP's F1 score was still 17.0\% lower than that of Merlin. 

\color{black}

\FloatBarrier

\clearpage
\subsection{Phenotype Classification}

\begin{table*}[h!]
\small  
\begin{center}
{
\setlength{\tabcolsep}{4.3pt}
\begin{tabular}{llllcccccccc}
\toprule
\textbf{$Encoder$}
& \textbf{$Init$}
& \textbf{$Labels$}
& \textbf{$Stem$}
& \multicolumn{4}{c}{\textbf{\small All Phecodes}}
& \multicolumn{2}{c}{\textbf{\small Upper Quartile}}
& \multicolumn{2}{c}{\textbf{\small Lower Quartile}}
\\
& & & \footnotesize KS$_z$/ & \multicolumn{4}{c}{\footnotesize N=691, Prev=3.1\%}
& \multicolumn{2}{c}{\textbf{\small by Prevalence}}
& \multicolumn{2}{c}{\textbf{\small by Prevalence}} \\
& & & \footnotesize Stride$_z$ & & &  \multicolumn{2}{c}{\footnotesize Phecodes w/ AUC}
& \multicolumn{2}{c}{\footnotesize N=173; Prev=8.7\%}
& \multicolumn{2}{c}{\footnotesize N=173, Prev=0.6\%} \\
\cmidrule(l{3pt}r{0pt}){5-6}
\cmidrule(l{3pt}r{0pt}){7-8}
\cmidrule(l{3pt}r{0pt}){9-10}
\cmidrule(l{3pt}r{0pt}){11-12}
&
&
&
& \footnotesize AUROC
& \footnotesize AUPRC
& \footnotesize >0.85
& \footnotesize >0.9
& \footnotesize AUROC
& \footnotesize AUPRC
& \footnotesize AUROC
& \footnotesize AUPRC
\\ 
\midrule
\small SwinUNETR & \small MAE & \small EHR & 4/4 & .736 & .088 & 52 & 8 & .734 & .207 & .738 & .029
\vspace{5pt} \\
\small ConvNeXt-T & \small I3D & \small EHR & 7/2 & .768 & .106 & 115 & 20 & .765 & .239 & .764 & .035 \\
\small ConvNeXt-S & \small I3D & \small EHR & 7/2 & .761 & .102 & 105 & 14 & .760 & .234 & .750 & .031 \\
\small ConvNeXt-B & \small I3D & \small EHR & 7/2 & .773 & .110 & 132 & 28 & .768 & .244 & .766 & .036 \\
\small ConvNeXt-B* & \small I3D & \small EHR & 3/2 & .789 & .131 & 180 & 46 & .784 & .270 & .781 & \cellcolor[HTML]{EFEFEF}.052
\vspace{5pt} \\
\small ResNet50 & \small I3D & \small EHR & 3/1 & .798 & .137 & 226 & 72 & .787 & \cellcolor[HTML]{EFEFEF}.280 & .797 & .049  \\
\small ResNet152 & \small I3D & \small EHR &  7/2 & .798 & .135 & 221 & 65 & .785 & .272 & .796 & .050 \\
\hspace{15pt} \small $\downarrow$ & \small I3D & \small EHR & 3/2 & .798 & .136 & 221 & 74 & .788 & .275 & .792 & .049 \\
\small & \small I3D & \small EHR & 3/1 & \cellcolor[HTML]{EFEFEF}.801 & \cellcolor[HTML]{EFEFEF}.140 & \cellcolor[HTML]{EFEFEF}235 & \cellcolor[HTML]{EFEFEF}79 & \cellcolor[HTML]{EFEFEF}.789 & .279 & \cellcolor[HTML]{EFEFEF}.801 & \cellcolor[HTML]{C0C0C0}\textbf{.054} \\
\small (Merlin) & \small I3D & MTL & 3/1 & \cellcolor[HTML]{C0C0C0}\textbf{.812} & \cellcolor[HTML]{C0C0C0}\textbf{.142} & \cellcolor[HTML]{C0C0C0}\textbf{259} & \cellcolor[HTML]{C0C0C0}\textbf{93} & \cellcolor[HTML]{C0C0C0}\textbf{.804} & \cellcolor[HTML]{C0C0C0}\textbf{.290} & \cellcolor[HTML]{C0C0C0}\textbf{.808} & .050 \\
\bottomrule
\end{tabular}
}
\end{center}
\caption{\textit{Phenotype classification.} We only include phenotypes that have more than 20 positive examples in the test set in order to ensure a meaningful measure of performance. * in ConvNext-B* indicates that instead of inflating the z dimension to a size equal to the 2D kernel height and width of 7, the kernel is inflated to a depth of 3.}
\label{table:phenotype_classification}
\end{table*}

\begin{table*}[h!]
\small  
\begin{center}
{
\setlength{\tabcolsep}{4.3pt}
\begin{tabular}{lllllcccccccc}
\toprule
\textbf{$Init$}
& \textbf{$Labels$}
& \textbf{$Split$}
& \textbf{$Stem$}
& \multicolumn{4}{c}{\textbf{\small All Phecodes}}
& \multicolumn{2}{c}{\textbf{\small Upper Quartile}}
& \multicolumn{2}{c}{\textbf{\small Lower Quartile}}
\\
& & \textbf{$Text$} & \footnotesize KS$_z$/ & \multicolumn{4}{c}{\footnotesize N=691, Prev=3.1\%}
& \multicolumn{2}{c}{\textbf{\small by Prevalence}}
& \multicolumn{2}{c}{\textbf{\small by Prevalence}} \\
& & & \footnotesize Stride$_z$ & & & \multicolumn{2}{c}{\footnotesize Phecodes w/ AUC}
& \multicolumn{2}{c}{\footnotesize N=173; Prev=8.7\%}
& \multicolumn{2}{c}{\footnotesize N=173, Prev=0.6\%} \\
\cmidrule(l{3pt}r{0pt}){5-6}
\cmidrule(l{3pt}r{0pt}){7-8}
\cmidrule(l{3pt}r{0pt}){9-10}
\cmidrule(l{3pt}r{0pt}){11-12}
&
&
&
& \footnotesize AUROC
& \footnotesize AUPRC
& \footnotesize >0.85
& \footnotesize >0.9
& \footnotesize AUROC
& \footnotesize AUPRC
& \footnotesize AUROC
& \footnotesize AUPRC
\\ 
\midrule
\small I3D & \small Staged & \xmark & 3/1 & .804 & .145 & 240 & 84 & .792 & .287 & .798 & \cellcolor[HTML]{EFEFEF}.056 \\
\small I3D & \small Staged & \cmark & 3/1 & .807 & \cellcolor[HTML]{EFEFEF}.146 & 249 & 87 & .795 & .291 & .801 & \cellcolor[HTML]{EFEFEF}.056 \\
\small I3D & \small MTL & \xmark & 3/1 & \cellcolor[HTML]{C0C0C0}\textbf{.814} & \cellcolor[HTML]{C0C0C0}\textbf{.153} & \cellcolor[HTML]{C0C0C0}\textbf{267} & \cellcolor[HTML]{C0C0C0}\textbf{103} & \cellcolor[HTML]{C0C0C0}\textbf{.806} & \cellcolor[HTML]{C0C0C0}\textbf{.302} & \cellcolor[HTML]{EFEFEF}.807 & \cellcolor[HTML]{C0C0C0}\textbf{.058} \\
\small Rand & \small MTL & \cmark & 3/1 & .786 & .124 & 117 & 46 & .778 & .261 & .779 & .042 \\ 
\small I3D & \small MTL & \cmark & 3/1 & \cellcolor[HTML]{EFEFEF}.812 & .142 & \cellcolor[HTML]{EFEFEF}259 & \cellcolor[HTML]{EFEFEF}93 & \cellcolor[HTML]{EFEFEF}.804 & \cellcolor[HTML]{EFEFEF}.290 & \cellcolor[HTML]{C0C0C0}\textbf{.808} & .050 \\
\bottomrule
\end{tabular}
}
\end{center}
\caption{\textit{Phenotype classification ablation study.} We perform ablation studies where we examine the impact of I3D initialization, staged training versus multi-task learning (MTL) with EHR and radiology reports, and splitting the report text with every other batch for finer grain contrastive learning.}
\label{table:phenotype_classification_ablation}
\end{table*}

\underline{Scaling Law Experiments:} We expand Figure \ref{fig:phecodes}b by computing data scaling law curves and performing data analysis to determine power laws. The resulting relationships are $\overline{AUROC} = 0.479D^{0.0568}$ and $\overline{AUPRC} = 0.0157x^{0.239}$, where $D$ represents the number of paired pretraining CT imaging and report data.

\underline{\textit{Counterfactual Analysis}} 

\underline{Methods:} Counterfactual analysis (Figure \ref{fig:phecodes}(e)) sought to evaluate whether a model is using expected features during phenotype classification. Models can instead learn shortcuts from spurious correlations in the training data \cite{Ross2017rrr, geirhos2020shortcut,Cohen2021Problems}, which are easier to learn compared to more subtle physiological features that the model should use. We employ the Latent Shift approach \cite{Cohen2021gif} for counterfactual generation. This method uses a latent variable model, which represents the input image in a low dimensional space and reconstructs the image in its output. The gradient of the Merlin prediction is passed back to the low-dimensional space, where the latent representation is modified such that Merlin's prediction decreases. This generates a modified image at the output of the latent variable model, which can be observed. The latent variable model regularizes the change in pixels so that the image remains close to the data manifold. The result of the Merlin counterfactual generation process is a new CT volume, which is evaluated by observing how the classifier prediction changes and how the image features driving the classification output are exaggerated or curtailed. We leverage this method to qualitatively verify that the features used during classification are consistent with what we expect the model to have learned. For this evaluation, we identify pleural effusion and splenomegaly as phenotypes with features that are well understood and can be observed in 2D slices. We select the presented volumes such that the phenotype prediction is above the decision boundary determined by AUROC curve analysis, to ensure that we are explaining a positive prediction. 

\begin{figure}[h!]
\centering
\includegraphics[width=1.0\textwidth]{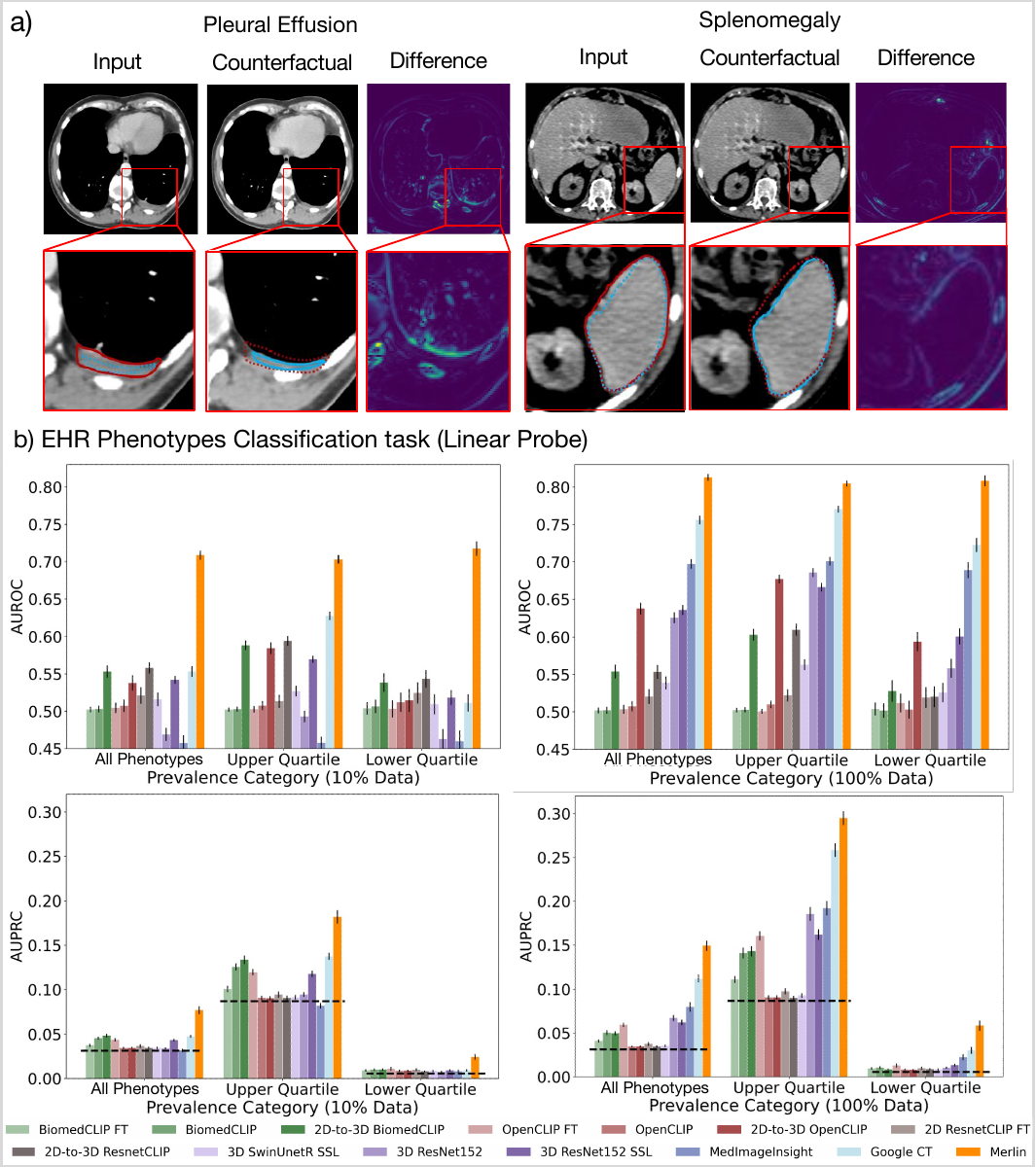}
\caption{\textit{Phenotype classification extended.} (a) Counterfactual analyses of pleural effusion classification (left; image from TCIA~\cite{clark2013cancer}) and splenomegaly classification (image from our internal test set). We annotate the zoomed in images by outlining the pathologies. The red lines border pathologies in the original images. The blue lines border pathologies in the counterfactual images. Counterfactual outlines are drawn over the original images with dotted lines and the original image outlines are also drawn over the counterfactual images with dotted lines. This allows comparing the size and shape of the pathologies between the original images and the counterfactuals, indicating that Merlin is indeed using appropriate features for image classification. (b) Average AUROC (second row) and AUPRC (third row) at 10\% and 100\% pretraining on the EHR phenotype classification task. Merlin pretraining results in consistently improved performance between the few-shot and fully-supervised data regimes compared to other baselines.}
\label{fig:phecode_classification_extended}
\end{figure}

\underline{Results:} In Figure \ref{fig:phecode_classification_extended}e, we apply the latent shift counterfactual method~\cite{Cohen2021gif} to qualitatively investigate the image features that Merlin uses to perform image classification. We examine an instance of ``pleural effusion'' (left) where the effusion is localized to the left lung and is reduced in the counterfactual, indicating that Merlin is leveraging the features we expect. We also examine ``splenomegaly'' classification (right). We observe that the size of the spleen is reduced in the counterfactual relative to the original image, adding credence to the validity of imaging features that Merlin uses for image classification.

\underline{\textit{EHR Phenotypes Task}} 

Aside from the results provided in Section \ref{result:beyond_merlin}, 2D and 2D-to-3D VLMs lifted offered benefits in data scarce settings with 10\% training data. 

\noindent\textbf{2D VLM baselines:} For the EHR phenotype classification task, the performance between finetuned and non-finetuned OpenCLIP and BiomedCLIP models remained similar, with a slight 2.6\% increase in AUROC and 4.9\% increase in AUPRC observed for all phenotypes when finetuning was applied given 10\% of the training data (Figure \ref{fig:phecode_classification_extended}b). In contrast, ResnetCLIP achieved a more significant 11.0\% AUROC and 13.0\% AUPRC improvement compared to finetuned OpenCLIP and BiomedCLIP at 10\% training data setting (Figure \ref{fig:phecode_classification_extended}b). These results suggest that finetuning provides a performance advantage in few-shot settings, specifically for 2D VLMs. This highlights that while finetuning 2D VLMs can enhance performance in limited data settings, their overall capacity to capture the volumetric data representations necessary for both tasks is limited.

\noindent\textbf{2D-to-3D lifted VLM baselines:} Larger performance gains were observed on the EHR phenotypes classification task, with a 6.8\% increase in AUROC and a 15.0\% increase in AUPRC for all phenotypes in the 10\% training data setting (Figure \ref{fig:phecode_classification_extended}b). These results suggest that, similar to 2D VLMs, 2D-to-3D lifted VLMs offer advantages in few-shot settings. 

\FloatBarrier
\clearpage

\subsection{Cross-modal Retrieval}

\begin{table*}[h!]
\small
\begin{center}
\setlength{\tabcolsep}{5.6pt}
\begin{tabular}{lllllcccccc}
\toprule
\textbf{$Task$}
& \textbf{$Method$}
& \multicolumn{3}{c}{\textbf{Recall@1}}
& \multicolumn{3}{c}{\textbf{Recall@8}}
\\
\cmidrule(l{3pt}r{0pt}){3-5}
\cmidrule(l{3pt}r{0pt}){6-8}
&
& \small N=32
& \small N=64
& \small N=128
& \small N=32
& \small N=64
& \small N=128


\\ 
\midrule
\small Img$\rightarrow$F & \small OpenCLIP & .030 & .016 & .009 & .243 & .125 & .062 \\ 
& \small BioMedCLIP & \cellcolor[HTML]{EFEFEF}.040 & \cellcolor[HTML]{EFEFEF}.021 & \cellcolor[HTML]{EFEFEF}.010 & \cellcolor[HTML]{EFEFEF}.298 & \cellcolor[HTML]{EFEFEF}.156 & \cellcolor[HTML]{EFEFEF}.083 
\\
& \small Merlin & \cellcolor[HTML]{C0C0C0}\textbf{.780} & \cellcolor[HTML]{C0C0C0}\textbf{.696} & \cellcolor[HTML]{C0C0C0}\textbf{.608} & \cellcolor[HTML]{C0C0C0}\textbf{.989} & \cellcolor[HTML]{C0C0C0}\textbf{.968} & \cellcolor[HTML]{C0C0C0}\textbf{.927} \\

\midrule
\small F$\rightarrow$Img & \small OpenCLIP & .033 & .017 & .009 & .250 & .125 & .061 \\ 
& \small BioMedCLIP & \cellcolor[HTML]{EFEFEF}.044 & \cellcolor[HTML]{EFEFEF}.021 & \cellcolor[HTML]{EFEFEF}.012 & \cellcolor[HTML]{EFEFEF}.306 & \cellcolor[HTML]{EFEFEF}.156 & \cellcolor[HTML]{EFEFEF}.079 
\\
& \small Merlin & \cellcolor[HTML]{C0C0C0}\textbf{.776} & \cellcolor[HTML]{C0C0C0}\textbf{.687} & \cellcolor[HTML]{C0C0C0}\textbf{.594} & \cellcolor[HTML]{C0C0C0}\textbf{.988} & \cellcolor[HTML]{C0C0C0}\textbf{.965} & \cellcolor[HTML]{C0C0C0}\textbf{.920} \\

\midrule
\small Img$\rightarrow$I & \small OpenCLIP & .030 & .016 & \cellcolor[HTML]{EFEFEF}.010 & .256 & .128 & .061 \\ 
& \small BioMedCLIP & \cellcolor[HTML]{EFEFEF}.036 & \cellcolor[HTML]{EFEFEF}.017 & .009 & \cellcolor[HTML]{EFEFEF}.273 & \cellcolor[HTML]{EFEFEF}.141 & \cellcolor[HTML]{EFEFEF}\cellcolor[HTML]{EFEFEF}.073 
\\
& \small Merlin & \cellcolor[HTML]{C0C0C0}\textbf{.352} & \cellcolor[HTML]{C0C0C0}\textbf{.253} & \cellcolor[HTML]{C0C0C0}\textbf{.174} & \cellcolor[HTML]{C0C0C0}\textbf{.796} & \cellcolor[HTML]{C0C0C0}\textbf{.663} & \cellcolor[HTML]{C0C0C0}\textbf{.532} \\

\midrule
\small I$\rightarrow$Img & \small OpenCLIP & .032 & .017 & .008 & .252 & .126 & .064 \\ 
& \small BioMedCLIP & \cellcolor[HTML]{EFEFEF}.046 & \cellcolor[HTML]{EFEFEF}.024 & \cellcolor[HTML]{EFEFEF}.012 & \cellcolor[HTML]{EFEFEF}.322 & \cellcolor[HTML]{EFEFEF}.169 & \cellcolor[HTML]{EFEFEF}.081 
\\
& \small Merlin & \cellcolor[HTML]{C0C0C0}\textbf{.384} & \cellcolor[HTML]{C0C0C0}\textbf{.277} & \cellcolor[HTML]{C0C0C0}\textbf{.194} & \cellcolor[HTML]{C0C0C0}\textbf{.854} & \cellcolor[HTML]{C0C0C0}\textbf{.706} & \cellcolor[HTML]{C0C0C0}\textbf{.564} \\
\bottomrule
\end{tabular}
\end{center}
\caption{\textit{Cross-modality retrieval.} We compare performance of OpenCLIP, BioMedCLIP, and Merlin across several settings: retrieving the correct findings section given an image (Img $\rightarrow$ F), retrieving the correct image given a findings section (F $\rightarrow$ Img), retrieving the correct impressions section given an image (Img $\rightarrow$ I), and retrieving the correct image given an impressions section (I $\rightarrow$ Img). We perform retrieval within pools of sizes N=32, N=64, and N=128.}
\label{table:retrieval_findings}
\end{table*}

\begin{table*}[h!]
\small
\begin{center}
\setlength{\tabcolsep}{5.6pt}
\begin{tabular}{llllcccccc}
\toprule
\textbf{$Task$}
& \textbf{$Init$}
& \textbf{$Labels$}
& \textbf{$Split$}
& \multicolumn{3}{c}{\textbf{Recall@1}}
& \multicolumn{3}{c}{\textbf{Recall@8}}
\\
\cmidrule(l{3pt}r{0pt}){5-7}
\cmidrule(l{3pt}r{0pt}){8-10}
&
&
& \textbf{$Text$}
& \small N=32
& \small N=64
& \small N=128
& \small N=32
& \small N=64
& \small N=128


\\ 
\midrule
\small Img$\rightarrow$F & I3D & Report & \cmark & .778 & .692 & .598 & \cellcolor[HTML]{C0C0C0}\textbf{.989} & .967 & .921 \\
& I3D & Staged & \xmark & .654 & .547 & .449 & .967 & .920 & .844 \\
& I3D & Staged & \cmark & .672 & .561 & .457 & .972 & .925 & .848 \\
& I3D & MTL & \xmark & \cellcolor[HTML]{C0C0C0}\textbf{.812} & \cellcolor[HTML]{C0C0C0}\textbf{.726} & \cellcolor[HTML]{C0C0C0}\textbf{.639} & .988 & \cellcolor[HTML]{C0C0C0}\textbf{.969} & \cellcolor[HTML]{C0C0C0}\textbf{.937} \\
& Rand & MTL & \cmark & .690 & .583 & .468 & .978 & .937 & .867 \\
& I3D & MTL & \cmark & \cellcolor[HTML]{EFEFEF}.780 & \cellcolor[HTML]{EFEFEF}.696 & \cellcolor[HTML]{EFEFEF}.608 & \cellcolor[HTML]{C0C0C0}\textbf{.989} & \cellcolor[HTML]{EFEFEF}.968 & \cellcolor[HTML]{EFEFEF}.927 \\

\midrule
\small F$\rightarrow$Img & I3D & Report & \cmark & .775 & .686 & .584 & .991 & \cellcolor[HTML]{C0C0C0}\textbf{.969} & .921 \\
& I3D & Staged & \xmark & .646 & .539 & .434 & .965 & .913 & .841 \\
& I3D & Staged & \cmark & .664 & .555 & .445 & .970 & .923 & .841 \\
& I3D & MTL & \xmark & \cellcolor[HTML]{C0C0C0}\textbf{.801} & \cellcolor[HTML]{C0C0C0}\textbf{.718} & \cellcolor[HTML]{C0C0C0}\textbf{.626} & \cellcolor[HTML]{C0C0C0}\textbf{.988} & .968 & \cellcolor[HTML]{C0C0C0}\textbf{.933} \\
& Rand & MTL & \cmark & .683 & .571 & .455 & .980 & .940 & .869 \\
& I3D & MTL & \cmark & \cellcolor[HTML]{EFEFEF}.776 & \cellcolor[HTML]{EFEFEF}.687 & \cellcolor[HTML]{EFEFEF}.594 & \cellcolor[HTML]{C0C0C0}\textbf{.988} & \cellcolor[HTML]{EFEFEF}.965 & \cellcolor[HTML]{EFEFEF}.920 \\

\midrule
\small Img$\rightarrow$I & I3D & Report & \cmark & .364 & \cellcolor[HTML]{EFEFEF}.265 & \cellcolor[HTML]{EFEFEF}.187 & \cellcolor[HTML]{C0C0C0}\textbf{.812} & \cellcolor[HTML]{C0C0C0}\textbf{.681} & \cellcolor[HTML]{C0C0C0}\textbf{.549} \\
& I3D & Staged & \xmark & .307 & .220 & .159 & .737 & .590 & .449 \\
& I3D & Staged & \cmark & .328 & .228 & .163 & .780 & .634 & .500 \\
& I3D & MTL & \xmark & \cellcolor[HTML]{C0C0C0}\textbf{.372} & \cellcolor[HTML]{C0C0C0}\textbf{.275} & \cellcolor[HTML]{C0C0C0}\textbf{.196} & \cellcolor[HTML]{EFEFEF}.799 & \cellcolor[HTML]{EFEFEF}.667 & \cellcolor[HTML]{EFEFEF}.543 \\
& Rand & MTL & \cmark & .288 & .202 & .131 & .740 & .592 & .453 \\
& I3D & MTL & \cmark & \cellcolor[HTML]{EFEFEF}.352 & .253 & .174 & .796 & .663 & .532 \\

\midrule
\small I$\rightarrow$Img & I3D & Report & \cmark & .382 & .274 & .192 & \cellcolor[HTML]{EFEFEF}.850 & \cellcolor[HTML]{C0C0C0}\textbf{.709} & \cellcolor[HTML]{C0C0C0}\textbf{.574} \\
& I3D & Staged & \xmark & .324 & .234 & .161 & .770 & .616 & .490 \\
& I3D & Staged & \cmark & .348 & .246 & .168 & .811 & .672 & .523 \\
& I3D & MTL & \xmark & \cellcolor[HTML]{C0C0C0}\textbf{.400} & \cellcolor[HTML]{C0C0C0}\textbf{.294} & \cellcolor[HTML]{C0C0C0}\textbf{.216} & .817 & .698 & \cellcolor[HTML]{EFEFEF}.568 \\
& Rand & MTL & \cmark & .289 & .200 & .127 & .779 & .613 & .460 \\
& I3D & MTL & \cmark & \cellcolor[HTML]{EFEFEF}.384 & \cellcolor[HTML]{EFEFEF}.277 & \cellcolor[HTML]{EFEFEF}.194 & \cellcolor[HTML]{C0C0C0}\textbf{.854} & \cellcolor[HTML]{EFEFEF}.706 & .564 \\
\bottomrule
\end{tabular}
\end{center}
\caption{\textit{Cross-modality retrieval ablation study.} We compare retrieval performance across three axes of weight initializations, methods for incorporating EHR and radiology reports into training, and splitting or using the full findings during training.}
\label{table:retrieval_findings_ablation}
\end{table*}

\begin{figure}[h!]
\centerline{\includegraphics[width=1.0\textwidth]{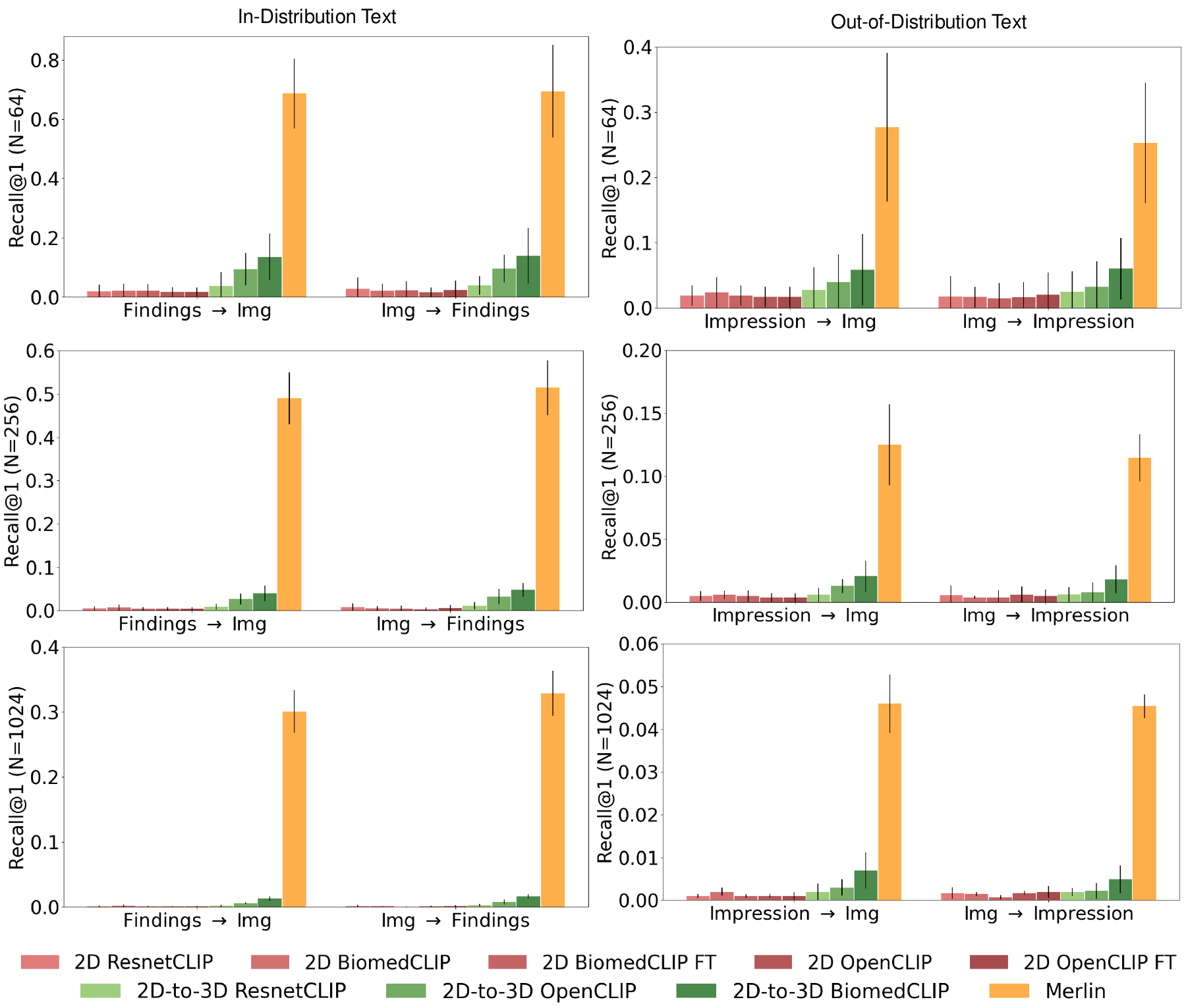}}
\caption{\textit{Retrieval extended.} Retrieval performance across CLIP-style models with pools of 64, 256, and 1024. The findings are in-distribution text, consistent with the data used during model training, while the impressions are out-of-distribution and shorter in length (Main paper, Figure 4b). All models, except for 2D BiomedCLIP and OpenCLIP, would have encountered the findings during training. Merlin maintains robust performance compared to other baselines.}
\label{fig:retrieval_extended}
\end{figure}

We extended the retrieval experiment presented in the main paper (Figure \ref{fig:retrieval}c) to include larger pools of data, specifically N = 64, 256, and 1024 (Figure \ref{fig:retrieval_extended}). Retrieval was performed on both the findings, which represent in-distribution text consistent with the data used during model training, and the impressions, which are out-of-distribution text and shorter in length (Figure \ref{fig:retrieval}b). We compared the performance of 2D VLM models and 2D-to-3D lifted VLMs to Merlin. Notably, all models except for 2D BiomedCLIP and OpenCLIP would have encountered the findings during training. These experiments offer insight into how medical VLMs may scale to larger clinical retrieval tasks.

\underline{Results:} Generally, as anticipated, we observe a decrease in retrieval performance across baselines as the pool size increases, but the relative retrieval performance remains consistent (Figure \ref{fig:retrieval_extended}). Merlin outperforms the 2D baselines, showing orders of magnitude better performance, and is considerably more effective than the 2D-to-3D lifted methods. Even for retrieval out of a large pool size of 1,024, Merlin achieves a recall of over 0.3, indicating that even with over a thousand examples to choose from, the correct example is chosen one in three times. In comparison, the 2D baselines consistently perform poorly on the retrieval tasks. Among the 2D-to-3D lifted methods, BiomedCLIP outperforms other lifted strategies for both in-distribution and out-of-distribution text and demonstrate slightly better performance than random results. 

\FloatBarrier
\clearpage

\subsection{Future Disease Prediction}

\begin{table*}[h!]
\setlength{\tabcolsep}{4.5pt}
\scriptsize
\begin{center}
{
\begin{tabular}{llllccccccc}
\toprule
\textbf{$Encoder$}
& \textbf{$Init$}
& \textbf{$Labels$}
& \textbf{$\% Tr$}
& \textbf{CKD}
& \textbf{DM}
& \textbf{HTN}
& \textbf{IHD}
& \textbf{CVD}
& \textbf{OST}
& \textbf{Average} 
\vspace{2pt} \\
& & & 10\% & 14/46 & 9/46 & 11/34 & 9/49 & 20/52 & 10/47 \\
& & & 100\% & 90/513 & 80/474 & 111/404 & 69/518 & 136/504 & 68/527 \\
\midrule
Swin Transformer & - & - & 10\% & .46 [.40, .50] & \cellcolor[HTML]{EFEFEF}.70 [.66, .75] & .43 [.39, .48] & .53 [.49, .57] & .53 [.50, .57] & .56 [.51, .61] & .54 [.52, .55] \\
ResNet152 & - & - & 10\% & .53 [.48, .58] & .66 [.61, .71] & .60 [.55, .64] & .49 [.45, .54] & .58 [.54, .62] & .50 [.44, .56] & .56 [.54, .58] \\
$\downarrow$ & I3D & - & 10\% & \cellcolor[HTML]{EFEFEF}.72 [.67, .76] & \cellcolor[HTML]{C0C0C0}\textbf{.72 [.67, .76]} & .67 [.63, .71] & .67 [.63, .71] & .67 [.64, .71] & \cellcolor[HTML]{EFEFEF}.66 [.61, .71] & \cellcolor[HTML]{EFEFEF}.68 [.67, .70] \\
& I3D & EHR & 10\% & .58 [.53, .63] & .64 [.59, .69] & .47 [.42, .51] & \cellcolor[HTML]{C0C0C0}\textbf{.75 [.71, .78]} & \cellcolor[HTML]{EFEFEF}.69 [.65, .72] & .62 [.57, .67] & .62 [.60, .64] \\
(Merlin) & I3D & MTL & 10\% & \cellcolor[HTML]{C0C0C0}\textbf{.74 [.70, .78]} & \cellcolor[HTML]{EFEFEF}.70 [.66, .75] & \cellcolor[HTML]{C0C0C0}\textbf{.69 [.65, .73]} & \cellcolor[HTML]{EFEFEF}.70 [.67, .74] & \cellcolor[HTML]{C0C0C0}\textbf{.73 [.69, .76]} & \cellcolor[HTML]{C0C0C0}\textbf{.69 [.65, .73]} & \cellcolor[HTML]{C0C0C0}\textbf{.71 [.69, .72]} \\
\midrule
Swin Transformer & - & - & 100\% & .55 [.50, .59] & \cellcolor[HTML]{EFEFEF}.73 [.68, .77] & .61 [.57, .65] & .52 [.48, .56] & .54 [.49, .57] & .60 [.55, .66] & .59 [.57, .61]  \\
ResNet152 & - & - & 100\% & .63 [.58, .67] & \cellcolor[HTML]{C0C0C0}\textbf{.74 [.69, .78]} & .65 [.61, .69] & .65 [.61, .69] & .57 [.54, .61] & .60 [.55, .66] & .64 [.62, .66] \\
$\downarrow$ & I3D & - & 100\% & .74 [.70, .77] & \cellcolor[HTML]{C0C0C0}\textbf{.74 [.70, .78]} & .71 [.67, .75] & .68 [.64, .72] & .68 [.64, .71] & .74 [.69, .78] & .71 [.70, .73] \\
& I3D & EHR & 100\% & \cellcolor[HTML]{EFEFEF}.76 [.73, .80] & .72 [.68, .76] & \cellcolor[HTML]{EFEFEF}.74 [.70, .77] & \cellcolor[HTML]{EFEFEF}.74 [.70, .78] & \cellcolor[HTML]{EFEFEF}.73 [.69, .76] & \cellcolor[HTML]{EFEFEF}.68 [.64, .73] & \cellcolor[HTML]{EFEFEF}.73 [.71, .74] \\
(Merlin) & I3D & MTL & 100\% & \cellcolor[HTML]{C0C0C0}\textbf{.77 [.74, .81]} & .72 [.68, .76] & \cellcolor[HTML]{C0C0C0}\textbf{.75 [.72, .79]} & \cellcolor[HTML]{C0C0C0}\textbf{.76 [.72, .79]} & \cellcolor[HTML]{C0C0C0}\textbf{.74 [.71, .77]} & \cellcolor[HTML]{C0C0C0}\textbf{.80 [.76, .84]} & \cellcolor[HTML]{C0C0C0}\textbf{.76 [.74, .77]} \\
\bottomrule
\end{tabular}
}
\end{center}
\caption{\textit{Multi-disease 5-year prediction.} We fine-tune Merlin for 5-year disease prediction. All data used in this evaluation, including train, val, and test splits, are held out from pretraining.}
\label{table:disease_prediction_5yr}
\end{table*}

\FloatBarrier

\subsection{Report Generation}

\begin{table*}[h!]
\setlength{\tabcolsep}{3.5pt}
\small
\begin{center}
\begin{tabular}{lcccccccc}
\toprule
& \multicolumn{2}{c}{\textbf{BLEU $\uparrow$}} & \multicolumn{2}{c}{\textbf{ROUGE-2 $\uparrow$}} & \multicolumn{2}{c}{\textbf{BERT $\uparrow$}} & \multicolumn{2}{c}{\textbf{RadGraph-F1 $\uparrow$}} \\
\cmidrule(lr){2-3} \cmidrule(lr){4-5} \cmidrule(lr){6-7} \cmidrule(lr){8-9}
\textbf{$Section$} & RadFM & Merlin & RadFM & Merlin & RadFM & Merlin & RadFM & Merlin \\
\midrule
Lower thorax & .001 & .019 & .070 & .332 & .406 & .615 & .020 & .319 \\
Liver and biliary tree & .001 & .269 & .025 & .389 & .328 & .641 & .080 & .380 \\
Gallbladder & .000 & .006 & .006 & .632 & .534 & .851 & .152 & .721 \\
Spleen & .000 & .002 & .004 & .710 & .382 & .853 & .283 & .805 \\
Pancreas & .000 & .001 & .010 & .700 & .447 & .849 & .091 & .748 \\
Adrenal glands & .006 & .030 & .067 & .882 & .490 & .942 & .106 & .879 \\
Kidneys and ureters & .005 & .269 & .040 & .385 & .368 & .654 & .091 & .387 \\
Gastrointestinal tract & .001 & .013 & .037 & .152 & .398 & .531 & .092 & .167\\
Peritoneal cavity & .000 & .206 & .005 & .390 & .387 & .702 & .050 & .335 \\
Pelvic organs & .000 & .233 & .009 & .358 & .328 & .656 & .036 & .432 \\
Vasculature & .000 & .026 & .004 & .485 & .232 & .748 & .006 & .548 \\
Lymph nodes & .003 & .023 & .119 & .601 & .502 & .775 & .031 & .542 \\
Musculoskeletal & .001 & .046 & .018 & .303 & .449 & .689 & .008 & .293 \\
\midrule
Full report & .000 & .102 & .011 & .262 & .224 & .588 & .008 & .293 \\
\bottomrule
\end{tabular}
\caption{\textit{Radiology report generation.} We compare Merlin and RadFM for generating radiology report sections corresponding to various anatomies, as well as the full findings.}
\label{table:perceptual}
\end{center}
\end{table*}

\begin{figure}[h!]
\centering
\includegraphics[width=1.0\textwidth]{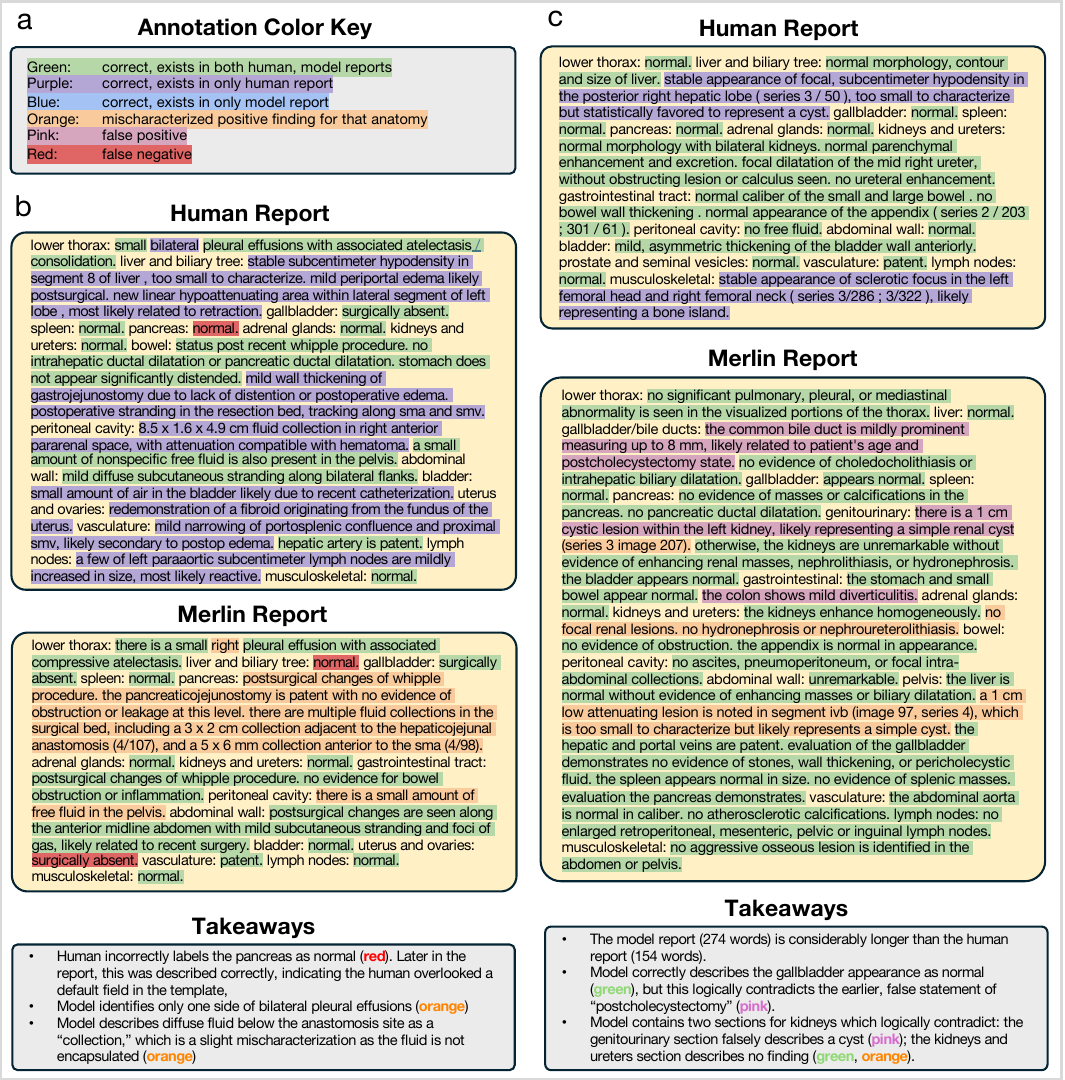}
\caption{\textit{Radiology report generation.} (a) As shown in the annotation color key, we annotate individual phrases to be correct, mischaracterized, false positive, or false negative. (b - c) We provide dense annotations of two sets of human and Merlin generated reports.}
\label{fig:report_generation_supp}
\end{figure}

\FloatBarrier

\clearpage
\section{Additional Comparisons}

\subsection{Additional Alternative Architecture Comparison}
\label{suppl:embedding_comparison}

\paragraph{Full-Finetuning:} We extend the results from Figure \ref{fig:baseline_experiments} and present full fine-tuning to enable a fair comparison between Merlin and alternative architecture baselines.

\underline{Methods:} Here, we describe the evaluation tasks, which involve finetuning both the image encoder and classification heads. For the 3D vision-only models, the embedding dimensions were 768 for SwinUNETR and 2048 for I3D Resnet152. For the 2D and the 2D-to-3D VLMs, the embedding dimensions were 768 for BiomedCLIP, 2048 for ResnetCLIP, and 1024 for OpenCLIP.

\noindent\textit{Findings-based disease classification task:} The findings-based disease classification task was adapted from the zero-shot findings classification task (Section \ref{method:zero_shot}), where the 30 findings originally used for zero-shot evaluation were reformulated for supervised learning. We parsed the radiology reports and categorized each finding as positive, negative, or missing from the report. While most conventional strategies involve discarding missing labels during training, we adopted a more inclusive strategy to use all our data. We framed the task as a multilabel classification task, introducing thirty separate binary heads, one for each finding. We masked the loss for subjects missing a specific finding before backpropogation\cite{xue2024ai, blankemeier2022opportunistic}. This approach standardizes the finetuning of image encoders across various models: vision-only foundation models, 2D vision-language models, 2D finetuned vision-language models, 2D-to-3D lifted vision-language models, and Merlin. We evaluated the performance of these baselines using 10\% and 100\% of the training data over 10 epochs for thoroughness.

\noindent\textit{Phenotype classification} The EHR phenotype classification task (Section \ref{method:phenotype_cls}) required minimal modification to the model architectures, by including linear classification heads for the 1692 phenotype tasks. For the vision-language models finetuned with MTL, the EHR linear layer from the MTL pretraining was already included, allowing for straightforward finetuning. For the vision-only and original 2D vision-language models (without finetuning), we appended a linear classification head to the image encoder. Similar to the findings-based disease classification task, we fully finetuned (image encoder plus linear classification heads) the models on 10\% and 100\% of the training data to assess classification performance across different dataset sizes.

\begin{figure}[h!]
\centerline{\includegraphics[width=1.0\textwidth]{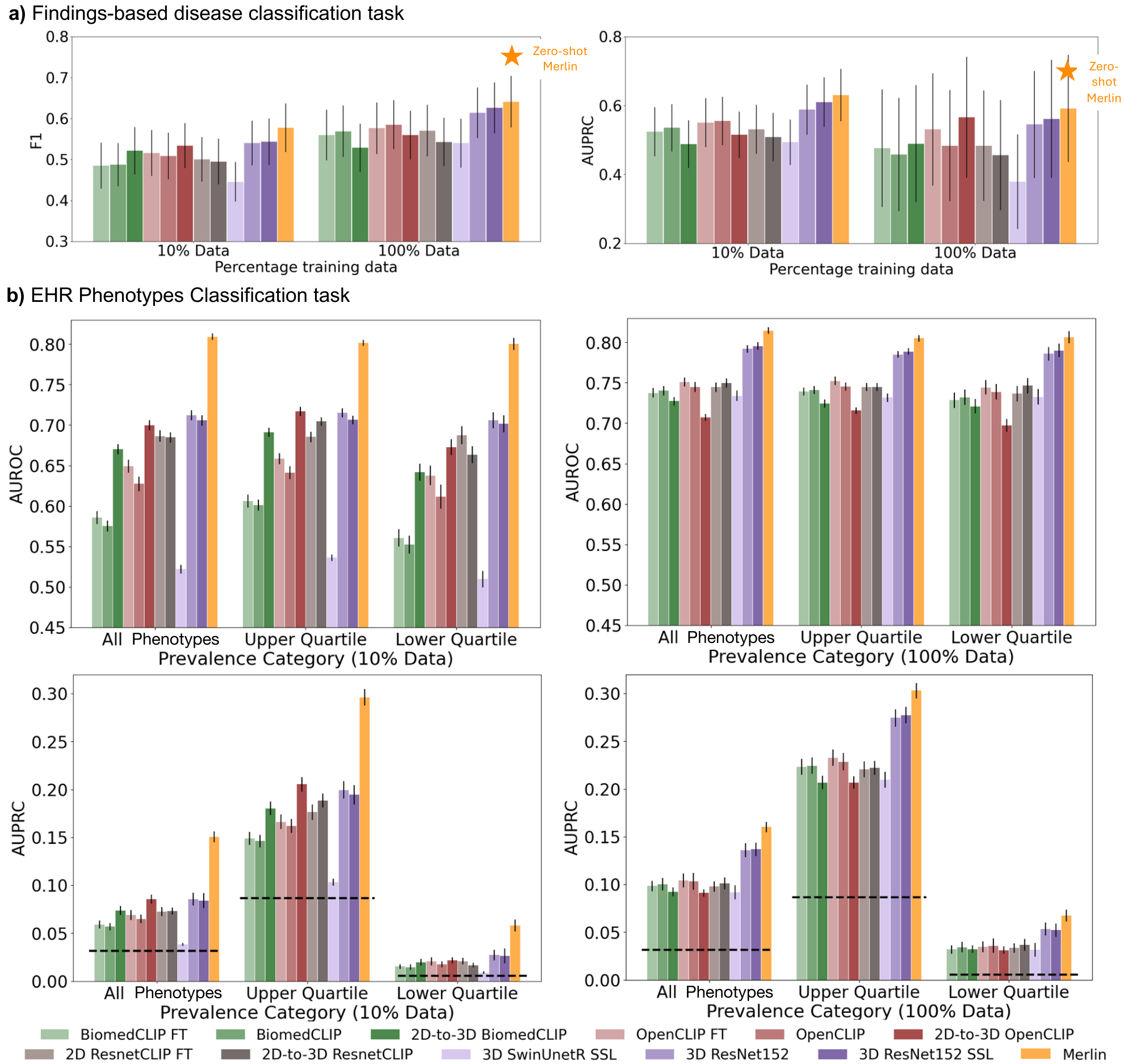}}
\caption{\textit{Alternative Architecture Baselines compared to Merlin (Full Finetuning).} Alternative architecture baselines versus Merlin, where each baselines is fully finetuned (image encoder plus classification head). Average F1 (left chart) and AUPRC (right chart) on 10\% and 100\% pretraining data for the findings-based disease classification task. (b) Average AUROC (second row) and AUPRC (third row) at 10\% and 100\% pretraining on the EHR phenotype classification task. Merlin pretraining results in consistently improved performance between the few-shot and fully-supervised data regimes compared to other baselines.}
\label{suppl_fig:full_finetuning}
\end{figure}

\underline{Results:} \noindent\textit{2D VLMs:} The 2D VLMs demonstrated relatively consistent performance between both finetuning and no finetuning scenarios. Specifically, finetuned OpenCLIP and BiomedCLIP exhibited an increase of 6.8\% in AUPRC with relatively consistent F1 scores compared to their non-finetuned counterparts on the findings-based disease classification task at the 100\% training data setting (Figure \ref{suppl_fig:full_finetuning}a). Similarly, finetuned 2D ResnetCLIP yielded similar performance to the other finetuned models. For example, when trained on 100\% of the data for the findings-based disease classification task, the model achieved an F1 score of 0.570 (95\% CI [0.505-0.631]), which was comparable to OpenCLIP (0.576, 95\% CI [0.511-0.636]) and BiomedCLIP (0.556, 95\% CI [0.496-0.620]). However, 2D VLMs underperformed compared to Merlin, with a 8.1\% lower F1 score for findings-based disease classification and 9.4\% lower AUROC on the EHR phenotypes task in the 100\% training data setting (Figure \ref{suppl_fig:full_finetuning}a-b). 

\noindent\textit{2D-to-3D lifted VLMs:} 2D-to-3D lifted VLMs demonstrated comparable performance to their 2D VLM baselines, with some improvements in specific scenarios. On the findings-based disease classification task, 2D-to-3D lifted VLMs performed similarly to their finetuned 2D VLM counterparts, with a 3.2\% increase in F1 score in the 10\% training data setting and a 1.4\% increase in AUPRC in the 100\% training data setting (Figure \ref{suppl_fig:full_finetuning}a). Performance among the 2D-to-3D lifted models varied. For instance, 2D-to-3D OpenCLIP achieved a 3.4\% increase in F1 score and a 19.0\% increase in AUPRC for the findings-based disease classification task compared to the other 2D-to-3D baselines (Figure \ref{suppl_fig:full_finetuning}a). In contrast, 2D-to-3D ResnetCLIP outperformed the other 2D-to-3D VLMs on the EHR phenotypes classification task, showing a 4.5\% increase in AUROC and a 10.0\% increase in AUPRC.

Despite these improvements, 2D-to-3D lifted VLMs underperformed relative to Merlin. Specifically, Merlin had 18.0\% higher F1 scores on the findings-based disease classification task and a 12.0\% higher AUROC on the EHR phenotype classification task compared to the 2D-to-3D lifted VLMs (Figure \ref{suppl_fig:full_finetuning}a-b). 

\noindent\textit{3D vision-only models:} 3D models trained from scratch or with SSL pretraining had a 8.1\% decrease in F1 score on the findings-based disease classification task and a 5.3\% decrease in AUROC on the EHR phenotypes task compared to Merlin in the 100\% training data setting  (Figure \ref{suppl_fig:full_finetuning}a-b). Among the 3D vision-only baselines, SwinUNETR underperformed 3D Resnet, with a 16.0\% decrease in F1 score on the findings-based disease classification task and a 9.1\% decrease in AUROC on the EHR phenotypes task. SSL pretraining modestly improved 3D Resnet performance, increasing the F1 score by 1.8\% and AUPRC by 3.1\% on the findings-based disease classification task (Figure \ref{suppl_fig:full_finetuning}a-b). However, these 3D Resnet models underperformed relative to Merlin, with a 3.5\% decrease in F1 score on the findings-based disease classification task and a 17.0\% decrease in AUPRC on the EHR phenotypes task (Figure \ref{suppl_fig:full_finetuning}a-b). Although 3D vision-only models outperformed the 2D and 2D-to-3D VLMs with a 5.7\% increase in F1 score on the findings-based disease classification task and 4.8\% increase in AUROC on the EHR phenotypes task, Merlin still achieved the best quantitative performance across all tasks (Figure \ref{suppl_fig:full_finetuning}a-b).

\begin{figure}[h!]
\centering
\includegraphics[width=1.0\textwidth]{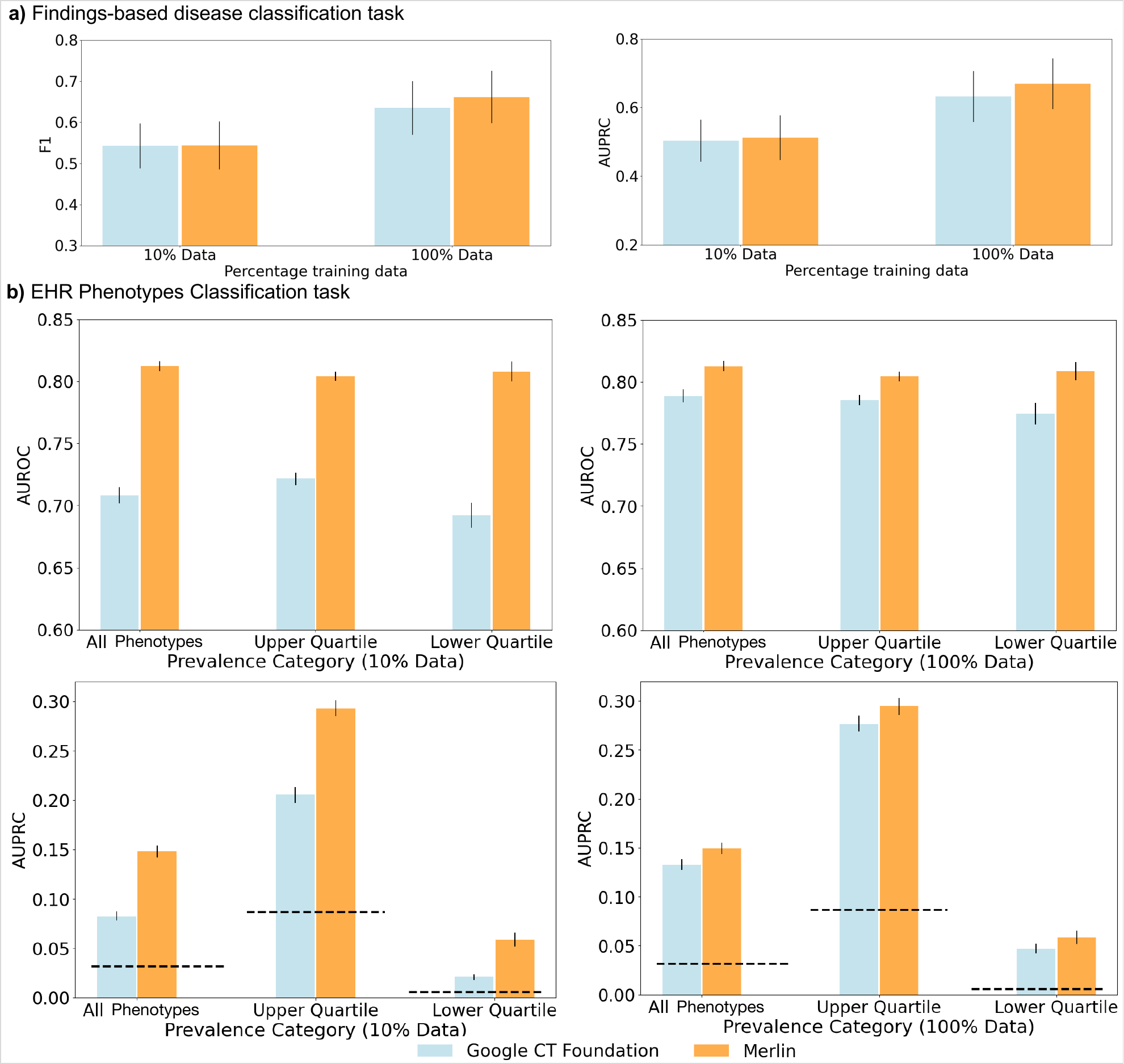}
\caption{\textit{MLP Embedding Comparison.} Comparison of 3D embeddings between Google CT Foundation and Merlin. To clarify, the evaluation models were trained using embeddings only, as Google CT Foundation produces embeddings exclusively. Lightweight multi-layer perceptrons were then trained for both classification tasks. All hyperparameters were kept consistent during training. (a) Average F1 (left chart) and AUPRC (right chart) on 10\% and 100\% training data for the findings-based disease classification task. (b) Average AUROC (second row) and AUPRC (third row) at 10\% and 100\% pretraining.}
\label{suppl_fig:embedding_comparison}
\end{figure}

\textbf{MLP Experiments:} We further evaluate the embeddings from a recent 3D CT embedding model from Google~\cite{kiraly2024medical}. We find that Merlin’s embeddings are more information-rich compared to those produced by the Google 3D CT vision-language embedding model using multi-layer perceptrons (Figure \ref{suppl_fig:embedding_comparison}).

\underline{Method (Google CT Pre-processing):} We further compare Merlin with Google CT Foundation on the findings-based disease and EHR phenotype classification task. CT Foundation is a recently released foundation model that generates embeddings from CT volumes~\cite{yang2024advancing}. We do note that the CT Foundation model presents challenges due to limited reproducibility: the model weights, training methodology, and data pre-processing details have not been disclosed. Interaction with the model is restricted to an API, which outputs a 1,408-dimensional 32-bit embedding vector for each CT volume. To use the API, we provide Google with a Google Cloud Bucket containing our 25,528 NIfTI files, and the API returns the corresponding 1,408-dimensional vectors as pickle files. We agree that there is limited methodological detail and differences in preprocessing could influence the results. To enable a fair comparison, we further processed each abdominal CT scan (a total of 25,528 volumes) to generate 2048-dimensional Merlin embedding vectors. Embeddings from both models (Merlin and Google CT Foundation) were then used as inputs to lightweight multi-layer perceptrons for findings-based disease and EHR phenotype classification tasks, with all hyperparameters kept consistent during evaluation. This approach allowed us to assess which 3D CT foundation model produces the most information-rich embedding vectors for these tasks.

\underline{Results:} Compared to the Google CT Foundation, Merlin had 4.1\% higher F1 score and a 5.8\% higher AUPRC for the findings-based disease classification task (Figure \ref{suppl_fig:embedding_comparison}). Both models depicted similar performance in the 10\% training data setting for this tsk. On the EHR phenotyping task in the 100\% training data setting, Merlin had a 3.0\% higher AUROC and a 13.0\% higher AUPRC than Google CT Foundation (Figure \ref{suppl_fig:embedding_comparison}b). In the 10\% training data setting, Merlin had 14.7\% higher AUROC and a 79.0\% higher AUPRC. Overall, while Google CT Foundation served as a strong baseline, Merlin provided higher-quality embeddings for these evaluation tasks.

\paragraph{MedImageInsight Processing Details:}  MedImageInsight (Microsoft) is a multimodal CLIP-style model that leverages the DaViT architecture with UniCL as its objective function, providing embeddings for both medical images and text~\cite{codella2024medimageinsight}. Trained on 3D medical imaging data, including abdominal CT scans, this model produces image and text representations suitable for downstream tasks. To evaluate its performance on the Merlin internal test set, we applied MedImageInsight’s image embeddings to zero-shot findings classification and EHR phenotype prediction. Following the Merlin preprocessing pipeline, abdominal CT scans were resampled to 1.5 × 1.5 × 3 mm voxel spacing, reoriented to RAS, clipped to a Hounsfield range of –1000 to 1000, and center-cropped to 224 × 224 × 160 voxels. Each volume was decomposed into 160 axial slices, converted to PNG images, and processed by MedImageInsight. Slice-level embeddings were then aggregated using a median-based strategy to obtain a single representation per CT volume (see Methods~\cite{codella2024medimageinsight}).

\FloatBarrier

\clearpage
\subsection{Image data augmentation}

We evaluated the impact of image data augmentation for vision-language pretraining. We observed that augmentation strategies significantly reduced performance compared to Merlin, which was trained without data augmentation.

\underline{Methods:} We explored two variants of image data augmentation strategies that attempted to preserve the original image signal - one weak and one strong, inspired by nnUNet~\cite{isensee2021nnu} and prior work on optimal augmentation strategies for medical imaging~\cite{van2024exploring}. Below, we list the augmentations used and their probability of being applied per instance ($P$). The weak image augmentation involved: zoom scaling between 0.8 and 1.2 ($P=0.16$), Gaussian noise addition with a standard deviation of up to 0.1 ($P=0.1$), and gamma-based intensity adjustments ranging from 0.7 to 1.5 ($P=0.15$). The more aggressive augmentations included: rotations up to 15 degrees ($P=0.16$), zoom scaling between 0.6 and 1.4 ($P=0.16$), Gaussian noise addition with a standard deviation of up to 0.3 ($P=0.1$), gamma-based intensity adjustments ranging from 0.7 to 1.5 ($P=0.15$), and horizontal flips ($P=0.5$).  These models were trained from scratch, initialized with inflated Imagenet weights, and run for 100 epochs (same as Merlin). 

\begin{figure}[h!]
\centerline{\includegraphics[width=0.75\textwidth]{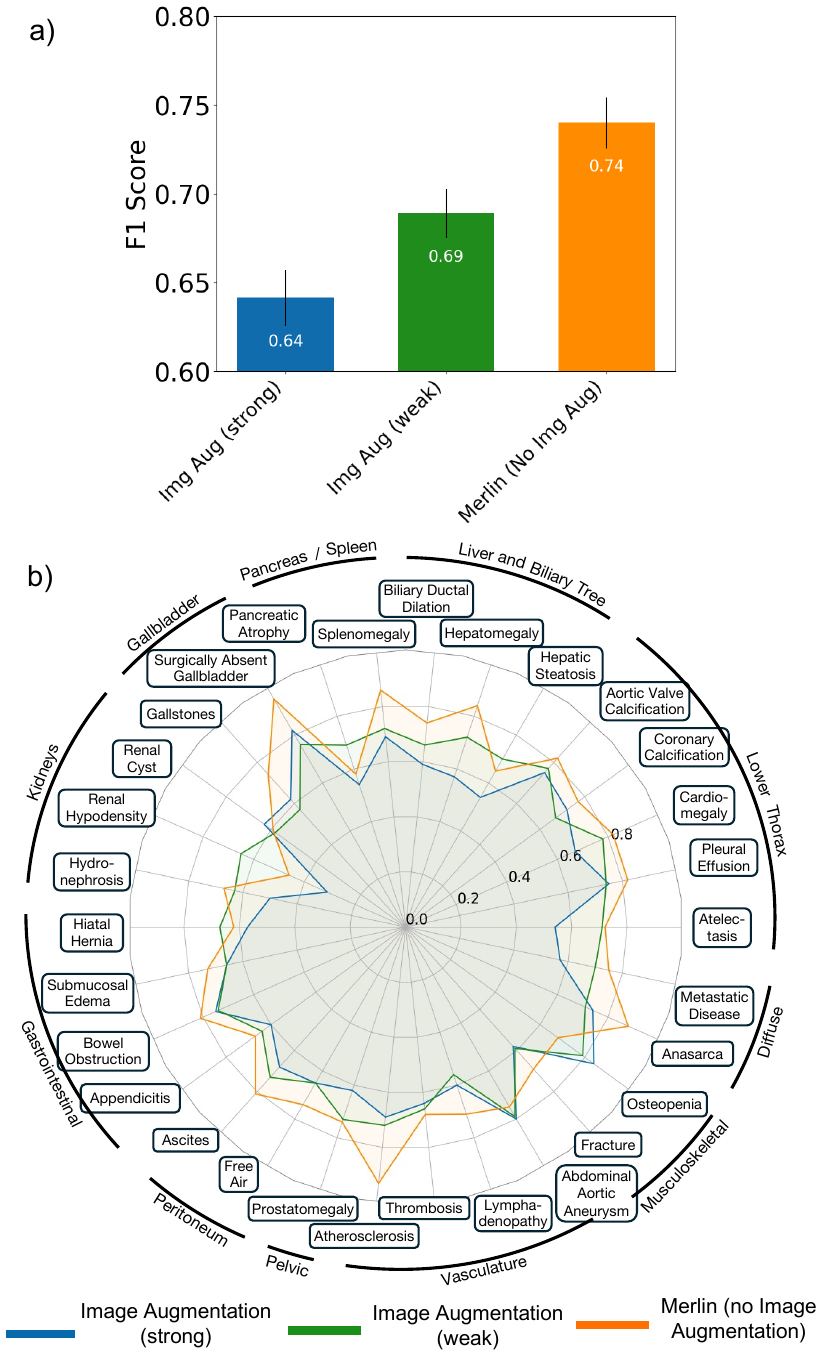}}
\caption{\textit{Image data augmentation experiments.} Image data augmentation experiments on zero-shot findings classification task. (a) Average F1 score for models trained with Merlin's architecture (I3D Resnet152; Clinical Longformer) with weak versus strong image data augmentation. Increased augmentation results in reduced performance. (b) Performance on models trained with various image data augmentation strategies, including Merlin, across 30 findings assessed on abdominal CT scans.}
\label{suppl_fig:data_aug}
\end{figure}

\underline{Results:} We evaluated the image data-augmented models on the zero-shot findings classification task and observed that augmentation significantly reduced performance compared to Merlin trained without data augmentation (Figure \ref{suppl_fig:data_aug}a). Specifically, for many diseases in the findings task, the use of augmentation strategies appeared to progressively degrade performance (Figure \ref{suppl_fig:data_aug}b). This result may initially seem counterintuitive, as existing literature suggests that image data augmentation improves CLIP performance~\cite{li2024scaling}. However, the same study also emphasized the importance of training CLIP models on high-quality datasets, showing that training on 60\% of the highest-quality data yielded nearly equivalent performance to using the full dataset. It is possible that CLIP-style training with abdominal CT images and radiology reports requires high-quality data that maintains precise image-text alignment. Image data augmentations may inadvertently alter critical elements of the abdominal CT images, causing misalignment with the corresponding radiology reports. This misalignment likely leads the model to learn noise rather than meaningful signal, reducing the effectiveness of CLIP pretraining (Figure \ref{suppl_fig:data_aug}). Furthermore, while data augmentations can introduce robustness in models trained on larger datasets commonly found in natural imaging, our comparatively smaller-scale medical training dataset may amplify the noise introduced by these augmentations rather than improving robustness. Scaling to larger datasets could mitigate this issue, potentially enhancing the benefits of augmentations and improving the overall effectiveness of CLIP pretraining. We therefore find that Merlin, trained without any image data augmentation, still outperforms variants using augmentation strategies. We hope that our open source dataset can inspire additional work on optimal augmentations and training strategies in the future.

\color{black}

\subsection{External Validation Experiments}

We expand on our external validation experiments (Section~\ref{result:external_val}), providing a summary of results and an explanation for observed discrepancies in dataset distributions.

\begin{figure}[h!]
\centerline{\includegraphics[width=1.0\textwidth]{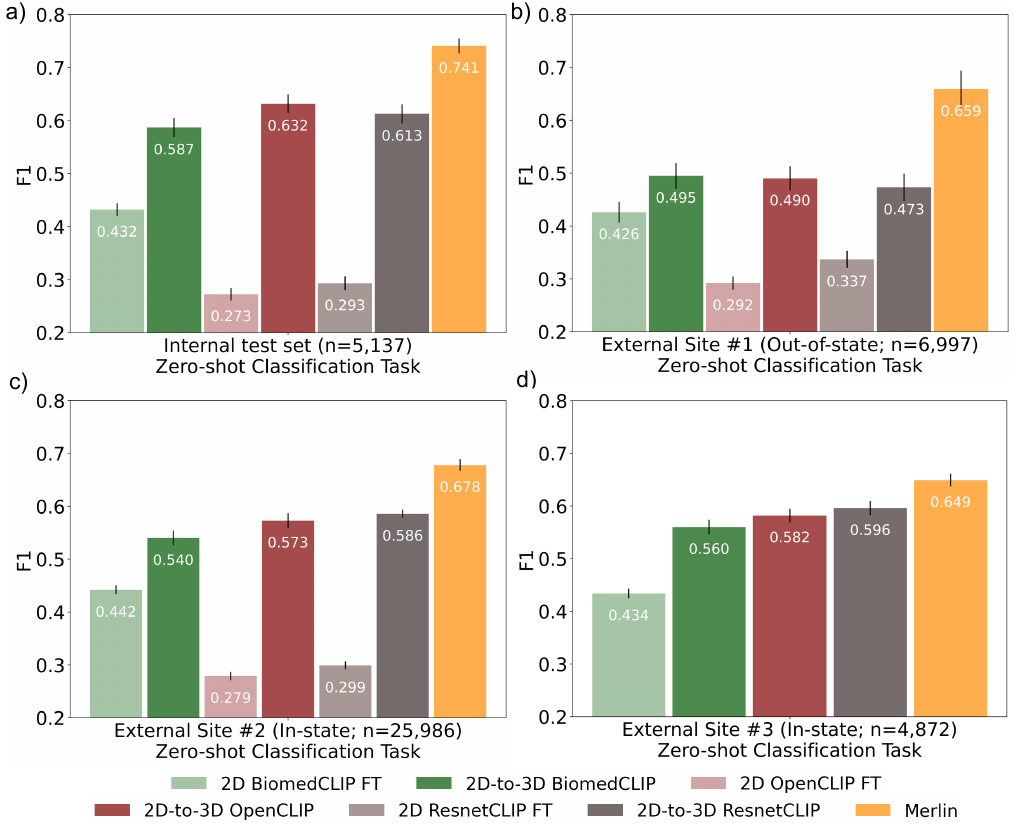}}
\caption{\textit{External Validation Experiments Summarized.} Merlin external validation on 37,885 abdominal CTs from three external sites and 5,137 internal CTs compared to alternative architecture baselines. We extend the internal and external validation (Figure \ref{fig:zero_shot}b) on abdominal CTs to include alternative architectures evaluated on: (a) the Merlin internal test set, (b) the External Site \#1, (c) External Site \#2, and (d) External Site \#3, using the zero-shot classification task and reporting F1 scores. Error bars represent 95\% confidence intervals. Across all baselines, Merlin consistently achieves the highest performance.}
\label{suppl_fig:external_val_summarized}
\end{figure}

\underline{Zero-shot findings classification:} We evaluate Merlin on the zero-shot findings classification task using abdominal CT scans from four separate datasets (Figure~\ref{suppl_fig:external_val_summarized}).

Specifically, we observe:

\begin{itemize}
    \item \textbf{Internal test set:} Merlin outperforms the second-best model, 2D-to-3D OpenCLIP, by 17.2\% (Figure ~\ref{suppl_fig:external_val_summarized}a).
    \item \textbf{Out-of-state site \#1:} All 2D-to-3D models experienced a 25.5\% drop in external performance compared to the internal test set. Merlin on the other hand, had a drop of 12.4\%. Additionally, Merlin outperformed the second-best model, 2D-to-3D BiomedCLIP, by 34.4\% (Figure ~\ref{suppl_fig:external_val_summarized}b).
    \item \textbf{In-state site \#2:} Merlin outperforms the second-best model, 2D-to-3D ResnetCLIP, by 15.7\% (Figure ~\ref{suppl_fig:external_val_summarized}c).
    \item \textbf{In-state site \#3:} Merlin outperforms the second-best model, 2D-to-3D ResnetCLIP, by 8.89\% (Figure ~\ref{suppl_fig:external_val_summarized}d).
\end{itemize}

\underline{External dataset distribution shifts:} We summarize dataset characteristics in Table ~\ref{tab:dataset_characteristics}. Across all external sites, the mean patient age is higher than in our internal test set, which may contribute to performance differences given that older populations can exhibit distinct disease prevalence and imaging characteristics. These cohorts exhibited differences in demographics and acquisition parameters, particularly age distributions, slice thickness, scanner manufacturer, and radiologist reporting patterns. This diversity enables us to assess Merlin's ability to generalize across varied clinical settings.

We note that External Site \#1 is an out-of-state dataset exhibiting distribution shifts not only at the patient, scanner, and imaging protocol level, but also in how radiologists interpret and report findings. In our training dataset, negative findings are documented using a standardized template across all radiologists, ensuring consistent phrasing. In contrast, radiologists in the external test set employ varied styles and terminology when reporting negative cases. Consequently, report text for normal organs is more homogeneous in the training data than in the external dataset. The slight performance differences observed on this external set are likely due to such reporting-level variations. Additionally, we observe that most CTs from External Site \#1 have slice thicknesses greater than 3 mm, which substantially degrades image quality by reducing spatial resolution and obscuring anatomical details critical for accurate findings detection. We also observe that every other alternative architecture baseline performs worse than random chance on Site \#1's dataset. Since all baselines shared identical pretraining, including the Clinical Longformer~\cite{li2022clinical} text encoder, Merlin’s superior performance can be attributed to its full 3D image encoder. The consistent, substantial gains demonstrate the quality of the Merlin model. 

For External site \#2,  performance differences to the internal test set are primarily driven by scanner-level variation. From Table ~\ref{tab:dataset_characteristics}, we observe that the internal test set has an approximately 60/40 split between GE Healthcare and Siemens scanners, whereas 95\% of scans in Site \#2 were acquired by Canon scanners. Different scanner manufacturers can introduce variations in reconstruction algorithms, resolution, noise levels, and contrast profiles~\cite{hammad2025explainable}. Such differences can alter CT scan appearance, making it more challenging for Merlin to align image and text representations under these distribution shifts.

For External Site \#3, we observe both protocol-level differences, particularly for kVp, and scaner-level differences. In the internal test set, most scans use either 100 or 120 kVp, whereas in External Site \#3, nearly all scans are acquired at 120 kVp (Table ~\ref{tab:dataset_characteristics}). Additionally, nearly all scanners at External Site \#3 are GE Healthcare, compared to a more balanced distribution of GE Healthcare and Siemens scanners in the internal test set (Table ~\ref{tab:dataset_characteristics}). Variations in kVp can influence image noise and exposure, especially in dense anatomical regions~\cite{chen2021variation}.

\begin{table}[h!]
\begin{center}
\resizebox{\textwidth}{!}{%
\begin{tabular}{l*{5}{|cc}}
\toprule
\textbf{Dataset Characteristics} 
& \multicolumn{2}{c}{\textbf{Train / Validation Set}} 
& \multicolumn{2}{c}{\textbf{Internal Test Set}} 
& \multicolumn{2}{c}{\textbf{Site \#1}}
& \multicolumn{2}{c}{\textbf{Site \#2}}
& \multicolumn{2}{c}{\textbf{Site \#3}} \\
\cmidrule(lr){2-3} \cmidrule(lr){4-5} \cmidrule(lr){6-7} \cmidrule(lr){8-9} \cmidrule(lr){10-11}
& \textbf{n} & \textbf{\%} 
& \textbf{n} & \textbf{\%} 
& \textbf{n} & \textbf{\%}
& \textbf{n} & \textbf{\%}
& \textbf{n} & \textbf{\%} \\
\midrule
\textbf{Abdominal CTs} & 20,332 & -- & 5,137 & -- & 6,997 & -- & 25,986 & -- & 4,872 & -- \\
\textbf{Age (mean)} & $54.1 \pm 18.9$ & -- & $54.5 \pm 19.1$ & -- &  $62.3 \pm 14.5$ & -- & $60.3 \pm 19.3$ & -- & $60.3 \pm 16.7$ & -- \\
\multicolumn{9}{l}{\textbf{Sex}} \\
\quad Female & 11,377 & 56.0 & 2,927 & 57.0 & 3,631 & 52.1 & 14,132 & 54.4 & 2,454 & 50.4 \\
\quad Male & 8,953 & 44.0 & 2,210 & 43.0 & 3,364 & 48.2 & 11,854 & 45.6 & 2,418 & 49.6 \\
\quad Other & 2 & 0.00 & 0 & 0.00 & 1 & 0.00 & 0 & 0.00 & 1 & 0.00 \\
\multicolumn{9}{l}{\textbf{kVp Count}} \\
\quad 100 kVp & 7,901 & 38.9 & 1,975 & 38.4 & 636 & 9.08 & 782 & 3.00 & 73 & 1.50 \\
\quad 120 kVp & 11,501 & 56.5 & 2,911 & 56.7 & 4184 & 59.8 & 24,779 & 95.4 & 4,459 & 91.5 \\
\quad 140 kVp & 480 & 2.36 & 134 & 2.60 & 2155 & 30.8 & 2 & 0.00 & 268  & 5.50 \\
\quad Other & 450 & 2.20 & 117 & 2.30 & 22 & 0.00 & 423 & 1.63 & 72 & 1.47 \\
\multicolumn{9}{l}{\textbf{Slice thickness}} \\
\quad 1-3mm & 19,904 & 99.9 & 4,995 & 99.9 & 132 & 1.88 & 25,926 & 99.7 & 4,218 & 94.5  \\
\quad $>$3mm & 18 & 0.00 & 4 & 0.00 & 6,865 & 98.1 & 60 & 0.00 & 247  & 5.53 \\
\textbf{Tube current (mA)} & 440 & 440 & -- & 161 & -- & 220 & -- & 329 & --\\
\multicolumn{9}{l}{\textbf{Scanner Manufacturer}} \\
\quad GE Healthcare & 12,399 & 61.0 & 3,119 & 60.9 & 6,971 & 99.6 & 1,104 & 4.25 & 4,458 & 91.5 \\
\quad Canon & 0 & 0.00 & 0 & 0.00 & 26  & 0.00 & 24,881 & 95.7 & 0 & 0.00 \\
\quad Siemens & 7,926 & 39.0 & 2,001 & 39.1 & 0 & 0.00 & 1 & 0.00 & 241 & 5.00 \\
\quad Philips & 2 & 0.00 & 1 & 0.00 & 0 & 0.00 & 0 & 0.00 & 173 & 3.50  \\
\bottomrule
\end{tabular}
}
\end{center}
\caption{\textit{Summary of Dataset Characteristics across internal and external sites on abdominal CTs. This includes the number of abdominal CTs, age, sex distribution, CT scan parameters (kVp, slice thickness, mean tube current), and scanner manufacturer details for each site used as a testing dataset. Values are reported as counts (n) and percentages.}}
\label{tab:dataset_characteristics}
\end{table}

\FloatBarrier


\end{document}